\documentclass[10pt,journal,compsoc]{IEEEtran}

\usepackage{multirow}
\usepackage[utf8]{inputenc} % allow utf-8 input
\usepackage[T1]{fontenc}    % use 8-bit T1 fonts
\usepackage{hyperref}       % hyperlinks
\usepackage{url}            % simple URL typesetting
\usepackage{booktabs}       % professional-quality tables
\usepackage{amsfonts}       % blackboard math symbols
\usepackage{nicefrac}       % compact symbols for 1/2, etc.
\usepackage{microtype}      % microtypography
\usepackage{amsmath}
\usepackage{ upgreek }
\usepackage{bm}
\usepackage{color,soul}
\usepackage{stmaryrd }
\usepackage{algorithm}
\usepackage{algorithmic}
\usepackage{graphicx}
\usepackage{amsmath}
\usepackage{amssymb}
\usepackage{amsthm}

\newtheorem{thm}{Theorem}
\newtheorem{lem}{Lemma}
\newtheorem{prop}{Proposition}
\newtheorem{defn}{Definition}

\begin{document}
\title{Scalable Label Propagation for Multi-relational Learning on the Tensor Product of Graphs}

\author{Zhuliu Li*,
        Raphael Petegrosso*,
        Shaden Smith,
        David Sterling,
        George Karypis,
        Rui Kuang$^{\dag}$% <-this % stops a space
\IEEEcompsocitemizethanks{\IEEEcompsocthanksitem Z. Li, R. Petegrosso, S. Smith, G. Karypis and R. Kuang are with Department of Computer Science and Engineering, University of Minnesota Twin Cities, MN, 55455, USA\protect\\
E-mail: \{lixx3617, peteg001, shaden, karypis, kuang\}@umn.edu

\IEEEcompsocthanksitem D. Sterling is with Department of Radiation Oncology, University of Minnesota Twin Cities, MN, 55455, USA\protect\\
E-mail: sterl035@umn.edu \protect\\
(*Co-first author, $^{\dag}$Corresponding author) 
}% <-this % stops a space
}

\IEEEtitleabstractindextext{%
\begin{abstract}
Multi-relational learning on knowledge graphs infers high-order relations among the entities across the graphs. 
This learning task can be solved by label propagation on the tensor product of the knowledge graphs to learn the high-order relations as a tensor. 
%Empirically, it is not feasible to compute label propagation on tensor product of more than three large graphs due to the exponential complexity of computing the tensor. 
In this paper, we generalize a widely used label propagation model to the normalized tensor product graph, and propose an optimization formulation and a scalable Low-rank Tensor-based Label Propagation algorithm (LowrankTLP) to infer multi-relations for two learning tasks, hyperlink prediction and multiple graph alignment. The optimization formulation minimizes the upper bound of the noisy tensor estimation error for multiple graph alignment, by learning with a subset of the eigen-pairs in the spectrum of the normalized tensor product graph. We also provide a data-dependent transductive Rademacher bound for binary hyperlink prediction. 
%With efficient tensor computations,  LowrankTLP takes either a sparse tensor of observed multi-relations or a CP-form tensor estimated from pairwise relations between all the graph pairs as the input, to infer the queried multi-relations among graph vertices for hyperlink prediction or multiple graph alignment.} 
We accelerate LowrankTLP with parallel tensor computation which enables label propagation on a tensor product of 100 graphs each of size 1000 in less than half hour in the simulation. LowrankTLP was also applied to predicting the author-paper-venue hyperlinks in publication records, alignment of segmented regions across up to 26 CT-scan images and alignment of protein-protein interaction networks across multiple species. The experiments demonstrate that LowrankTLP indeed well approximates the original label propagation with better scalability and accuracy.\\
\textbf{Source code:} \url{https://github.com/kuanglab/LowrankTLP}
\end{abstract}

\begin{IEEEkeywords}
multi-relational learning, tensor product graph, label propagation, hyperlink prediction, multiple graph alignment, tensor decomposition and completion
\end{IEEEkeywords}}

\maketitle

% is not selected <OR> if conference mode is selected - because compsoc
% conference papers position the abstract like regular (non-compsoc)
\IEEEdisplaynontitleabstractindextext

% \fi
\IEEEpeerreviewmaketitle

\begin{figure}[htb]
	\begin{center}
		\centerline{\includegraphics[width=\columnwidth]{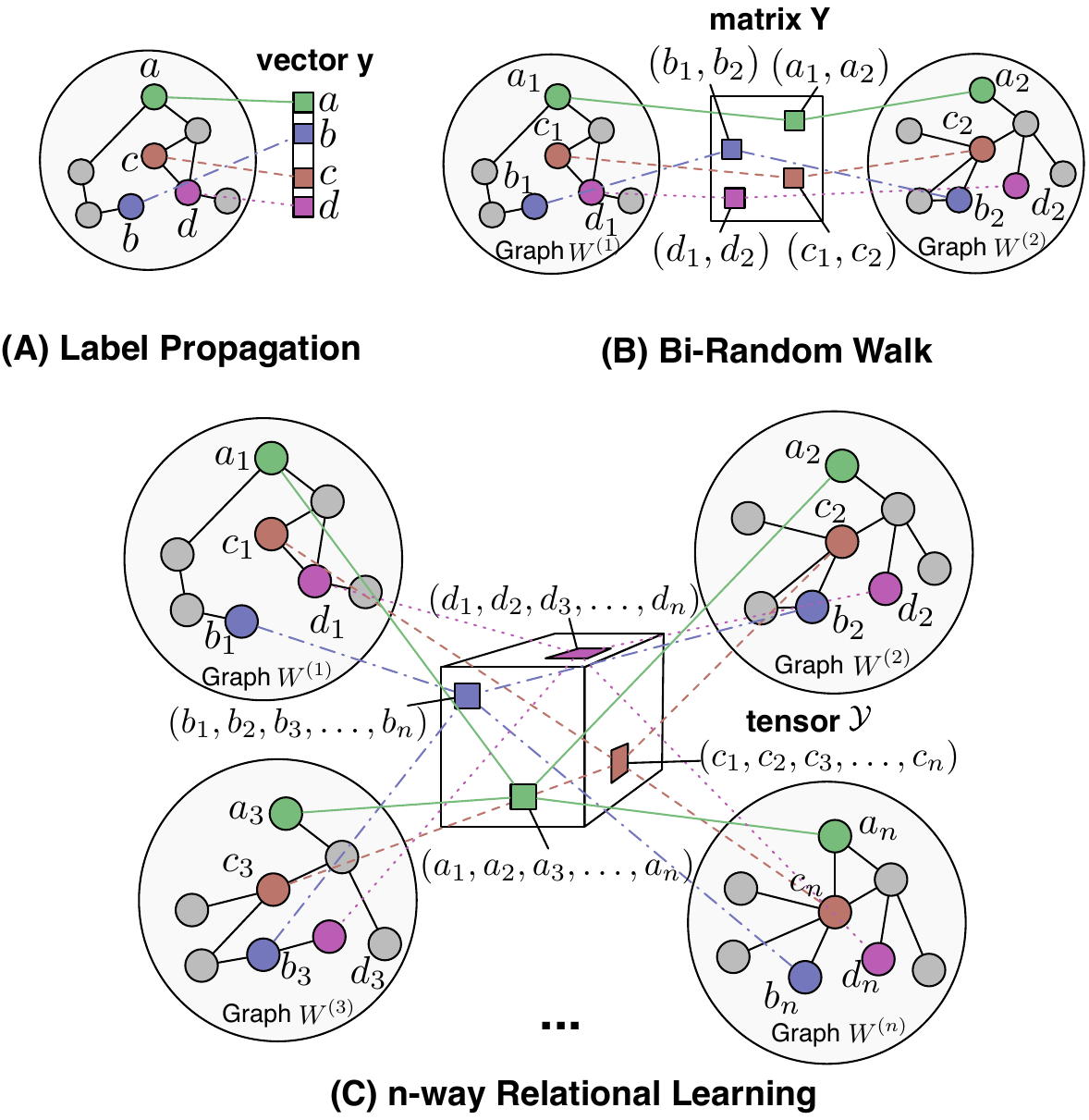}}
		\caption{{\bf Label propagation generalized on tensor product graphs.} \textbf{(A)} Label propagation on a graph predicts the labels of the vertices for semi-supervised learning; \textbf{(B)} Label propagation on the Kronecker product of two graphs predicts links between the vertices across the two graphs; \textbf{(C)} Label propagation on an $n$-way tensor product graph learns $n$-way multi-relations across the vertices in $n$ graphs. Each $n$-way tuple of graph vertices in the same color is represented as an entry in the $n$-way tensor.%At the top left, label propagation on a graph predicts the labels of the nodes for semi-supervised learning; at the bottom left, label propagation on the Kronecker product of two graphs predicts the links between the nodes across the two graphs; finally, at the right, label propagation on the $N$-way tensor product graph learns the $N$-way multi-relations across the nodes in $N$ graphs. Each set of $N$ nodes in the same color is represented by an entry in the $N$-way tensor.
		}
		\label{TLP}
	\end{center}
	\vskip -0.3in
\end{figure}

\ifCLASSOPTIONcompsoc
\IEEEraisesectionheading{\section{Introduction}\label{sec:introduction}}
\else
\section{Introduction}
\fi
\IEEEPARstart{L}{abel} propagation has been widely used for semi-supervised learning on the similarity graph of labeled and unlabeled samples \cite{szummer2001partially, zhu2002learning, zhou2003learning}. As illustrated in Figure \ref{TLP} (A), label propagation propagates training labels on a graph to learn a vector $\boldsymbol{y}$ predicting the labels of vertices. 
A generalization of label propagation on the Kronecker product of two graphs (also called bi-random walk) can infer the pairwise relations in a matrix $Y$ between the vertices from the two graphs as shown in Figure \ref{TLP} (B). This approach has been applied to aligning biological and biomedical networks \cite{singh2008global, xie2012prioritizing}. Similar learning problems on Kronecker product graphs also exist in link prediction \cite{kashima2009link,raymond2010fast}, matching images \cite{duchenne2011tensor}, image segmentation \cite{yang2013affinity}, collaborative filtering \cite{liu2015bipartite}, citation network analysis \cite{liu2016cross} and multi-language translation \cite{xu2016cross}. When applied on the tensor product of $n$ graphs to predict the matching across the $n$ graphs in an $n$-way tensor $\mathcal{Y}$ as shown in Figure \ref{TLP} (C), label propagation accomplishes $n$-way relational learning from knowledge graphs \cite{nickel2015review}. Label propagation on the tensor product graph explores the graph topologies for associating vertices across the graphs assuming the global relations among the vertices reveal the vertex identities \cite{xie2012prioritizing}. However, the tensor formulation of label propagation is computationally intensive to solve. Empirically, most of the existing methods are only scalable to learn 3-way relations in large graphs even if the graphs are sparse  \cite{kashima2009link,liu2016cross}. In particular, each multiplication with tensor will exponentially increase the number of nonzero entries and after one or two iterations, a dense tensor is expected. 
The main objective of this study is to provide a principled and theoretical approximation approach, and scalable algorithms and implementations to tackle the scalability issue. Our contributions in this paper are summarized as follows:
\begin{itemize}
\item We propose a novel optimization formulation to approximate the transformation matrix in the closed-form solution of label propagation on the tensor product graph, by learning with a subset of eigen-pairs from the normalized tensor product graph. We provide a theoretical justification that the globally optimal solution of the optimization problem minimizes an estimation error bound of recovering the true tensor that is structured by the tensor product graph manifolds, for multiple graph alignment. We also provide a data-dependent error bound using the \emph{transductive Rademacher complexity} for binary hyperlink prediction.
\item We develop an efficient eigenvalue selection algorithm to sequentially select the eigen-pairs from each individual normalized graph considering the global spectrum of the transformation matrix. We then show that the spectrum of the low-rank normalized tensor product graph constructed by the selected eigen-pairs is guaranteed to be the globally optimal solution to the proposed optimization formulation.
\item We propose a Lowrank Tensor-based Label Propagation algorithm (LowrankTLP) using efficient tensor operations to compute the approximated solution based on the selected eigen-pairs from the knowledge graphs. We provide an efficient parallel implementation of LowrankTLP using SPLATT library \cite{splattsoftware} with shared-memory parallelism to increase the scalability by a large magnitude.
\item Our work comprehensively generalizes the label propagation model proposed in \cite{zhou2003learning} on tensor product of multiple graphs in several aspects including normalized tensor product graph, regularization framework, iterative tensor-based label propagation algorithm and computation of the closed-form solution.
\item We validate the effectiveness, efficiency and scalability of LowrankLTP on the simulation data by controlling the graph size and topology. We also demonstrate the practical use of LowrankLTP on three real datasets for hyperlink prediction and multiple graph alignment, across a large number of knowledge graphs.
\end{itemize}
\vskip -0.1in

\begin{figure*}[ht]
	\begin{center}
		\centerline{\includegraphics[width=\textwidth]{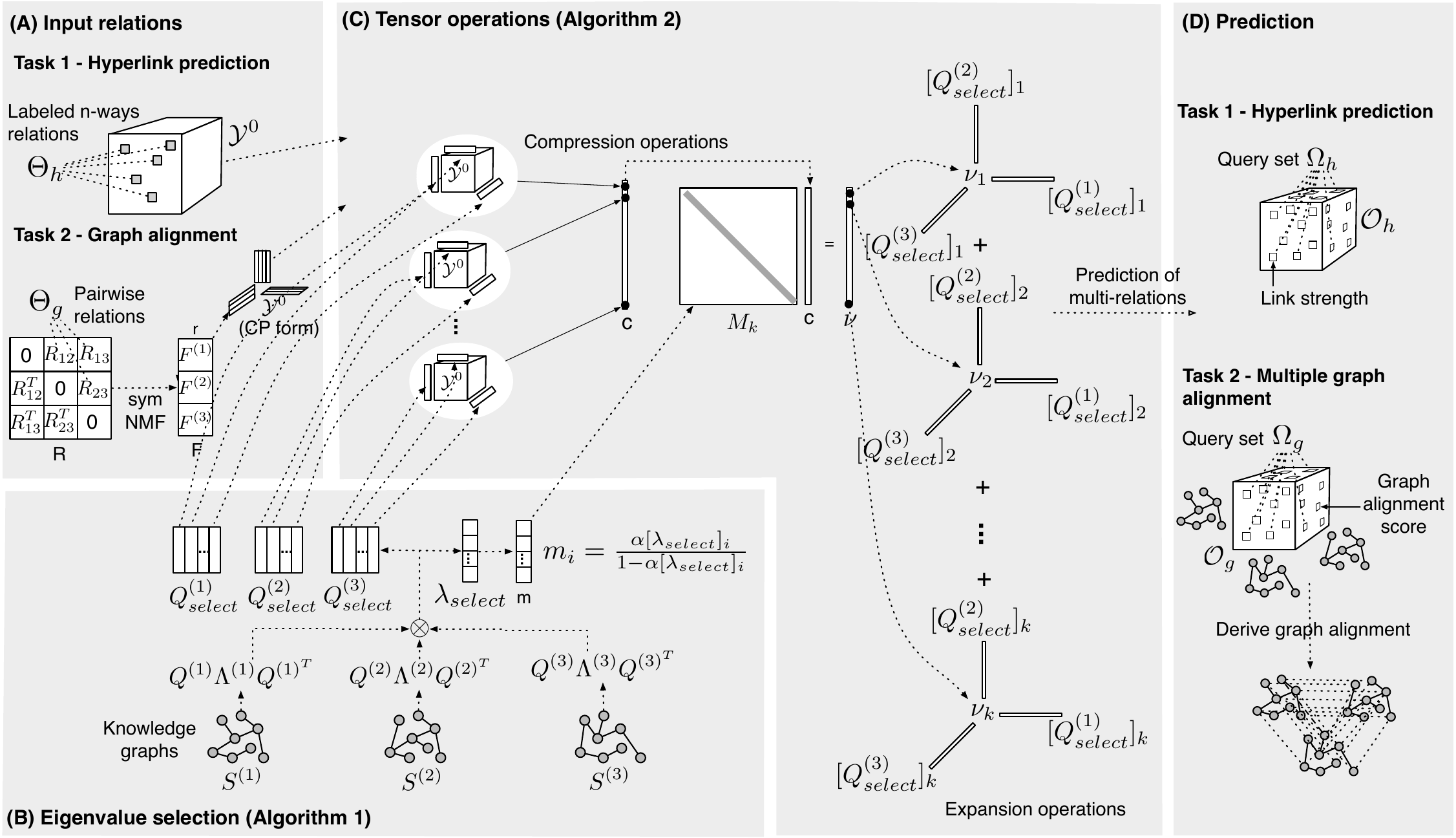}}
		\vskip -0.1in
		\caption{The overview of LowrankTLP illustrated on the tensor product of three graphs. \textbf{(A) Input:} the initial input tensor $\mathcal{Y}^0$ is 1) a sparse tensor of labeled multi-relations (given in set $\Theta_h$) for hyperlink prediction or 2) a CP-form estimated from pairwise similarities (given in set $\Theta_g$) between every pair of the graphs for multiple graph alignment. \textbf{(B) Knowledge Graphs:} Three normalized undirected graphs $S^{(1)}$, $S^{(2)}$ and $S^{(3)}$ are given. Algorithm \ref{alg:selecteig} will obtain the $k$ optimal eigenvalues and the corresponding eigenvectors of the tensor product graph based on the eigen-decomposition of each graph for computing the approximation of the closed-form solution. \textbf{(C) Efficient Tensor Operations:} $\mathcal{Y}^0$ and the selected eigen-pairs are used to perform compression and prediction operations to obtain the approximated closed-form solution of label propagation in a CP-form. \textbf{(D) Output:} The scores of the hyperlink strengths queried in set $\Omega_h$ are predicted in a sparse tensor $\mathcal{O}_h$; or, the scores of the graph alignment queried in the set $\Omega_g$ are predicted in a sparse tensor $\mathcal{O}_g$, which are then used to derive the graph alignment.}
		\label{lrTLP}
	\end{center}
	\vskip -0.3in
\end{figure*}

\section{Problem formulation} \label{sec:problem}
We first define the notations and the tensor product graph. Several useful lemmas and the definitions of tensor CANDECOMP/PARAFAC (CP) decomposition and Tucker decomposition are also given in Appendix \ref{sec:lemma}. For more general tensor computations, we direct the readers to the survey paper \cite{kolda2009tensor}. \\ \\
\textbf{Notations and operators:} 
\begin{table}[h]
	\centering
	\label{my-label}
	\resizebox{\columnwidth}{!}{\begin{tabular}{llllll}
			vector: & $\boldsymbol{x}$ & Hadamard product: &  $\circledast$ & outer product:   & $\circ$ \\
			matrix: & $X$ & Kronecker product: & $\otimes$ & the $i$-mode product: & $\times_i$ \\
			tensor: & $\mathcal{X}$ & Khatri–Rao product: & $\odot$ & vectorization of tensor: & $vec(\mathcal{X})$
	\end{tabular}}
\end{table} \\
\textbf{Tensor product graph (TPG):} 
Let $\{W^{(i)} :  i=1,\dots,n\}$ be $n$ distinct symmetric matrices,  where each $I_i \times I_i$  matrix $W^{(i)}$ denotes the adjacency matrix of the $i$-th undirected graph with $I_i$ vertices. The tensor product graph adjacency matrix is defined as $W = \otimes_{i=1}^n W^{(i)}$ with dimension $(\prod_i^n I_i) \times (\prod_i^n I_i)$. The edge  $W_{(a_1, a_2, \dots, a_n), (b_1, b_2, \dots, b_n)} = \prod_{l=1}^n W^{(l)}_{a_l,b_l}$ of $W$ encodes the similarity between a pair of $n$-way tuples of graph vertices $(a_1, a_2, \dots, a_n)$ and $(b_1, b_2, \dots, b_n), \forall \{a_l, b_l\} \in [1,I_l]$ as illustrated in Figure \ref{TLP} (C). The network properties of the TPG including the spectral distribution are discussed in \cite{leskovec2010kronecker}.

Now, we introduce the two multi-relational learning problems studied in this paper. The objective is to score the queried $n$-way relations among the vertices across multiple undirected knowledge graphs $\{W^{(i)} :   i=1, \dots, n\}$, for either hyperlink prediction or multiple graph alignment given the input as 1) the labels of a small number of observed $n$-way relations,  2) or the similarity scores between the vertices of every pair of graphs respectively, as shown in Figure \ref{lrTLP} (A). \\ \\
\emph{\textbf{Task 1: hyperlink prediction}}
      	\begin{itemize}
		\item \emph{Input relations}:  a small set $\Theta_h = \{(i_1, i_2, \dots, i_n)  :   i_j \in [1,I_j], \forall  j = 1,\dots, n \}$ of labeled  $n$-way relations (hyperlinks) among the vertices across $n$ graphs, where $i_j$  denotes the $i$-th vertex of graph $W^{(j)}$.
		\item \emph{Queried multi-relations}:  a set $\Omega_h = \{(j_1, j_2, \dots, j_n)  :  j_l \in [1,I_l],  \forall l = 1,\dots, n $\}  of the queried $n$-way relations (hyperlinks), chosen per user's interests.
		\item \emph{Learning task}: given $n$ knowledge graphs and the set $\Theta_h$ of labeled hyperlinks, predict the link strengths  of the queried set $\Omega_h$ of hyperlinks in a sparse tensor $\mathcal{O}_h \in  \mathbb{R}^{I_n\times I_{n-1} \times ... \times I_1}$ with $|\Omega_h|$ nonzero entries.
		\end{itemize} 
\emph{\textbf{Task 2: multiple graph alignment}}
      	\begin{itemize}
		\item \emph{Input relations}:  a set  $\Theta_g = \{R_{ij} \in \mathbb{R}_+^{I_i \times I_j} : \forall i,j \in [1,n] \ \text{and} \  i <  j\}$  of non-negative matrices, where $R_{ij}$ holds the  similarity scores between the vertices of a pair of graphs $W^{(i)}$ and $W^{(j)}$.
		\item \emph{Queried multi-relations}:  a set $\Omega_g= \{(j_1, j_2, \dots, j_n)  :  j_l \in [1,I_l],  \forall l = 1,\dots, n $\} of the queried $n$-way relations, which can be derived from the pairwise relations by heuristic as in \cite{gligorijevic2015fuse} and \cite{hashemifar2016joint}.
		\item \emph{Learning task}: given $n$ knowledge graphs and the set $\Theta_g$ of pairwise relations, predict the alignment scores between  the queried set $\Omega_g$ of $n$-way tuples  of graph vertices in a sparse tensor  $\mathcal{O}_g \in  \mathbb{R}^{I_n\times I_{n-1} \times ... \times I_1}$ with $|\Omega_g|$ nonzero entries. 
		\end{itemize}

As outlined in Figure \ref{lrTLP}, given either the labels of the observed $n$-way relations or the similarity scores between the vertices of all the graph pairs, we propose a scalable label propagation algorithm for manifold learning on the tensor product graph (TPG) to infer the scores of a query set of $n$-way relations in a sparse output tensor, based on the topological information of the knowledge graphs.

\section{Label propagation on tensor product graph} \label{sec:lpt}
In this section, we focus on generalizing the graph-based semi-supervised learning model proposed in \cite{zhou2003learning} to the tensor product of multiple graphs. We first derive the normalized tensor product graph (TPG), and then generalize the objective function for graph-based semi-supervised learning on TPG to solve the multi-relational learning tasks defined in Section \ref{sec:problem}. Next, we generalize the label propagation algorithm on TPG and give its closed-form solution. Finally, we analyze the scalability issues of applying label propagation on the tensor product of multiple graphs. 

%Let $W$ be the adjacency matrix of  a graph with $m$ vertices, and $D$ be the degree matrix of $W$. In \cite{zhou2003learning}, labels of the graph vertices in a vector $\boldsymbol{y} \in \mathbb{R} ^ m$ can be inferred from a small portion of observed labels in a sparse vector $\boldsymbol{y}^0 \in \mathbb{R} ^ m$, through  minimizing the the following cost function
%\begin{equation}
%\mathcal{J}(\boldsymbol{y})=\frac{1}{2}(\sum_{i,j=1}^m W_{i, j}(\frac{\boldsymbol{y}_i}{\sqrt{D_{i, i}}}-\frac{\boldsymbol{y}_j}{\sqrt{D_{j, j}}})^2+\mu \sum_{i=1}^m (\boldsymbol{y}_i-\boldsymbol{y}^0_i)^2). \label{eq:reg}
%\end{equation} 
%
%The first term $\sum_{i,j=1}^m W_{i, j}(\frac{\boldsymbol{y}_i}{\sqrt{d_{i, i}}}-\frac{\boldsymbol{y}_j}{\sqrt{D_{j, j}}})^2 = \boldsymbol{y}^T (I-D^{-\frac{1}{2}} W D^{-\frac{1}{2}}) \boldsymbol{y}$ is often called the \emph{smoothness constraint} or \emph{graph regularization} which force the labeling of the vertices $i$ and $j$ to be similar if their connection in the graph is strong. The term $D^{-\frac{1}{2}} W D^{-\frac{1}{2}}$ is called  \emph{normalized graph}, whose edges is normalized by the corresponding vertices degree. The second term in $\mathcal{J}(\boldsymbol{y})$ is called \emph{fitting constraint}, which penalize the difference between the learning labels and the initial labels. $\mu$ is a hyperparameter balancing the impacts of both terms. In this section, we propose the generalization of label propagation on TPG, to solve the two learning tasks defined in Section \ref{sec:problem}.
\subsection{Normalized TPG}
Let $S^{(i)}=[D^{(i)}]^{-\frac{1}{2}}W^{(i)}[D^{(i)}]^{-\frac{1}{2}}$ be the normalized graph of $W^{(i)}$ for $i=1,\dots, n$, where  $D^{(i)}$ is the degree matrix. The TPG $W = \otimes_{i=1}^n W^{(i)}$ has its normalization $S = D^{-\frac{1}{2}}WD^{-\frac{1}{2}}$ derived as follows
\begin{align}
S & =(\otimes_{i=1}^n [D^{(i)}]^{-\frac{1}{2}}) (\otimes_{i=1}^n W^{(i)})(\otimes_{i=1}^n [D^{(i)}]^{-\frac{1}{2}}) \label{eq:NTPG_1}\\
& = \otimes_{i=1}^n ([D^{(i)}]^{-\frac{1}{2}}W^{(i)} [D^{(i)}]^{-\frac{1}{2}})\label{eq:NTPG_2} \\
& = \otimes_{i=1}^n S^{(i)}. \nonumber
\end{align} 
Equation \eqref{eq:NTPG_1} is obtained by the fact that the degree matrices are diagonal. Equation \eqref{eq:NTPG_2} is obtained by Appendix \ref{sec:lemma} Lemma \ref{l1}. Using Appendix \ref{sec:lemma} Lemma \ref{l4}, it can be shown that the eigenvalues of $S$ are bounded between -1 and 1. This property will be used later in the derivations in the forthcoming sections.

\subsection{Regularization framework with normalized TPG} 
\label{sec: regTPG}
Let $\mathcal{Y}^0 \in \mathbb{R}^{I_n\times I_{n-1} \times ... \times I_1}$ be an initial tensor which is either incomplete with labels of the observed $n$-way relations or complete with noisy labels of all $n$-way relations,  the true labels of all the $n$-way relations can be inferred in tensor $\mathcal{Y} \in \mathbb{R}^{I_n\times I_{n-1} \times ... \times I_1}$ by minimizing the following objective function:
\begin{align}
\mathcal{J}(\mathcal{Y}) = &\frac{1}{2}( vec(\mathcal{Y})^T (I - \otimes_{i=1}^n S^{(i)})vec(\mathcal{Y}) \nonumber \\
& +\mu ||vec(\mathcal{Y} )-vec(\mathcal{Y}^0 )||_2^2). \label{eq:regTPG}
\end{align} 
The first term in $\mathcal{J}(\mathcal{Y})$ is called \emph{smoothness constraint} or \emph{graph regularization}, where $I - \otimes_{i=1}^n S^{(i)}$ is called  \emph{normalized graph
Laplacian} of the TPG,  ensures the values (inferred multi-relations) in tensor $\mathcal{Y}$  to be smooth on the manifolds  of the normalized TPG $S$. In other words, the degree-normalized $(a_1,a_2,\dots,a_n)$-th and the $(b_1,b_2,\dots,b_n)$-th entries in tensor $\mathcal{Y}$  are forced to be close if the
edge weight  $W_{(a_1,a_2,\dots,a_n),(b_1,b_2,\dots,b_n)} = \prod_{l=1}^n W^{(l)}_{a_l,b_l}$ is large. The second term in $\mathcal{J}(\mathcal{Y})$ is called \emph{fitting constraint}, which penalizes the difference between the inferred tensor $\mathcal{Y}$ and its initialization $\mathcal{Y}^0$, where $\mu > 0$ is a hyperparameter balancing the impacts of both terms. 
\begin{itemize}
	\item \emph{\textbf{Formulation of hyperlink prediction (Task 1)}}: the learning task 1 is a \emph{transductive learning} problem \cite{el2009transductive, gu2012towards} of inferring a tensor $\mathcal{Y}$ from a sparse initial tensor $\mathcal{Y}^0$. The nonzero entries in $\mathcal{Y}^0$ are the labels (link types) of the observed $n$-way relations (hyperlinks) given in set $\Theta_h$ (defined in Section \ref{sec:problem}), such that the label of the $(i_1,i_2,\dots,i_n)$-th hyperlink is $\mathcal{Y}_{i_n,i_{n-1},\dots,i_1}^0$. The zero entries in $\mathcal{Y}^0$ represent the unobserved $n$-way relations. The inferred tensor $\mathcal{Y}$ is composed of the link strengths of all the $n$-way hyperlinks.
	\item \emph{\textbf{Formulation of multiple graph alignment (Task 2)}}: we convert the set $\Theta_g$ of pairwise similarity matrices to the rank-$r$ CP-form (Appendix \ref{sec:lemma} Definition \ref{def:CPD}) of $\mathcal{Y}^0$ as following: first, symmetric NMF (symNMF) \cite{ding2005equivalence} is applied on a symmetric matrix $R$ built by stacking all $R_{ij}$'s to obtain a nonnegative factor matrix $F \in \mathbb{R}_+^{(\sum_{i=1}^n I_i) \times r}$ such that $R \approx F F^T$  as illustrated in Figure \ref{lrTLP} (A). Then, the rank-$r$ CP-form of $\mathcal{Y}^0$ is approximated as $\mathcal{Y}^0 = \llbracket F^{(n)},F^{(n-1)},\dots, F^{(1)} \rrbracket$ where  $F^{(i)}\in \mathbb{R}_+^{I_i \times r}$ is the $i$-th submatrix of $F$. The assumption is that more similar pairwise relations between every pair $(i_a,i_b) \subset (i_1, i_2, \dots, i_n)$ imply a stronger $n$-way relation in tuple $(i_1, i_2, \dots, i_n)$. This representation has been widely adopted in real graph alignment problems as in \cite{gligorijevic2015fuse}, \cite{chen2016fascinate} and \cite{wang2018antenna}. As $\mathcal{Y}^0$ is guessed from the pairwise relations, we call the learning task 2 \emph{structured signal recovery from noisy observation}. Our goal is to recover the true tensor $\mathcal{Y}$ which is structured by the TPG manifolds, from its noisy observation $\mathcal{Y}^0$.  
\end{itemize}

\subsection{Label propagation algorithm on normalized TPG} \label{sec: LPTPG}
The objective function $\mathcal{J}(\mathcal{Y})$ in Equation \eqref{eq:regTPG} can be minimized by performing the fixed-point iteration \eqref{eq4}, which is a generalization of the graph-based semi-supervised learning algorithm in \cite{zhou2003learning} to TPG.
\begin{align} 
vec(\mathcal{Y}^{t+1}) =\alpha (\otimes_{i=1}^n S^{(i)}) vec(\mathcal{Y}^t) + (1-\alpha)
vec(\mathcal{Y}^0), \label{eq4}
\end{align}
where $\alpha = \frac{1}{1+\mu} \in (0,1)$ is a balancing hyperparameter and $t$ denotes the iteration number. Applying the \emph{vectorization property} of Tucker decomposition (Appendix \ref{sec:lemma} Definition \ref{def:tucker}), the Equation \eqref{eq4} can be rewritten as 
\begin{align}
\mathcal{Y}^{t+1} =\alpha \mathcal{Y}^t \times_1 S^{(n)} \times_2 S^{(n-1)} \dots \times_n S^{(1)}+ (1-\alpha)\mathcal{Y}^0.\nonumber
\end{align} 
Even if $\mathcal{Y}^0$ is in a sparse form, the density of tensor $\mathcal{Y}$ increases exponentially in each iteration. Therefore, the space complexity is $O(\prod_{i=1}^n I_i)$ for store the dense tensor and the time complexity is $O((\prod_{i=1}^n I_i)(\sum_{i=1}^n I_i))$ per iteration. Due to the necessity of computing the full tensor, the same space and time complexities are required by the \emph{Link Propagation} method proposed in \cite{kashima2009link}, which applies conjugate gradient descent \cite{Bertsekas/99} to minimize a similar objective function defined on the original (unnormalized) TPG.

Since the eigenvalues of $S$ are in $[-1, 1]$ and $\alpha \in (0,1)$, iteration \eqref{eq4} converges to the following closed-form solution of $\mathcal{J}(\mathcal{Y})$ as
\begin{equation}
vec(\mathcal{Y}^*) = \lim_{t \to \infty} vec(\mathcal{Y}^t) =(1-\alpha)(I-\alpha S)^{-1}
vec(\mathcal{Y}^0). \label{eq:oldcf}
\end{equation}
Furthermore, given the eigen-decomposition of each $S^{(i)}$ as $\{S^{(i)} = Q^{(i)} \Lambda^{(i)} Q^{(i)T} :  i=1, \dots, n \}$,  the eigen-decomposition of $S$ can be expressed as 
\begin{align}
S =  Q \Lambda Q^T= (\otimes_{i=1}^n Q^{(i)})(\otimes_{i=1}^n \Lambda^{(i)}) ( \otimes_{i=1}^n Q^{(i)T} ) \nonumber, 
\end{align}
according to Appendix \ref{sec:lemma} Lemma \ref{l1} and \ref{l4}.
%an eigen-decomposition $S = Q \Lambda Q^T$ can simplify the closed-form solution as  
%\begin{align} 
%\overrightarrow{\mathcal{Y}^*} & =(1-\alpha)(I-\alpha S)^{-1}
%\overrightarrow{\mathcal{Y}^0} \label{eq5} \nonumber \\
%   & =(1-\alpha)Q(I-\alpha \Lambda)^{-1} Q^T
%\overrightarrow{\mathcal{Y}^0}. \
%\end{align}
Substituting $S$ into Equation \eqref{eq:oldcf} we have 
\begin{align}
vec(\mathcal{Y}^*) & =(1-\alpha)(\otimes_{i=1}^n Q^{(i)}) (I-\alpha (\otimes_{i=1}^n\Lambda^{(i)}))^{-1} \nonumber \\
   & (\otimes_{i=1}^n Q^{(i)T}) vec(\mathcal{Y}^0). \label{eq:closeform}
\end{align}
It is not hard to see that computing the closed-form solution in Equation \eqref{eq:closeform} from right to left needs $2n$ matrix-tensor products in total with the \emph{vectorization property} of Tucker decomposition (Appendix \ref{sec:lemma} Definition \ref{def:tucker}). Therefore, the space and time complexity will be the same as running two iterations of label propagation, apart from computing the eigen-decompositions of all the normalized graphs $\{S^{(i)} :  i=1,\dots,n\}$. To tackle this challenge of computing label propagation of a high-order $n$-way tensor on TPG, we propose the LowrankTLP algorithm based on a principled approximation of the linear transformation matrix $(I-\alpha S)^{-1}$ in the closed-form solution in Equation \eqref{eq:oldcf} in the next section.
%In related works \cite{raymond2010fast} and \cite{liu2016cross}, Eckart-Young-Mirsky theorem \cite{eckart1936approximation} is applied on each individual graph to approximate their TPG. However, this approximation does not consider the global spectral distribution of the TPG (as stated in Lemma \ref{l5}) and thus, leads to a large error in both approximation of the TPG and prediction of the multi-relations. To scale the label propagation on the tensor product of multiple high dimensional graphs, we propose an lowrankTLP algorithm based on a proper approximation of the linear transformation matrix $(I-\alpha S)^{-1}$ in Section \ref{sec:lowrankTLP} .

\section{Low-rank label propagation} \label{sec:lowrankTLP}
In this section, we first propose an optimization formulation to approximate the closed-form solution in Equation \eqref{eq:oldcf}; then we develop Algorithm \ref{alg:selecteig} to select a subset of  eigen-pairs from the normalized tensor product graph $S$ which are guaranteed to be the optimal solution to the proposed optimization formulation. Next, we propose the LowrankTLP algorithm (illustrated in Figure \ref{lrTLP}) for scalable label propagation on TPG, using the selected eigen-pairs. Finally, we provide the theoretical justification of our optimization formulation for \emph{Task 2} by proposing an estimation error bound of recovering the true tensor that is structured by the TPG manifolds; we also provide a data-dependent error bound for a special case of \emph{Task 1}. 

\subsection{Optimization formulation} \label{sec:optimization}
We propose to approximate the closed-form solution in Equation \eqref{eq:oldcf} by minimizing the perturbation on transformation matrix $(I- \alpha S)^{-1}$ as follows,
\begin{equation}
\label{eq: optimization}
\begin{aligned}
& \underset{\emph{eig}(S_k)}{\text{minimize}}
& & ||(I- \alpha S)^{-1} - (I- \alpha S_k)^{-1}||_{2,F} \\
& \text{subject to}
& & \text{rank}(S_k)=k,\ \emph{eig}(S_k) \subseteq \emph{eig}(S), 
\end{aligned}
\end{equation}
where $S = \otimes_{i=1}^n S^{(i)}$ is the normalized TPG; $\emph{eig}(S_k)$ and $\emph{eig}(S)$ denote the sets of eigen-pairs of $S_k$ and $S$ respectively; $||.||_2$ is spectral norm and $||.||_F$ is Frobenius norm. 

The objective is to find a low-rank matrix $S_k$ defined by a subset of eigen-pairs of $S$ to give the lowest divergence on the overall transformation matrix $(I- \alpha S)^{-1}$.  In Section \ref{sec: task1 bound},  we will show this formulation minimizes the estimation error bound in Theorem \ref{the:task2}.
% \begin{lem}\label{l6}
% Let matrix $\tilde{S}_k$ denote the best rank-$k$ approximation to  $S$ per Eckart-Young-Mirsky theorem. $\tilde{S}_k$ is not guaranteed to be the optimal solution to the optimization problem \eqref{eq: optimization}.
% \end{lem}
It is noteworthy that simply computing the best rank-$k$ approximation to $S$ per Eckart-Young-Mirsky theorem does not guarantee the optimal solution. Instead, we will show that the global optimal solution to the optimization problem \eqref{eq: optimization} can be efficiently found by Algorithm \ref{alg:selecteig}.

\subsection{Selection of the optimal eigen-pairs} \label{sec:Select eig}
Let $S_k = Q_{1:k} \Lambda_{1:k} Q_{1:k}^T$ be the eigen-decomposition of $S_k$, where $\Lambda_{1:k}$ and $Q_{1:k}$ store the eigen-pairs $\{ (\lambda_j, \boldsymbol{q}_j) :   j=1,\dots, k \}$ of $S_k$ selected from $\emph{eig}(S)$. Also, define diagonal matrix $\Lambda_{\text{rest}}$ and matrix  $Q_{\text{rest}}$ to hold the remaining eigen-pairs $\{ (\lambda_i, \boldsymbol{q}_i) :  i=k+1,\dots, N \}$ of $\emph{eig}(S)$, where $N= \prod_{l=1}^n I_l$.  According to Appendix \ref{sec:lemma} Lemma \ref{l4},  we have 
\begin{align}
\lambda_j=\prod_{i=1}^n \lambda^{(i)}_j \ \ \text{and} \ \
\boldsymbol{q}_j=\otimes_{i=1}^n \boldsymbol{q}^{(i)}_j, \forall j=1,\dots, k, \nonumber
\end{align}
where $\lambda^{(i)}_j$ and $\boldsymbol{q}^{(i)}_j$ is an eigen-pair of $S^{(i)}$ contributing to $\lambda_j$ and $\boldsymbol{q}_j$ of $S$. This implies the eigen-pairs of $S_k$ are composed of the properly selected eigen-pairs from each $S^{(i)}$, for $i=1,\dots,n$.

\begin{prop} \label{the:lrrep} Define $A=(I-\alpha S)^{-1}$ and its approximation $\hat{A}=(I-\alpha S_k)^{-1}$. According to Woodbury formula \cite{higham2002accuracy}, we have
	\begin{align}
	\hat{A} = Q_{1:k}((I- \alpha \Lambda_{1:k})^{-1}-I)Q_{1:k}^T + I.
	\label{eq:approxA}
	\end{align} 
\end{prop}

%Proof: by definition of $S_k$ we have
%$
%\hat{A}  =(I-\alpha Q_{1:k} \Lambda_{1:k} Q_{1:k}^T)^{-1}$, which can be further expanded as $Q_{1:k}((I- \alpha \Lambda_{1:k})^{-1}-I)Q_{1:k}^T + I $
%by Woodbury formula \cite{higham2002accuracy}. (End of Proof)
\begin{thm} \label{the: perturbation}
	The optimal $k$ eigenvalues $\{\lambda_j :  j=1,\dots,k \}$ that solve the optimization problem \eqref{eq: optimization} are among the union of the $k$ largest (algebraic) and  $k$ smallest (algebraic) eigenvalues of $S$ and satisfy the following condition
	$$ 
	\frac{\alpha |\lambda_j|}{1-\alpha\lambda_j} \geq  \frac{\alpha |\lambda_i|}{1-\alpha\lambda_i}, \forall j\in [1,k], \forall i\in[k+1,N].
	$$
\end{thm}
\begin{proof}
Given Equation \eqref{eq:approxA}, the perturbation can be obtained as
\begin{align}
\hat{A}-A = Q_{\text{rest}} (I-(I-\alpha  \Lambda_{\text{rest}})^{-1})Q_{\text{rest}}^T, \nonumber
\end{align}
whose singular values are $\{\frac{\alpha |\lambda_{i}|}{1-\alpha\lambda_{i}} :  i=k+1,\dots, N \}$ and $k$ zeros. Thus, its spectral norm and Frobenius norm are 
\begin{align} \label{eq:purtspectral}
\begin{split}
||\hat{A}  - A||_2 &= \frac{\alpha |\lambda^*|}{1-\alpha\lambda^*} \text{ and}\ \\ ||\hat{A}  - A||_F &= \sqrt{\sum_{i=k+1}^N (\frac{\alpha |\lambda_i|}{1-\alpha\lambda_i})^2},   
\end{split}
\end{align}
where 
$\lambda^* = \text{argmax}_{\lambda \in \{\lambda_{k+1} ,\dots, \lambda_N\}} \frac{\alpha |\lambda|}{1-\alpha\lambda}$.
To minimize both norms in Equation \eqref{eq:purtspectral}, the $k$ selected eigenvalues $\{\lambda_j :   j=1,\dots,k \}$ should produce the largest elements in the set $\{\frac{\alpha |\lambda_j|}{1-\alpha\lambda_j} :   j=1,\dots,k \} $ among all the eigenvalues of $S$.  In addition, since $\alpha \in (0,1)$ and $\lambda_j \in [-1,1]$ for $ j=1,\dots, k$, the function $\frac{\alpha |\lambda_j|}{1-\alpha\lambda_j}$ is monotonically increasing in the positive orthant and decreasing in the negative orthant with $\lambda_j$. Thus, $\{\lambda_j :  j=1,\dots,k \}$ must be in the union of the $k$ largest (algebraic) eigenvalues and $k$ smallest (algebraic) eigenvalues of $S$. (End of Proof) 
\end{proof}

\begin{thm} \label{the:selecteig}
 Define function $\textbf{top\_bot\_2k}(\boldsymbol{x}) = \textbf{top\_k}(\boldsymbol{x}) \cup  \textbf{bot\_k}(\boldsymbol{x})$ where  $\textbf{top\_k}(\boldsymbol{x})$ and  $\textbf{bot\_k}(\boldsymbol{x})$ return the $k$ algebraically largest and smallest values of the vector $\boldsymbol{x}$ respectively. Given the vector $\boldsymbol{\lambda}^{(i)}$ of the eigenvalues of $S^{(i)}$ for $i=1,\dots, n $, we have
	\begin{align*}
	& \textbf{top\_bot\_2k}( \otimes_{i=1}^{n}\boldsymbol{\lambda}^{(i)}) = \\ &\textbf{top\_bot\_2k}(\boldsymbol{\lambda}^{(n)} \otimes \textbf{top\_bot\_2k}( \Gamma^{(n-1)})), \ \text{where}\\
	& \Gamma^{(i)}=
	\begin{cases}
	\boldsymbol{\lambda}^{(i)}\otimes \textbf{top\_bot\_2k}(\Gamma^{(i-1)}), & \text{if}\ \ i=2, \dots, n-1 \\
	\boldsymbol{\lambda}^{(1)}, & \text{if}\ \ i=1.
	\end{cases}
	\end{align*}
\end{thm} 
\begin{proof} 
Theorem \ref{the:selecteig} can be proven by induction based on the observation that the $k$ algebraically largest (smallest) elements in the outer product of two real vectors can only be among the multiplications between the union of the $k$ largest and smallest values  in the two vectors. Thus, only the numbers in $\textbf{top\_bot\_2k}(\Gamma^{(i-1)})$ are needed to compute the next $\Gamma^{(i)}$ in the recursion. Taking the elements in $\textbf{top\_bot\_2k}(\Gamma^{(i-1)})$ in the multiplication with each $\boldsymbol{\lambda}^{(i)}$ guarantees that the numbers needed for computing the $k$ largest (smallest) elements in $\otimes_{i=1}^{n}\boldsymbol{\lambda}^{(i)}$ will be kept in $\Gamma^{(i)}$. The
details of the proof are given in Appendix \ref{proof 1}. (End of Proof)
\end{proof} 
% \vskip -0.2in
\begin{algorithm}[]
	\caption{Select Eigenvalues}
	\label{alg:selecteig}
	\begin{algorithmic}[1]
		\STATE {\bfseries Input:}$\{S^{(i)} :  i=1,\dots, n \}$, $\alpha \in (0,1) $.
		\STATE {\bfseries Output:} $\boldsymbol{\lambda}_{\text{select}}$ and $\{Q^{(i)}_{\text{select}} : i=1,\dots,n\}$.
		\STATE Compute and store the eigenvalues and eigenvectors of $S^{(i)}$ in vector $\boldsymbol{\lambda}^{(i)}$ and matrix $Q^{(i)}$ respectively, for $i=1,\dots, n$.
		\STATE $\Gamma \leftarrow \boldsymbol{\lambda}^{(1)} $
		\FOR{$i=2$ to $n$}
		\STATE $\Gamma \leftarrow \boldsymbol{\lambda}^{(i)}\otimes \textbf{top\_bot\_2k}(\Gamma) $  
		\ENDFOR
		\STATE $\boldsymbol{\lambda}_{\text{select}} \leftarrow$  $k$ elements from $\Gamma$ with the largest $\frac{\alpha |\Gamma_j|}{1-\alpha \Gamma_j} , j=1,\dots, k$  
		\FOR{$i=n$ to $1$}
		\STATE return $Q^{(i)}_{\text{select}}$ from $Q^{(i)}$ by looking-up indexes of the values output by function $\textbf{top\_bot\_2k}()$.
		\ENDFOR
	\end{algorithmic}
\end{algorithm}

According to Theorem \ref{the: perturbation}, the selected eigenvalues $\{\lambda_j :  j=1, \dots, k\}$ from $S$ satisfying $ 
\frac{\alpha |\lambda_j|}{1-\alpha\lambda_j} \geq  \frac{\alpha |\lambda_i|}{1-\alpha\lambda_i}, \forall j\in [1,k], \forall i\in[k+1,N]
$ must be contained in the union of the $k$ largest and $k$ smallest eigenvalues of $S$. Thus, we only need to find the $\textbf{top\_bot\_2k}(\otimes_{i=1}^n \boldsymbol{\lambda}^{(i)})$ with Theorem \ref{the:selecteig}, and select $k$ eigenvalues which give the largest elements in the set $\{\frac{\alpha |\lambda_j|}{1-\alpha \lambda_j} :  j=1, \dots, k\}$.  Based on the idea, we propose Algorithm \ref{alg:selecteig} to select the eigen-pairs $\{(\lambda_j^{(i)}, \boldsymbol{q}_j^{(i)}) :  i=1,\dots, n,  j=1,\dots, k\}$ efficiently in time $O(\sum_{i=1}^n(kI_i \log(kI_i))$, plus the time for eigen-decomposition of each knowledge graph.  Algorithm \ref{alg:selecteig} starts with $\boldsymbol{\lambda}^{(1)}$, the eigenvalues of the first graph (line 4) and iteratively merges another $\boldsymbol{\lambda}^{(i)}$ one at a time in the for-loop between line 5-7 to compute $\textbf{top\_bot\_2k}(\otimes_{l=1}^i \boldsymbol{\lambda}^{(l)})$ in $\Gamma$. Each merge step computes and sorts $O(kI_i)$ numbers. Algorithm \ref{alg:selecteig} outputs a vector $\boldsymbol{\lambda}_{\text{select}}$ of the selected eigenvalues from $S$ and matrices $Q_{\text{select}}^{(i)}$ of the selected eigenvectors from $Q^{(i)}$, for $ i=1,\dots,n$ such that
\begin{align} 
\boldsymbol{\lambda}_{\text{select}} &=  [\lambda_1, \lambda_2, \dots, \lambda_k]^{T} \nonumber \\
Q_{\text{select}}^{(i)} &=[\boldsymbol{q}_1^{(i)}, \boldsymbol{q}_2^{(i)}, \dots , \boldsymbol{q}_k^{(i)}], \forall i=1,\dots, n. \nonumber
\end{align}
Define $M=(I- \alpha \Lambda_{1:k})^{-1}-I$, which is computed from $\boldsymbol{\lambda}_{\text{select}}$ as 
\begin{align}
M =\text{diag}\bigg(\bigg[\frac{\alpha \lambda_1}{1-\alpha \lambda_1 },\frac{\alpha \lambda_2}{1-\alpha \lambda_2 }, \dots , \frac{\alpha \lambda_k}{1-\alpha \lambda_k } \bigg]\bigg) \nonumber. 
\end{align}
The matrix $Q_{1:k}$ can be computed from $\{Q_{\text{select}}^{(i)} :  i=1,\dots, n\}$ as
$
Q_{1:k}= \odot_{i=1}^{n} Q_{\text{select}}^{(i)}.
$
By Equation \eqref{eq:approxA}, the closed-form solution in Equation \eqref{eq:oldcf} can be approximated as 
\begin{align} 
 vec(\widehat{\mathcal{Y}}^*) =&(1-\alpha)\hat{A} vec(\mathcal{Y}^0) \label{eq:approx_sol} \\
=&(1-\alpha)(\odot_{i=1}^{n} Q_{\text{select}}^{(i)})   M (\odot_{i=1}^{n} Q_{\text{select}}^{(i)})^T
vec(\mathcal{Y}^0) \nonumber \\
&+(1-\alpha)vec(\mathcal{Y}^0). \label{eq:lowRank}
\end{align}
\vskip -0.1in
\begin{algorithm}[tb]
	\caption{LowrankTLP }
	\label{alg:lowrankLP}
	\begin{algorithmic}[1]
		\STATE {\bfseries Input:} $\{S^{(i)} : i=1,\dots, n\}$, $\mathcal{Y}^0$, $\alpha$, $k$ and $\Omega$.
		\STATE {\bfseries Output:} Sparse tensor $O$.
		\STATE Apply Algorithm \ref{alg:selecteig} to obtain $\boldsymbol{\lambda}_{\text{select}}$, $\{Q_{\text{select}}^{(i)} : i=1,\dots, n\}$.
		\STATE Initialize $\boldsymbol{v}$ to be a $k$-D vector with all zeros.
		\IF{$\mathcal{Y}^0$ is sparse} 
		\FOR{j=1 to k}
		\STATE $\boldsymbol{v}_j\leftarrow  \mathcal{Y}^0 \bar{\times}_1 \boldsymbol{q}_j^{(n)} \bar{\times}_2 \boldsymbol{q}_j^{(n-1)} \dots \bar{\times}_n \boldsymbol{q}_j^{(1)}$ 
		\ENDFOR
		\ELSIF{$\mathcal{Y}^0$ is in CP-form $\llbracket F^{(n)},F^{(n-1)},\dots, F^{(1)} \rrbracket$}
		\STATE $\Psi \leftarrow Q^{(1)T}_{select} F^{(1)}$ 
		\FOR{j=2 to k}
		\STATE $\Psi \leftarrow \Psi \circledast (Q^{(j)T}_{select} F^{(j)})$
		\ENDFOR 
		\STATE $\boldsymbol{v} \leftarrow \Psi \boldsymbol{1}$
		\ENDIF
		\STATE $m\leftarrow  \alpha \boldsymbol{\lambda}_{\text{select}} / (1-\alpha  \boldsymbol{\lambda}_{\text{select}})$ 
		\STATE $\boldsymbol{\hat{v}'} \leftarrow (\boldsymbol{v} \circledast m)^{\boldsymbol{'}} $
		\STATE \label{line:expbegin}Initialize $\mathcal{Y^*}=\{\}$ to be an empty tensor
		\FOR{every tuple $(i_1, i_2, \dots, i_n)$ in $\Omega$}
		\STATE $\mathcal{O}_{i_n, i_{n-1}, \dots, i_1} \leftarrow (1-\alpha) (\sum_{j=1}^k  \boldsymbol{\hat{v}^{'}_j} \prod_{l=1}^n q_{i_l, k}^{(l)}+\mathcal{Y}^0_{i_n, i_{n-1}, \dots, i_1})$
		\ENDFOR 
		%with the same dimension as $\mathcal{Y}_0$ 
		%		\FOR{j = 1 to k} 
		%		\STATE $\mathcal{Y} \leftarrow \mathcal{Y} + \hat{\boldsymbol{v}}_{k-j+1}(\boldsymbol{q}_j^{(n)} \circ \boldsymbol{q}_j^{(n-1)} \circ \dots \circ \boldsymbol{q}_j^{(1)})$ 
		%		\ENDFOR \label{line:expend}
		%		\STATE $\mathcal{Y}^* \leftarrow  (1-\alpha) (\mathcal{Y} + \mathcal{Y}^0) $ 
	\end{algorithmic}
\end{algorithm}

\subsection{LowrankTLP algorithm} \label{sec:lrTLP}
Equation \eqref{eq:lowRank} implies a $2$-step tensor computation of the closed-form solution given in Algorithm \ref{alg:lowrankLP}. The two steps are also illustrated in Figure \ref{lrTLP}.\\

\noindent \textbf{\emph{Compression step}} (line 4-15 in Algorithm \ref{alg:lowrankLP})
\begin{itemize}
	\item \textbf{\emph{$\mathcal{Y}^0$ is sparse in hyperlink prediction (Task 1)}} (line 4-8):
	the role of $(\odot_{i=1}^{n} Q_{\text{select}}^{(i)})^T
	vec(\mathcal{Y}^0)$ in Equation \eqref{eq:lowRank} is to compress the original tensor $\mathcal{Y}^0$ to a $k$-D vector $\boldsymbol{v}$ with its $j$th element 
	\begin{align}
	\boldsymbol{v}_j = \mathcal{Y}^0 \bar{\times}_1 \boldsymbol{q}_j^{(n)} \bar{\times}_2 \boldsymbol{q}_j^{(n-1)} \dots \bar{\times}_n \boldsymbol{q}_j^{(1)}, \label{compression}
	\end{align}
	where each $\bar{\times}_i$ denotes mode-$i$ vector product of tensor. In Equation \eqref{compression}, the original tensor $\mathcal{Y}^0$ is compressed to a scalar by multiplying with $n$ vectors which is similar to computing the core tensor in Tucker decomposition. Denote the number of nonzeros in $\mathcal{Y}^0$ as $\boldsymbol{|}\mathcal{Y}^0\boldsymbol{|}$, the time complexity of the compression step is $O( \boldsymbol{|}\mathcal{Y}^0\boldsymbol{|} nk)$. 

\textbf{$\star$ \emph{Parallel implementation}}:
	The construction of $\boldsymbol{v}$ via Equation (\ref{compression}) performs $j$
	sequences of $n$-way tensor-vector products as 
		\begin{align}
	Z &\gets Y_{(1)}^{0} (Q_{\text{select}}^{(2)} \odot \dots \odot Q_{\text{select}}^{(n)}), \label{eq: kernel} \\
	\boldsymbol{v}_j &\gets \boldsymbol{q}_j^{(1)T} \boldsymbol{z}_j\qquad \forall j=1, \dots, k, \nonumber
	\end{align}
	where $Y_{(1)}^{0}$ denotes the matrix flattened from $\mathcal{Y}^0$.
	The kernel in Equation \eqref{eq: kernel} is similar to \emph{matricized
tensor times Khatri-Rao product (MTTKRP)} \cite{bader2007efficient} involving $n{-}1$ products during the computation of the
	CP decomposition. Therefore, we can leverage parallel algorithms developed to compute the CP decomposition for the computation. We adopt SPLATT \cite{smith2015splatt}, a C library with shared-memory parallelism  for fast \emph{MTTKRP} computation. Parallelized in $p$ threads, the parallel implementation reduces the complexity to $O(\frac{  \boldsymbol{|}\mathcal{Y}^0\boldsymbol{|} nk}{p})$.
	\textbf{\item \emph{$\mathcal{Y}^0$ is in CP-form in multiple graph alignment (Task 2)}} (line 9-15): when the initial tensor $\mathcal{Y}^0$ is in the CP-form $\llbracket F^{(n)},F^{(n-1)},\dots, F^{(1)} \rrbracket$ the $k$-D vector $\boldsymbol{v}$ can be obtained by 
	\begin{align}
	\boldsymbol{v} &= (\odot_{i=1}^n Q_{\text{select}}^{(i)} )^T (\odot_{i=1}^n F^{(i)})\boldsymbol{1} \label{eq: CP form 1} \\
	&= \circledast_{i=1}^n (Q_{\text{select}}^{(i)T} F^{(i)}) \boldsymbol{1} \label{eq: CP form 2},
	\end{align}
	where Equation \eqref{eq: CP form 1} is obtained by \emph{vectorization property} of CP-form (Appendix \ref{sec:lemma} Definition \ref{def:CPD}) and Equation \eqref{eq: CP form 2} can be derived from Appendix \ref{sec:lemma} Lemma \ref{l2}. Since each $Q_{\text{select}}^{(i)T} F^{(i)}$ takes $O(krI_i)$ (recall $r$ is the rank of the $\mathcal{Y}^0$ in CP-form), the time complexity of the compression step becomes only $O(kr \sum_{i=1}^n I_i)$.
\end{itemize}

\noindent \textbf{\emph{Expansion (Prediction) step:}} (line 18-21 in Algorithm \ref{alg:lowrankLP}) \\
After obtaining the $k$-D vector $\boldsymbol{v}$ which is then multiplied by the diagonal matrix $M$ to obtain another $k$-D vector $\hat{\boldsymbol{v}}$, the second step is to compute  
\begin{align}
vec(\widehat{\mathcal{Y}}^*) = (1-\alpha)((\odot_{i=1}^n Q_{\text{select}}^{(i)} )\hat{\boldsymbol{v}} +vec(\mathcal{Y}^0)). \label{expansion}
\end{align}
The left term of \eqref{expansion} has the same 
form as the vectorized CP decomposition with factor matrices $Q_{\text{select}}^{(i)} \in \mathbb{R}^{I_i \times k} $ for  $i=1,\dots,n$ (Appendix \ref{sec:lemma} Definition \ref{def:CPD}). Thus, the tensorized form can be obtained as 
\begin{align}
\widehat{\mathcal{Y}}^* = (1-\alpha)(\llbracket \boldsymbol{\hat{v}'}; Q_{\text{select}}^{(n)}, Q_{\text{select}}^{(n-1)}, \dots, Q_{\text{select}}^{(1)} \rrbracket + \mathcal{Y}^0), \label{com_rep}
\end{align}
%\vspace{-5pt}
where $\boldsymbol{\hat{v}'}$ is a reversal of the elements in $\boldsymbol{\hat{v}}$. According to Equation \eqref{com_rep}, the $k$-D vector  $\hat{\boldsymbol{v}}$ and matrices \{$Q_{\text{select}}^{(i)} :  i=1, \dots, n$\} together with $\mathcal{Y}^0$ store all the information for computing any entry of $\mathcal{\widehat{Y}}^*$ with a time complexity $O(nk)$.
Suppose the query set $\Omega$ (denoting either $\Omega_h$ or $\Omega_g$ defined in Section \ref{sec:problem}) has cardinality $|\Omega|$, the total time complexity for predicting the queried  $n$-way relations in a sparse tensor $\mathcal{O}$ is $O(nk|\Omega|)$.

\begin{table*}[ht]
	\centering
	\resizebox{\textwidth}{!}{\begin{tabular}{|l|l|l|}
\hline
                          & Compression step  & Prediction step\\ \hline
LowrankTLP (Task 1)        &           $O(\sum_{i=1}^n(I^3_i + kI_i \log (kI_i))+ |\mathcal{Y}^0| nk)$        &     $O(nk|\Omega|)$            \\ \hline
LowrankTLP (Task 2)            &         $O(\sum_{i=1}^n(I^3_i + kI_i \log (kI_i)+ krI_i))$           &        $O(nk|\Omega|)$          \\ \hline
ApproxLink (Task 1)        &    $O(\sum_{i=1}^n(I_i k_i^2 + k_i^3)+ |\mathcal{Y}^0| n (\prod_{i=1}^n k_i))$               &       $O(n (\prod_i k_i) |\Omega|)$           \\ \hline
GraphCP/GraphCP-W (Task 1) &            $O(iters * n(|\mathcal{Y}^0|nr + r^2 \sum_{i=1}^n  I_i + r\sum_{i=1}^n  I_i^2))$       &   $O(nr|\Omega|)$               \\ \hline
GraphCP (Task 2)     &    $O(iters * n( r^2 \sum_{i=1}^n  I_i + r\sum_{i=1}^n  I_i^2))$               &  $O(nr|\Omega|)$                \\ \hline
\end{tabular}}
	\caption{\textbf{Comparison of time complexity.} The time complexities of LowrankTLP, ApproxLink \cite{raymond2010fast} and GraphCP/GraphCP-W \cite{Narita2011tensor} are shown. Each method is also annotated by the applicability to hyperlink prediction (\textit{task 1}) or multiple graph alignment (\textit{task 2}).\label{Time_table}}
	\vskip -0.3in
\end{table*}

\subsection{Time and space complexity} \label{sec: complexity}
%\begin{itemize}
%  \item \textbf{\emph{Time complexity:}}
The time complexity of  selecting the optimal eigen-pairs using Algorithm \ref{alg:selecteig} is  $O(\sum_{i=1}^n I_i^3 + kI_i \log(kI_i))$. Therefore, the overall complexity of the computing the compressed representation in Algorithm \ref{alg:lowrankLP} (line 1 - 17) is $O(\sum_{i=1}^n(I^3_i + kI_i \log (kI_i))+ |\mathcal{Y}^0| nk)$ for sparse initialization, with $|\mathcal{Y}^0|$ denoting the number of nonzeros in $\mathcal{Y}^0$, and $O(\sum_{i=1}^n(I^3_i + kI_i \log (kI_i)+ krI_i))$ for CP-form initialization. The time complexity of the prediction step in Algorithm \ref{alg:lowrankLP} (line 18 - 21) is $O(nk|\Omega|)$ for both sparse initialization and CP-form initialization. Table \ref{Time_table} compares the time complexity of LowrankTLP with other existing methods described in Sections \ref{sec: related} and \ref{sec: baselines}. Note that compared with LowrankTLP, ApproxLink \cite{raymond2010fast} has slightly lower compression complexity when $n$ is small; however, when $n$ is big the term $\prod_{i=1}^n k_i$ in ApproxLink becomes a bottleneck in the computation. For GraphCP and GraphCP-W \cite{Narita2011tensor}, we assumed the first order method is applied to minimize the objective functions. Note that the overall complexity of GraphCP and GraphCP-W is the number of iterations multiplies the  per-iteration-complexity (computing the gradient) in the compression step while the empirical runtime complexity relies on the optimization method, line search type, initialization, stopping condition, etc.
%  \item \textbf{\emph{Space complexity:}}
 
The space required to store the eigenvectors of all the normalized graphs is $O(\sum_{i=1}^n I_i^2)$; to store the indexes of the selected eigen-pairs is  $O(k)$; and to store the initial tensor is $O(\boldsymbol{|}\mathcal{Y}^0\boldsymbol{|})$ and $O(\sum_{i=1}^n I_i r)$ for sparse and CP-form initial tensor respectively. Thus, the overall space complexity is $O(\boldsymbol{|}\mathcal{Y}^0\boldsymbol{|} +\sum_{i=1}^n I_i^2+ k)$ for sparse initialization and $O(\sum_{i=1}^n I_i^2 + \sum_{i=1}^n I_i r + k)$ for CP-form initialization.
%\end{itemize}

% \begin{table}[t]
% 	\centering
% 	\resizebox{\columnwidth}{!}{\begin{tabular}{|l|l|l|}
% \hline
%                           & Compression (decomposition for CP) step  & Prediction step\\ \hline
% LowrankTLP (Task 1)        &           $O(\sum_{i=1}^n(I^3_i + kI_i \log (kI_i))+ |\mathcal{Y}^0| nk)$        &     $O(nk|\Omega|)$            \\ \hline
% LowrankTLP (Task 2)            &         $O(\sum_{i=1}^n(I^3_i + kI_i \log (kI_i)+ krI_i))$           &        $O(nk|\Omega|)$          \\ \hline
% ApproxLink (Task 1)        &    $O(\sum_{i=1}^n(I_i k_i^2 + k_i^3)+ |\mathcal{Y}^0| n (\prod_{i=1}^n k_i))$               &       $O(n (\prod_i k_i) |\Omega|)$           \\ \hline
% GraphCP/GraphCP-W (Task 1) &            $O(iters * n(|\mathcal{Y}^0|nr + r^2 \sum_{i=1}^n  I_i + r\sum_{i=1}^n  I_i^2))$       &   $O(nr|\Omega|)$               \\ \hline
% GraphCP (Task 2)     &    $O(iters * n( r^2 \sum_{i=1}^n  I_i + r\sum_{i=1}^n  I_i^2))$               &  $O(nr|\Omega|)$                \\ \hline
% \end{tabular}}
% 	\caption{\textbf{Comparison of time complexity.} The time complexities of LowrankTLP, ApproxLink \cite{raymond2010fast} and GraphCP/GraphCP-W \cite{Narita2011tensor} are shown. Each method is also annotated by the applicability to hyperlink prediction (\textit{task 1}) or multiple graph alignment (\textit{task 2}).\label{Time_table}}
% \end{table}

\subsection{Error analysis of LowrankTLP algorithm}\label{sec:error}
In this section,  we first present an estimation error bound of LowrankTLP for multiple graph alignment, assuming the initial tensor $\mathcal{Y}^0$ is fully observed with Gaussian noise. The analysis provides a theoretical justification of the proposed optimization framework in Equation \eqref{eq: optimization} in Section \ref{sec:optimization}. Next, we use the \emph{transductive Rademacher complexity} \cite{el2009transductive} to derive a data-dependent error bound of LowrankTLP for binary hyperlink prediction. 
\subsubsection{Estimation error bound of TPG-structured tensor recovery from noisy observation} \label{sec: task1 bound}
Denote the transformation matrix $(1-\alpha)(I-\alpha S)^{-1}$ as $P$. We assume that the noisy tensor $\mathcal{Y}^0$ approximated by the pair-wise relations as described in Section \ref{sec: regTPG}, is generated with the true TPG-structured tensor  $\mathcal{Y}^{true}$ and a noise tensor $\mathcal{Z} \in \mathbb{R}^{I_n \times I_{n-1} \times \dots \times I_1}$ as 
$$vec(\mathcal{Y}^0) = P^{-1} vec(\mathcal{Y}^{true})  + vec(\mathcal{Z}),$$
where the entries of $\mathcal{Z} $ are  drawn from the i.i.d Gaussian distribution $\mathcal{N}(0,\sigma^2)$. 

\begin{thm} \label{the:task2} Let $\hat{P} = (1-\alpha)(I-\alpha S_k)^{-1}$, where $S_k$ is defined  in Section \ref{sec:optimization} with $eig(S_k)$ selected from $eig(S)$ by Algorithm \ref{alg:selecteig}. The inferred tensor $\hat{\mathcal{Y}}^*$ found by the LowrankTLP algorithm as $vec(\hat{\mathcal{Y}}^*) = \hat{P} vec(\mathcal{Y}^0)$ in Equation \eqref{eq:approx_sol}, has the following bounded recovery error to the true tensor $\hat{\mathcal{Y}}$
\begin{align} 
\mathbb{E}_{\mathcal{Z}} [||\hat{\mathcal{Y}}^* -  \mathcal{Y}^{true}||_{\mathcal{F}}]  \leq & (1-\alpha)(\frac{\alpha |\lambda^*|}{1-\alpha\lambda^*} ||\mathcal{Y}^0||_\mathcal{F} \nonumber \\+   &\sigma \sqrt{ \sum_{i=1}^N \frac{1}{(1-\alpha \lambda_i)^2}}), \label{ineq: the: task2}
\end{align}
where $\lambda_i$s are the eigenvalues of $S$, $\lambda^*$ is defined in Section \ref{sec:Select eig} Equation \eqref{eq:purtspectral}, and $||.||_\mathcal{F}$ denotes the Frobenius norm of a tensor.
\end{thm}
\begin{proof} 
\begin{align} 
&\mathbb{E}_{\mathcal{Z}} [||\hat{\mathcal{Y}}^* - \mathcal{Y}^{true}||_\mathcal{F}]  
= \mathbb{E}_{\mathcal{Z}} [|| (\hat{P}  - P) vec(\mathcal{Y}^0) +P vec(\mathcal{Z})||_2] \nonumber \\
&\leq|| (\hat{P}  - P) vec(\mathcal{Y}^0) ||_2 + \mathbb{E}_{\mathcal{Z}} [||P vec(\mathcal{Z})||_2] \label{ineq: Minkowski}\\
& \leq|| (\hat{P}  - P) vec(\mathcal{Y}^0) ||_2 + \sqrt{tr(\mathbb{E}_{\mathcal{Z}} [ vec(\mathcal{Z})vec(\mathcal{Z})^T]P^TP)} \label{ineq: Jensen}\\
& =|| (\hat{P}  - P) vec(\mathcal{Y}^0) ||_2 + \sigma ||P||_F \nonumber \\
& \leq||\hat{P} - P||_2|| vec(\mathcal{Y}^0)||_2  +   \sigma ||P||_F \nonumber \\
& =  (1-\alpha)(\frac{\alpha |\lambda^*|}{1-\alpha\lambda^*} ||\mathcal{Y}^0||_\mathcal{F} +   \sigma \sqrt{ \sum_{i=1}^N \frac{1}{(1-\alpha \lambda_i)^2}}), \nonumber
\end{align}
where inequalities \eqref{ineq: Minkowski} and \eqref{ineq: Jensen} are obtained with Minkowski’s and Jensen’s inequalities  respectively (End of Proof).
\end{proof} 

It is important to note that since all $\lambda_i \in [-1,1]$ are constants, the second term on the right of inequality \eqref{ineq: the: task2} is upper bounded by $O(\sqrt{N})$; the upper bound of the expected estimation error in Inequality \eqref{ineq: the: task2} can be minimized by properly choosing $\lambda^*$ to minimize $\frac{\alpha |\lambda^*|}{1-\alpha\lambda^*}$, which is the same as minimizing the optimization objective in \eqref{eq: optimization} in Section \ref{sec:optimization} by Theorem \ref{the: perturbation}. Thus, Theorem \ref{the:task2} also provides a theoretical justification of the proposed optimization formulation.

As discussed so far, we proposed to approximate the transformation matrix $P$ through properly selecting $k$ eigen-pairs of $S$ to minimize the estimation error bound in Theorem \ref{the:task2}. Note that instead of using our approximation, another natural alternative is to directly find the best rank-$k$ approximation to $P$. Proposition \ref{prop:error} below shows that our approximation strategy is a better solution (the proof is given in Appendix \ref{proof 2}). 
\begin{prop} \label{prop:error} Follow the definitions of $A$ and $\hat{A}$ in Proposition \ref{the:lrrep}, where $eig(S_k)$ is selected from $eig(S)$ by Algorithm \ref{alg:selecteig}. Define $A_k$ as the best rank-$k$ approximation to $A$ in both spectral and Frobenius norm.  Assuming $k < \prod_i I_i$, we have the following inequalities
	\begin{align}
	||\hat{A}-A||_2 < ||A_k-A||_2 \ \text{and} \ ||\hat{A}-A||_F < ||A_k-A||_F. \nonumber
	\end{align} 
\end{prop}
% \begin{proof} 
% Let $\sigma_1 > \sigma_2 > \dots > \sigma_N$ be the sorted eigenvalues of matrix $S$.
% Since $ \sigma_i \in [-1,1], \ \text{for} \  i=1,\dots, N$ and $\alpha \in (0,1)$, by Eckart-Young-Mirsky theorem, the nonzero eigenvalues of $A_k$ are $\{\frac{1}{1-\alpha \sigma_i} :  i=1, \dots, k \}$ and the perturbations are given below.
% \begin{align*}
% ||A_k  - A||_2 & = \frac{1}{1-\alpha\sigma_{k+1}}  \\
% ||A_k  - A||_F & = \sqrt{\sum_{i=k+1}^N (\frac{1}{1-\alpha\sigma_i})^2}
% \end{align*}
% Using $\{\sigma_i : i=1,\dots, k\}$ as eigenvalues and their corresponding eigenvectors of $S$ to construct a rank-$k$ matrix $L$, and define $B= (I-\alpha L)^{-1}$, we have $||\hat{A}-A||_2 \leq ||B -A||_2$ and $||\hat{A} - A||_F \leq ||B -A||_F$ according to the definition of $S_k$ in Section \ref{sec:optimization}. It is also not hard to show that $||B -A||_2 < ||A_k -A||_2$ and $||B -A||_F < ||A_k -A||_F$ by the facts that every $ \sigma_i \in [-1,1]$ and $\alpha \in (0,1)$ (see details of the proof in Appendix \ref{proof 2}). Thus, inequalities \eqref{ie:fro} hold. (End of Proof)
% \end{proof} 

\subsubsection{Transductive Rademacher bound for binary hyperlink prediction}
\label{sec: task2 bound}
Define $\Theta = \Theta_h \cup \bar{\Theta}_h = \{(i_1, i_2, \dots, i_n)  :  \forall  i_j \in [1,I_j], j = 1,\dots, n \}$ as the set of all $n$-way relations among the vertices across the $n$ knowledge graphs, where $\bar{\Theta}_h$ denotes the complement of $\Theta_h$. Define tensor $\mathcal{Y}^{true} \in \{+1,-1\}^{I_n\times I_{n-1} \times ... \times I_1}$ which stores the true labels of all the hyperlinks in set $\Theta$, where the label of the $(i_1, i_2, \dots, i_n)$-th hyperlink is either 1 (the link exists) or -1 (the link does not exist). Accordingly, $\mathcal{Y}^0$ contains a subset of known entries (hyperlinks) sampled from $\mathcal{Y}^{true}$ and zeros for the other unknown entries. Define $\mathcal{Y}_{out} \subset \mathbb{R}^{I_n\times I_{n-1} \times ... \times I_1}$ as the set of tensors outputted by the LowrankTLP algorithm over all possible $\Theta_h$  / $\bar{\Theta}_h$ partitions such that for every $\widehat{\mathcal{Y}}^* \in \mathcal{Y}_{out}$ we have $vec(\hat{\mathcal{Y}}^*) = \hat{P} vec(\mathcal{Y}^0)$  (as in Theorem \ref{the:task2}). In the following derivations, we assume $\mathcal{Y}^0$ is normalized by $||\mathcal{Y}^0||_{\mathcal{F}}$ so that its Frobenius norm is unit. This normalization is proper since it does not change the signs in $\mathcal{Y}^*$. In Theorem \ref{the: trans}, we provide a data-dependent error bound of LowrankTLP for binary hyperlink prediction, using the \emph{transductive Rademacher complexity} proposed in \cite{el2009transductive}.

\begin{thm}
\label{the: trans}
Denote $l = |\Theta_h |$ and $u = |\bar{\Theta}_h|$ to be the cardinalities of $\Theta_h$ and $\bar{\Theta}_h$ respectively.  Let $c_0 = \sqrt{\frac{32\ln{(4e)}}{3}}$, $Q = \frac{1}{l}  + \frac{1}{u}$ and $G = \frac{l+u}{(l+u-0.5)(1-0.5/\max{(l,u)})}$. For any fixed positive real $\gamma$, with probability of at least $1-\delta$ over the random choice of the set $\Theta_h$, for all $ \widehat{\mathcal{Y}}^* \in \mathcal{Y}_{out}$,
\begin{align}
\mathcal{L}^{\gamma}_{u}(\widehat{\mathcal{Y}}^*)\leq & \widehat{\mathcal{L}}_{l}^{\gamma}(\widehat{\mathcal{Y}}^*) + \frac{||\hat{P}||_F}{\gamma}\sqrt{\frac{2}{lu}}  + c_0 Q \sqrt{\min{(l,u)}} \nonumber\\ & +\sqrt{\frac{GQ}{2}\ln(\frac{1}{\delta})}, \label{eq: transbound}
\end{align}
where $\mathcal{L}^{\gamma}_{u}(\widehat{\mathcal{Y}}^*)$ and $\widehat{\mathcal{L}}_{l}^{\gamma}(\widehat{\mathcal{Y}}^*)$ are the $\gamma$-margin test and empirical errors respectively defined as
\begin{align}
\mathcal{L}^{\gamma}_{u}(\widehat{\mathcal{Y}}^*) & = \frac{1}{u} \sum_{(i_1,i_2,\dots,i_n) \in\bar{\Theta}_h} \ell_{\gamma} (\widehat{\mathcal{Y}}^*_{i_n,i_{n-1},\dots,i_1},\mathcal{Y}^{true}_{i_n,i_{n-1},\dots,i_1}) \nonumber \\
\widehat{\mathcal{L}}_{l}^{\gamma}(\widehat{\mathcal{Y}}^*) & = \frac{1}{l} \sum_{(i_1,i_2,\dots,i_n) \in \Theta_h} \ell_{\gamma} (\widehat{\mathcal{Y}}^*_{i_n,i_{n-1},\dots,i_1},\mathcal{Y}^{true}_{i_n,i_{n-1},\dots,i_1}), \nonumber
\end{align}
with $\ell_{\gamma} (a,b) = 0$ if $ab > \gamma$ and $\ell_{\gamma} (a,b) = \min{(1,1-\frac{ab}{\gamma})}$ otherwise.
\end{thm}

\begin{proof}
The bound \eqref{eq: transbound} in Theorem \ref{the: trans} is based on the transductive Rademacher bound using the \emph{transductive Rademacher complexity} \cite{el2009transductive} given in Appendix \ref{sec: appendix_TRC} Definition \ref{def: appendix_TRC}. According to Appendix \ref{sec: appendix_TRC} Theorem \ref{the: appendix_TRC}, we only need to bound the Rademacher complexity $R_{l+u}(\mathcal{Y}_{out})$ as below:
\begin{align}
R_{l+u}(\mathcal{Y}_{out}) & = (\frac{1}{l}  +\frac{1}{u}) \mathbb{E}_{\boldsymbol{\sigma}} \Big[\sup_{\widehat{\mathcal{Y}}^* \in \mathcal{Y}_{out}}  \boldsymbol{\sigma}^T  vec(\widehat{\mathcal{Y}}^*) \Big] \nonumber \\
& \leq (\frac{1}{l}  +\frac{1}{u}) \mathbb{E}_{\boldsymbol{\sigma}} \Big[\sup_{\mathcal{Y}^0: ||\mathcal{Y}^0||_\mathcal{F} = 1}  \boldsymbol{\sigma}^T  \hat{P}vec(\mathcal{Y}^0) \Big] \nonumber \\
& = (\frac{1}{l}  +\frac{1}{u})  \mathbb{E}_{\boldsymbol{\sigma}} \Big[ ||\hat{P}\boldsymbol{\sigma}||_2\Big] \label{eq: thans_2}\\
& \leq (\frac{1}{l}  +\frac{1}{u})  \sqrt{  tr(\mathbb{E}_{\boldsymbol{\sigma}} \Big[\boldsymbol{\sigma}\boldsymbol{\sigma}^T \Big] \hat{P}^T \hat{P} )} \label{eq: thans_3}\\
& = ||\hat{P}||_F\sqrt{\frac{2}{lu}}, \nonumber
\end{align}
where \eqref{eq: thans_2} and \eqref{eq: thans_3} are obtained using  Cauchy-Schwarz and Jensen’s inequalities  respectively. (End of Proof)
\end{proof}
Given the eigenvalues of $\hat{P}$ are bounded within $[\frac{1-\alpha}{1+\alpha} , 1]$,  it is clear that $ ||\hat{P}||_F \leq \sqrt{l+u}$. Assuming $l+u \rightarrow \infty$ and $l \ll u$, the error bound \eqref{eq: transbound} can be simplified as $\mathcal{L}^{\gamma}_{u}(\widehat{\mathcal{Y}}^*)\leq \widehat{\mathcal{L}}_{l}^{\gamma}(\widehat{\mathcal{Y}}^*) +O \Big(\sqrt{\frac{1}{l}}  \Big)$, which has a slower convergence rate compared with the estimation error bound of the convex tensor completion model \cite{tomioka2011statistical} under certain conditions. It is also important to note that when $l$ is very small i.e. the labeled $n$-way relations are extremely sparse, the term $\sqrt{\frac{1}{l}}$ increases relatively slow as $l$ decreases. Thus, the bound by $O \Big(\sqrt{\frac{1}{l}}  \Big)$ implies that empirically, the performance of LowrankTLP might deteriorate less with very sparse input tensors, which is consistent with our observations in both simulations and experiments on real datasets shown later in Section \ref{sec:exp}.
%\begin{align}
%\mathcal{L}^{\gamma}_{u}(\widehat{\mathcal{Y}}^*)\leq \widehat{\mathcal{L}}_{l}^{\gamma}(\widehat{\mathcal{Y}}^*) + O \bigg(\sqrt{\frac{1}{l}} \bigg).\label{ineq: transbound_simp}
%\end{align}
\section{Related work} \label{sec: related}
%Multi-relational learning with knowledge graphs remains a challenge due to the exponential number of possible multi-relations to evaluate even in a small number of graphs of medium sizes. 
Two categories of tensor-based techniques have been previously applied to the problem of multi-relational learning with multiple knowledge graphs.

The first category of methods also leverage the same idea of semi-supervised manifold learning on the tensor product graph (TPG) for predicting hyperlinks across the graphs. Given a set of labeled $n$-way tuples of graph vertices, these methods aim at labeling/scoring the unlabeled $n$-way tuples by learning with the manifold structure in the TPG. \cite{kashima2009link} proposed a semi-supervised \emph{link propagation} method using conjugate gradient descend to predict the hyperlinks in the multi-relational tensor. The \emph{link propagation} method is only empirically scalable to three-way tensors due to the necessity of computing the full tensor in every iteration. Alternatively, several methods were proposed to apply low-rank approximation on each individual graph rather than working with the original TPG for better scalability to the tensor product of two or three large graphs. For example, \emph{approximate link propagation} \cite{raymond2010fast} applies label propagation on the product of two low-rank knowledge graphs for pairwise link prediction, and \cite{dunlavy2011temporal} makes a similar low-rank assumption on a single bipartite graph for pairwise link prediction.
 %even though it was only used for predicting the links on a pair of graphs in the experiments. %The algorithm takes two graphs $W^{(1)} \in \mathbb{R}^{I_2 \times I_2}$ and $W^{(2)} \in {I_2 \times I_2}$ and a matrix $Y^{(0)} \in  \mathbb{R}^{I_1 \times I_2} $ stores the known pairwise link strength between the graph nodes as inputs, and output a matrix $Y$ whose elements are the pairwise link strength between the unknown nodes. 
Transductive learning over product graph (TOP) \cite{liu2016cross} is a graph-based one-class transductive learning algorithm, in which the Gaussian random fields prior proposed in \cite{zhu2003semi} is approximated as a regularization term with the product of multiple low-rank knowledge graphs to overcome the bottleneck of evaluating the prior.
%Transductive learning over product graph (TOP) \cite{liu2016cross} is a cross-graph relational learning algorithm for one-class classification, in which the product graph structure is introduced via a Gaussian random fields prior.
%taking $n$ knowledge graphs $\{W^{(i)} | i=1,\dots n\}$ and a tensor $Y^{(0)}$ containing the known binary multi-relations between the graph nodes, and output the unlabeled multi-relations by introducing the product graph structure via a Gaussian random fields prior. 
In general, these methods are not applicable to a large number of knowledge graphs: first, low-rank approximation of each individual graph does not guarantee a globally optimal approximation of the TPG as stated in Appendix \ref{sec:lemma} Lemma \ref{l5}; and second, the rank of the TPG is exponential in the number of graphs, and therefore the approximation is not scalable to many graphs. Therefore, these approximation methods do not preserve the performance of learning with the original TPG and still suffers scalability issues to learn from a large number of knowledge graphs.

The second category are tensor decomposition methods regularized by graph Laplacian, which decompose a noisy complete tensor or an incomplete sparse tensor into low-rank factor matrices to estimate the true tensor. For example, \cite{Narita2011tensor} proposed two types of graph Laplacian regularization for both CP and Tucker decomposition. The first type is called \emph{within-mode} regularization which extends the regularized matrix factorization methods to encourage the components in each factor matrix to be smooth among strongly connected vertices in its corresponding graph.
The second type is called \emph{cross-mode} regularization, which regularizes all the factor matrices jointly with the graph Laplacian of the TPG. For joint analysis of data from multiple sources including tensor, matrices  and knowledge graphs, \cite{zheng2010collaborative} introduced a coupled matrix tensor factorization (CMTF) model with \emph{within-mode} graph Laplacian regularization for collaborative filtering. Though these tensor decomposition models can be solved by any scalable first-order method based on the idea of \emph{all-at-once} optimization \cite{acar2011scalable, AcKoDu11}, they are however potentially lead to poor local minima especially for high-order tensors due to their non-convex formulations. Moreover, it has been observed that the accuracy of tensor completion tends to degrade severely when only a small fraction of multi-relations are observed \cite{Narita2011tensor}. 
% Over the past decade, convex tensor decomposition models have been extensively studied, which introduce convex relaxation on tensor multi-linear rank via employing variant types of trace norm  regularization \cite{tomioka2010estimation, gandy2011tensor, wimalawarne2018convex} on the unfolded tensor.  Although these convex tensor decomposition methods converge to global minimum and have better theoretical guarantees \cite{tomioka2011statistical} than conventional non-convex models, their scalability need to be significantly improved in order for large-scale applications.
% Note that convex tensor decomposition models which introduce convex relaxation on tensor multi-linear rank employing variants of trace norm regularization on the unfolded tensor \cite{tomioka2010estimation, tomioka2013convex, gandy2011tensor, wimalawarne2014multitask, wimalawarne2018convex}, converge to globally optimal solution and have theoretical guarantees of recovering the true tensor \cite{tomioka2011statistical}. These models, however, are generally not applicable to large-scale real-world problems due to the scalability issue of involving the trace-norm regularization.

%  Although these convex tensor decomposition methods converge to global minimum and have better theoretical guarantees \cite{tomioka2011statistical} than conventional non-convex models, these methods are generally not applicable  for large-scale high-order problems due to the scalability issue. }
% \vspace{-0.1in}
\begin{figure*}[ht]
	\begin{center}
		\begin{tabular}{cc}
			\includegraphics[width=0.5\textwidth]{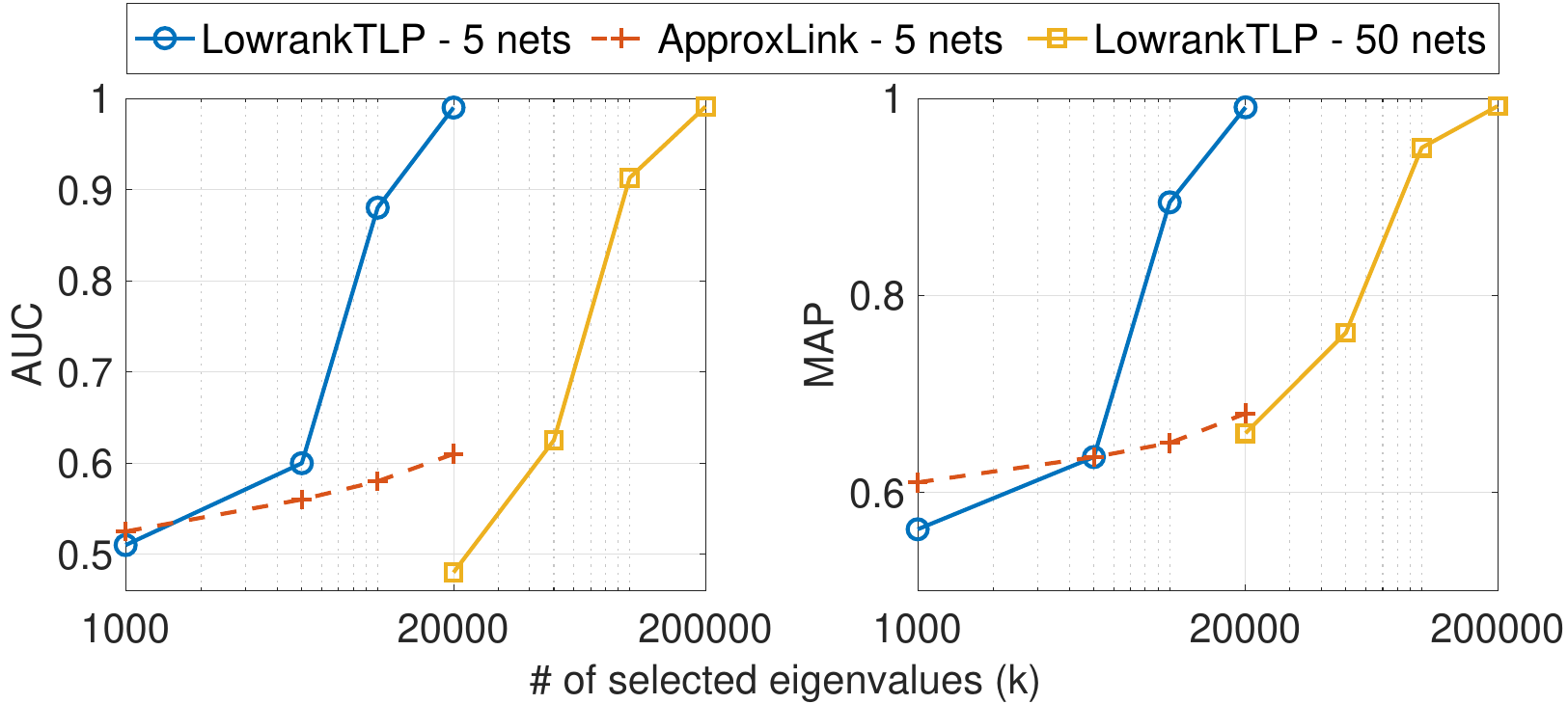} &
			\includegraphics[width=0.5\textwidth]{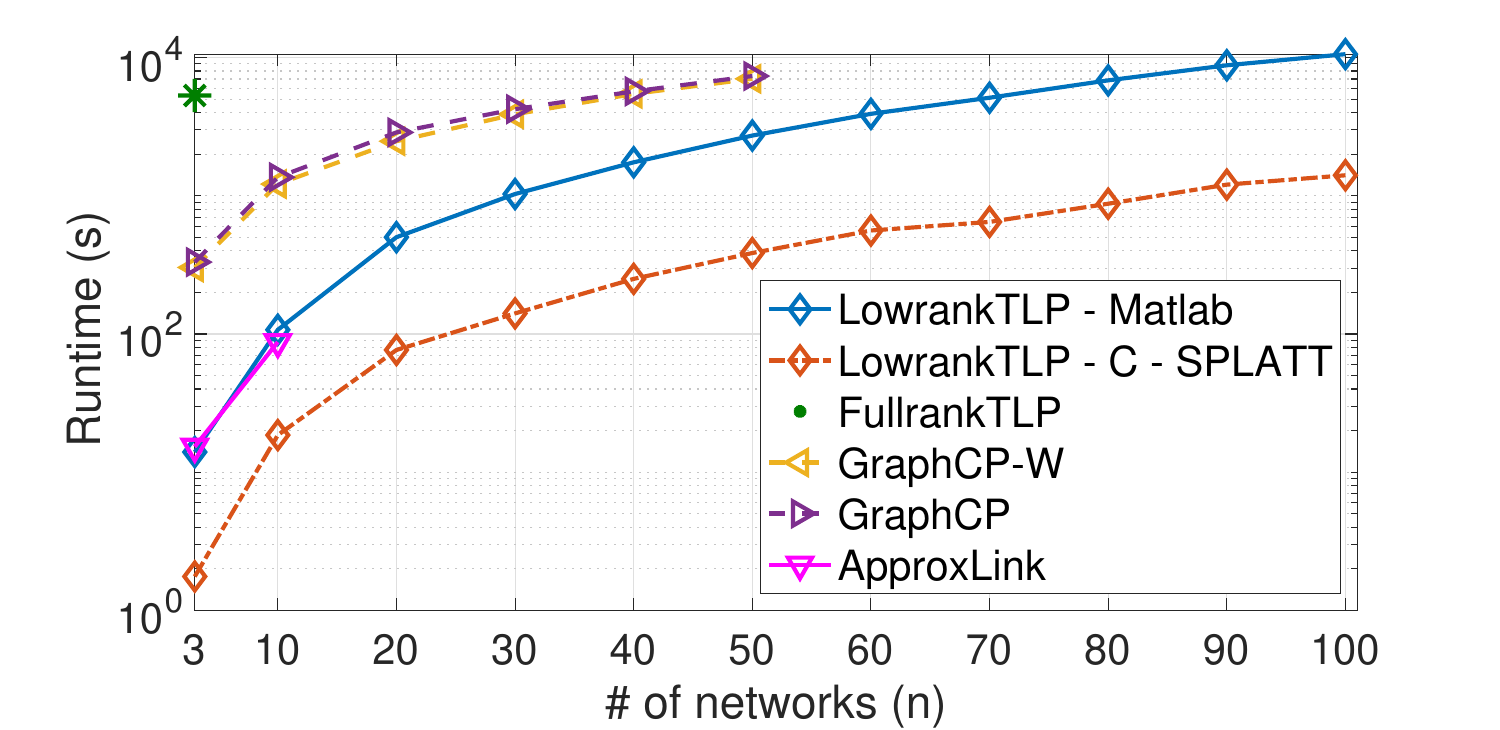} \\
			(A)  & (B)  \\
		\end{tabular}
		\vskip -0.1in
		\caption{\textbf{Simulation results.} (A) Effectiveness comparisons by varying TPG ranks. 
			%Blue solid line represents our LowrankTLP method; and red solid line represents the approximate Link Propagation method. Figure 
			(B) Efficiency and scalability comparisons. The curves are truncated if the method is not scalable to the larger sizes.}
		\label{Simulation_figures}
	\end{center}
	\vskip -0.2in
\end{figure*}
\section{Experiments} \label{sec:exp}
%In the experiments, the performance of LowrankTLP for hyperlink prediction (Task 1) and graph alignment (Task 2) was evaluated in simulation and three real datasets. LowrankTLP was first applied in simulation and DBLP data of scientific publication records for semi-supervised learning; then applied on segmented CT Scan images and protein-protein interaction (PPI) networks data for unsupervised learning. Four baselines: \textbf{ApproxLink} \cite{raymond2010fast}, \textbf{TOP} \cite{liu2016cross}, \textbf{TD} (tensor decomposition) and \textbf{GraphTD} \cite{Narita2011tensor}  were also included in the experiments for comparison. For global alignment of multiple PPI networks LowrankTLP was compared with \textbf{IsoRankN} \cite{liao2009isorankn} and \textbf{BEAMS} \cite{Alkan2013beams} which were developed specifically for PPI network alignment. 
In the experiments, the performance of LowrankTLP for hyperlink prediction (Task 1) and multiple graph alignment (Task 2) was evaluated in simulation and three real datasets. In Section \ref{sec: baselines}, we first explain the baseline methods implemented for performance comparisons. In Section \ref{sec-simulations}, we evaluate the effectiveness and efficiency of LowrankTLP for hyperlink prediction (Task 1) on the simulation data, through controlling the size and topology of multiple artificial graphs. In Section \ref{sec-DBLP}, we test the practical application of LowrankTLP for hyperlink prediction (Task 1) on the DBLP dataset of scientific publication records. In Section \ref{sec-CT} and \ref{sec-ppi}, we evaluate the practical application of LowrankTLP to multiple graph alignment (Task 2). We first apply LowrankTLP to align up to 26 CT scan images in Section \ref{sec-CT}. Next, we evaluate the performance of LowrankTLP for the global alignment of up to 4 full protein-protein interaction (PPI) networks across 4 different species in Section \ref{sec-ppi}. For better clarity, we also summarize the input/output of all the experiments in Appendix \ref{sec:Supplementary} Table \ref{summary_table}.

\subsection{Baseline methods and implementations} \label{sec: baselines}
\subsubsection{Baseline methods}
We compared LowrankTLP with seven baseline methods in the simulations and the experiments based on their applicability to hyperlink prediction and multiple graph alignment.
\begin{itemize}
\item{\emph{Approximate link propagation} (\textbf{ApproxLink}) \cite{raymond2010fast}: }ApproxLink was originally  designed for pair-wise link prediction in a matrix. We extended its operations for hyperlink prediction in a tensor. Given an incomplete initial tensor $\mathcal{Y}^0 \in \mathbb{R}^{I_n\times I_{n-1} \times ... \times I_1}$ with zeros representing the missing entries,  and $n$ knowledge graphs $\{W^{(i)} :  i=1,\dots,n\}$, ApproxLink can score the queried entries in the tensor.
\item{\emph{Transductive learning over product graph} (\textbf{TOP})} \cite{liu2016cross}:  TOP is designed for one-class classification. Given a binary incomplete initial tensor $\mathcal{Y}^0 \in \mathbb{R}^{I_n\times I_{n-1} \times ... \times I_1}$ with ones and zeros representing the observed positive entries and missing entries respectively,  and $n$ knowledge graphs $\{W^{(i)} :  i=1,\dots,n\}$, TOP can detect the queried positive entries in the tensor.
\item{\emph{CANDECOMP/PARAFAC decomposition} (\textbf{CP}): } An initial tensor $\mathcal{Y}^0 \in \mathbb{R}^{I_n\times I_{n-1} \times ... \times I_1}$ which is either noisily complete or incomplete with zeros representing the missing entries is decomposed into $n$ factor matrices by solving a \emph{least square} problem. The factor matrices are then used to construct the queried entries in the tensor.
%\item\textcolor{red}{{\emph{CP using weighted optimization} (\textbf{CP-W}) \cite{acar2011scalable}: }An incomplete initial tensor $\mathcal{Y}^0 \in \mathbb{R}^{I_n\times I_{n-1} \times ... \times I_1}$ is decomposed into $n$ factor matrices, by solving a \emph{weighted least square} problem. The factor matrices are then used to construct the queried entries in the tensor}.

\item{\emph{CP using weighted optimization} (\textbf{CP-W}) \cite{acar2011scalable}: } CP-W is a variation of CP where the differences are that the initial tensor $\mathcal{Y}^0$ is incomplete, and the $n$ factor matrices are found by solving a \emph{weighted least square} problem.

\item{\emph{Graph regularized CP} (\textbf{GraphCP}) \cite{Narita2011tensor}: }Given $n$ knowledge graphs $\{W^{(i)} :  i=1,\dots,n\}$, an initial tensor $\mathcal{Y}^0 \in \mathbb{R}^{I_n\times I_{n-1} \times ... \times I_1}$ which is either noisily complete or incomplete with zeros representing the missing entries is decomposed into $n$ factor matrices  by solving a \emph{least square} problem with \emph{cross-mode} regularization using TPG. The factor matrices are then used to construct the queried entries in the tensor.

\item{\emph{Graph regularized CP using weighted optimization} (\textbf{GraphCP-W}) \cite{Narita2011tensor}: } Similarly, GraphCP-W is a variation of GraphCP where the differences are that the initial tensor $\mathcal{Y}^0$ is incomplete, and the $n$ factor matrices are again learned by solving a \emph{weighted least square} problem with \emph{cross-mode} regularization using TPG.

%\item\textcolor{red}{{\emph{Graph regularized CP using weighted optimization} (\textbf{GraphCP-W}) \cite{Narita2011tensor}: }Given $n$ knowledge graphs $\{W^{(i)} :  i=1,\dots,n\}$, an incomplete initial tensor $\mathcal{Y}^0 \in \mathbb{R}^{I_n\times I_{n-1} \times ... \times I_1}$ is decomposed into $n$ factor matrices by solving a \emph{weighted least square} problem with \emph{cross-mode} regularization using TPG. The factor matrices are then used to construct the queried entries in the tensor.}

\item{\emph{Spectral methods for multiple PPI network alignment} (\textbf{IsoRankN}) \cite{liao2009isorankn}: }Given $n$ PPI networks $\{W^{(i)} : i=1,\dots,n\}$ of $n$ species, and BLAST sequence similarities $\{R_{ij} \in \mathbb{R}_+^{I_i \times I_j} : \forall i,j \in [1,n] \ \text{and} \  i <  j\}$ between proteins from each pair of species, IsoRankN  finds a global alignment of the $n$ PPI networks based on spectral clustering on the induced graph of pairwise alignment scores.

\item{\emph{Backbone extraction and merge strategy for multiple PPI network alignment} (\textbf{BEAMS})  \cite{Alkan2013beams}: }Given $n$ PPI networks $\{W^{(i)} : i=1,\dots,n\}$ of $n$ species, and BLAST sequence similarities $\{R_{ij} \in \mathbb{R}_+^{I_i \times I_j} : \forall i,j \in [1,n] \ \text{and} \  i <  j\}$ between proteins from each pair of species, BEMAS finds a global alignment of the $n$ PPI networks by solving a combinatorial optimization problem with a heuristic approach.
\end{itemize}

%For the experiment of multiple PPI network alignment  in Section \ref{sec-ppi}, LowrankTLP was compared with \textbf{IsoRankN} \cite{liao2009isorankn} and \textbf{BEAMS} \cite{Alkan2013beams} which were developed specifically for PPI network alignment. \\
% \usepackage{multirow}
%\begin{table}[t]
%\resizebox{\columnwidth}{!}{\begin{tabular}{|c|c|c|c|c|c|c|c|c|c|}
%\hline
%\multicolumn{1}{|c|}{} & \multicolumn{1}{c|}{LowrankTLP} & \multicolumn{1}{c|}{ApproxLink} & \multicolumn{1}{c|}{TOP} & \multicolumn{1}{c|}{CP} & \multicolumn{1}{c|}{GraphCP} & \multicolumn{1}{c|}{CP-W} & \multicolumn{1}{c|}{GraphCP-W} & \multicolumn{1}{c|}{IsoRankN} & \multicolumn{1}{c|}{BEAMS} \\ \hline
%Task 1                 & \checkmark                              & \checkmark & \checkmark                       & \checkmark                       & \checkmark & \checkmark                         & \checkmark                              &                             &                          \\ \hline
%Task 2                 & \checkmark                               &                               &                         &                        & \checkmark                            &                          &                               & \checkmark                            & \checkmark                         \\ \hline
%		\end{tabular}}
%	\caption{\textbf{Correspondences between learning tasks and methods}} \label{table: task_method}
%\end{table}

\subsubsection{Implementation details}
The graph regularization hyperparameter $\alpha$ defined in section \ref{sec: LPTPG} was chosen from the pool \{0.001, 0.1, 0.9, 0.99\} for LowrankTLP,  ApproxLink and TOP;  rank $\left\lceil\sqrt[n]{k}\right\rceil$ approximation was applied to each individual graph for ApproxLink and TOP to guarantee the approximated TPG has the same or larger rank than $k$.

For better scalability, we adopted the first-order method ADAM \cite{kingma2014adam} based on the \emph{all-at-once} optimization \cite{acar2011scalable, AcKoDu11} to minimize the objective functions of CP, CP-W, GraphCP and GraphCP-W. The factor matrices were randomly initialized; the stopping criteria was chosen to be $||\nabla f(\boldsymbol{x}^t)||_2 \leq 10^{-3} ||\nabla f(\boldsymbol{x}^0)||_2$, where $\boldsymbol{x}^t$ denotes the stack of all the vectorized factor matrices in the $t$-th iteration; the maximum number of iterations was set to 1000. Note that,  for GraphCP and GraphCP-W the gradient scale of the \emph{cross-mode} regularization term increases much faster than the gradient scale of the decomposition term, as the tensor order (the number of graphs) increases. Therefore, when the tensor order is high, the graph hyperparameter $\alpha$ as defined in \cite{Narita2011tensor} tends to be set very small. Unless otherwise stated, we chose $\alpha$ from $\{10^{-5}, 10^{-4}, 10^{-3}, 10^{-2}, 10^{-1}\}$ as suggested in \cite{Narita2011tensor} and $r$ (tensor rank) from $\{10, 50, 100\}$ for all CP based methods in Task 1; for Task 2 the CP rank $r$ is equal to the rank of the initial CP-form tensor and is chosen by PCA to cover at least 90\% of the spectral energy in the stacking matrix $R$ defined in Section \ref{sec: regTPG}.

The baselines IsoRankN and BEAMS were developed specifically for PPI network alignment. We downloaded and ran the original packages$^[$\footnote{IsoRankN package: \url{http://cb.csail.mit.edu/cb/mna/}.}$^{,}$\footnote{BEAMS package: \url{http://webprs.khas.edu.tr/~cesim/BEAMS.tar.gz}.}$^]$ to obtain the alignment scores with the graph hyperparameter $\alpha$ selected from $\{0.1, 0.3,0.5, 0.7, 0.9 \}$ as suggested in their packages.

All the experiments were performed using our server with Intel(R) Xeon(R) CPU E5-2450 with 32 cores (2.10GHz) in 2 CPUs and 196GB of RAM. All the baseline methods except IsoRankN and BEAMS were implemented using MATLAB R2018b. 
\begin{table}[t]
	\centering
	\resizebox{\columnwidth}{!}{
\begin{tabular}{|l|l|l|l|l|}
\hline
\multirow{2}{*}{} & \multicolumn{2}{l|}{5 nets} & \multicolumn{2}{l|}{10 nets} \\ \cline{2-5} 
                  & AUC          & MAP          & AUC           & MAP          \\ \hline
LowrankTLP        & 0.990        & 0.991        & 0.942         & 0.952        \\ \hline
ApproxLink        & 0.610        & 0.679        & 0.554         & 0.672        \\ \hline
GraphCP           & 0.540        & 0.539        & 0.536         & 0.533        \\ \hline
GraphCP-W         & 0.543        & 0.535        & 0.533         & 0.550        \\ \hline
CP                & 0.549        & 0.528        & 0.520         & 0.527        \\ \hline
CP-W              & 0.522        & 0.486        & 0.472         & 0.496        \\ \hline
\end{tabular}
}
\caption{\textbf{Effectiveness comparison in simulations.}}\label{simluation_table}
\vskip -0.3in
\end{table}

\subsection{Simulations} \label{sec-simulations}
Synthetic graphs were generated to evaluate the performance and the scalability. We started with a graph of density $0.1$ and size $I$ to generate $n$ distinct graphs by randomly permuting $10\%$ of edges from the common ``ancestor'' graph so that they share similar structures that can be utilized for matching the multi-relations. The inputs are the $n$ graphs and a sparse $n$-way tensor $\mathcal{Y}^0 \in \mathbb{R}^{I \times ... \times I}$ with $I/2$ (half) of its diagonal entries set to 1s. We set the other $I/2$ diagonal entries and $I/2$ randomly sampled off-diagonal entries to 0.9 and treated them as positive and negative test samples, respectively.  The outputs are scores of the $I$ test entries after label propagation, which can be used to distinguish the positive and negative classes based on the assumption that the vertices indexed by the diagonal entries of the tensor should have high similarities since they come from the same ``ancestor'' graph and this information should be captured by the TPG. 
\begin{figure*}[t]
	%%\vskip 0.2in
	\begin{center}
		\begin{tabular}{cr}
			\includegraphics[width=0.33\textwidth,clip,trim=0 -10 0 0]{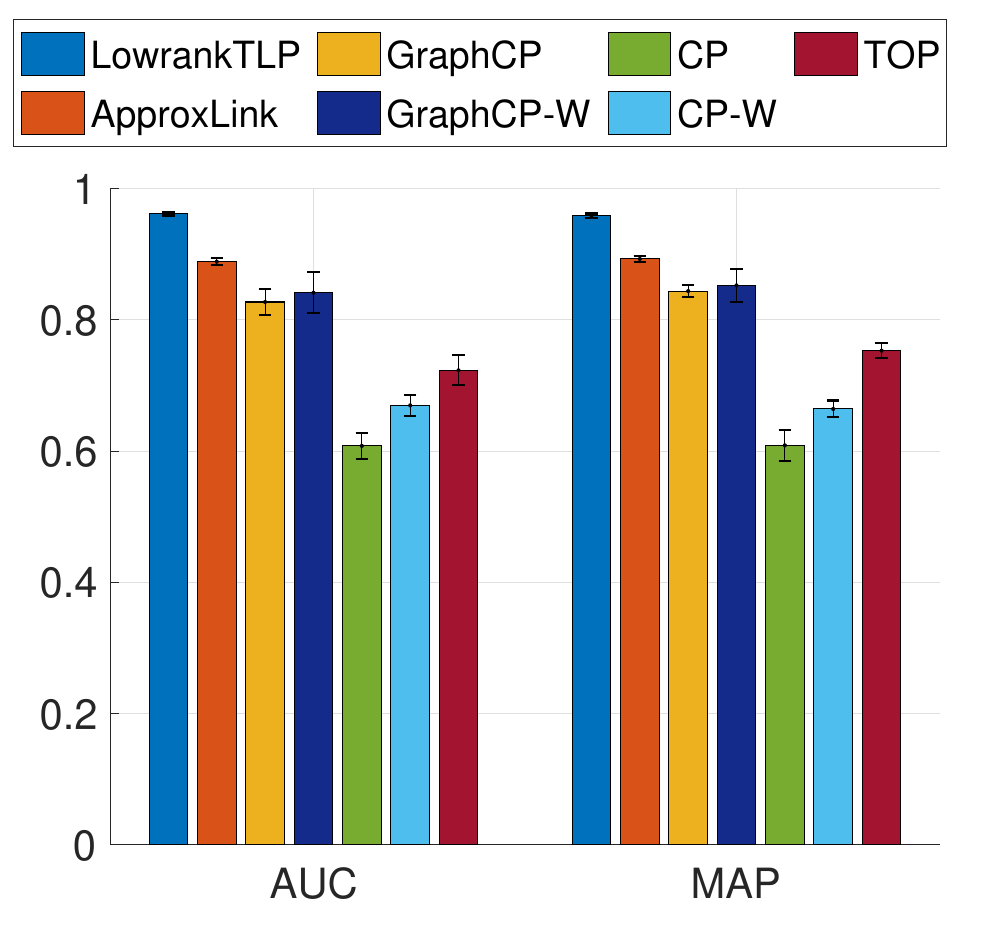} & 
			\includegraphics[width=0.66\textwidth,clip,trim=30 0 0 0]{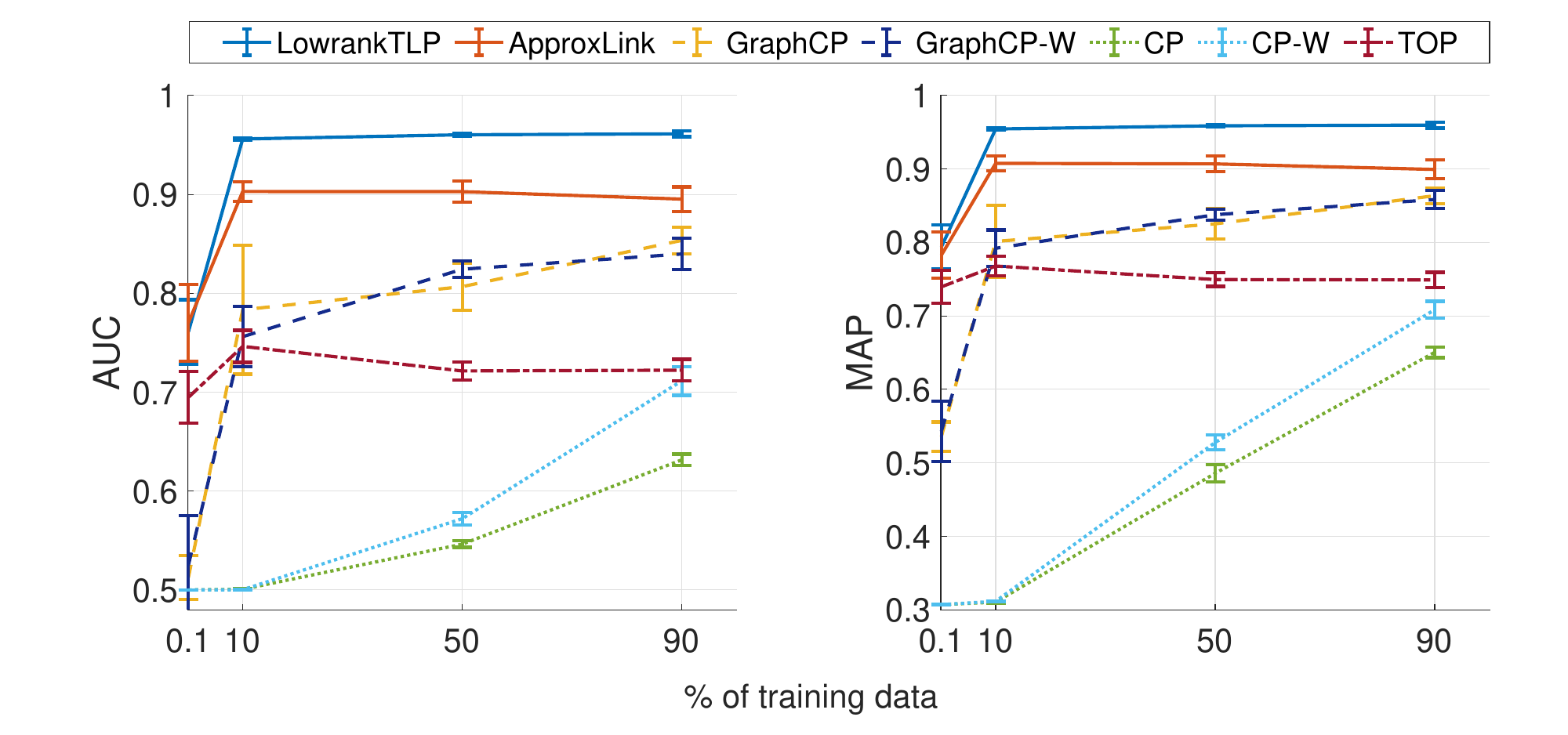}  \\
			(A) & (B) \hspace{152px} (C) \hspace{68px} \\
		\end{tabular}
		\caption{\textbf{DBLP results.} (A) The performance of 5-fold cross-validation. The average and standard deviation across the 5 folds are shown. (B) \& (C) The performance of using various percentages of training data. The average and standard deviation are shown for each percentage across different random samplings. }
		\label{DBLP_figures}
	\end{center}
	\vspace{-0.3in}
\end{figure*}
\begin{itemize}
\item{\textbf{\emph{Effectiveness}}}: We  compare LowrankTLP with ApproxLink, CP, CP-W, GraphCP and GraphCP-W using the same sparse tensor $\mathcal{Y}^0$ as input.  For fair comparisons, we fixed  $\alpha=0.1$ for  LowrankTLP and used the best hyperparameters for all the baseline methods.  The area under the curve (AUC) and mean average precision (MAP) are the evaluation metrics. Each experiment was repeated five times and the average performances are reported. Table \ref{simluation_table} shows that LowrankTLP clearly outperforms all the baselines to learn multi-relations among 5 and 10 graphs. The prediction of the CP-based methods is almost random, which is not surprising given the fact that tensor $\mathcal{Y}^0$ is extremely sparse; this observation also agrees with the previous observations that the accuracy of tensor decomposition degrades severely when only a small fraction of entries is observed \cite{Narita2011tensor, tomioka2010estimation}.  
Note that ApproxLink performs better than the CP-based methods, which implies that label propagation is a more robust approach for sparse inputs than tensor decomposition for hyperlink prediction.  Figure \ref{Simulation_figures}(A) shows that the LowrankTLP outperforms ApproxLink in different TPG ranks, which validates the advantage of our optimization formulation \eqref{eq: optimization}.  It also shows that LowrankTLP requires only a moderate rank $k \geq 10,000$ when $n=5$, and achieves a high performance with $k\geq$ 100,000 when $n=50$, whereas ApproxLink is not applicable to such a large number of graphs.

\item{\textbf{\emph{Efficiency and scalability}}}: We further compared the running time of the MATLAB implementation of  LowrankTLP using Tensor Toolbox \cite{TTB_Software} version 2.6 and the parallel implementation using SPLATT library \cite{splattsoftware} (described in Section \ref{sec:lrTLP}) with the baseline methods applicable to the knowledge graphs. We chose a small tensor rank $r=10$ for GraphCP and GraphCP-W. The TPG rank $k=\frac{10^4}{5}n$ was chosen for LowrankTLP which achieves AUC $\approx 0.9$ empirically.
In Figure \ref{Simulation_figures}(B), we observe that the parallel LowrankTLP results in a speedup of about one order of magnitude compared to the MATLAB version. The parallel implementation of LowrankTLP improved the running time to $10^3$s compared with $10^4$s by the sequential implementation to align 100 graphs of size 1000 each. ApproxLink has a similar running time as sequential LowrankTLP on 3 and 10 graphs, while it is not applicable for more graphs due to the exponential growth of the number of components as discussed in Section \ref{sec: complexity}. The empirical running time of GraphCP/GraphCP-W are worse than LowrankTLP even if the theoretical time complexity for computing the gradients in each iteration is fast as analyzed in Table \ref{Time_table}. 
\end{itemize}
% \begin{figure*}[
% 	\begin{center}
% 		\begin{tabular}{cr}
% 			\includegraphics[width=0.33\textwidth,clip,trim=0 -10 0 0]{img_DBLP_folds-eps-converted-to.pdf} & 
% 			\includegraphics[width=0.66\textwidth,clip,trim=30 0 0 0]{img_DBLP_percent_AUC_MAP-eps-converted-to.pdf}  \\
% 			(A) & (B) \hspace{152px} (C) \hspace{68px} \\
% 		\end{tabular}
% 		\caption{\textbf{DBLP results.} (A) The performance of 5-fold cross-validation. The average and standard deviation across the 5 folds are shown. (B) \& (C) The performance of using various percentages of training data. The average and standard deviation are shown for each percentage across different random samplings. }
% 		\label{DBLP_figures}
% 	\end{center}
% 	\vskip -0.3in
% \end{figure*}

% \begin{table*}[ht]
% 	\centering
% 	\resizebox{\textwidth}{!}{\begin{tabular}{|c|c|c|c|c|c|c|c|}
% \hline
% \multicolumn{2}{|l|}{}             & LowrankTLP & ApproxLink & GraphCP & GraphCP-W & CP    & CP-W  \\ \hline
% \multirow{2}{*}{5 nets}  & AUC & 0.990      & 0.610      & 0.540   & 0.543     & 0.549 & 0.522 \\ \cline{2-8} 
%                              & MAP & 0.991      & 0.679      & 0.539   & 0.535     & 0.528 & 0.486 \\ \hline
% \multirow{2}{*}{10 nets} & AUC & 0.942      & 0.554      & 0.536   & 0.533     & 0.520 & 0.472 \\ \cline{2-8} 
%                              & MAP & 0.952      & 0.672      & 0.533   & 0.550     & 0.527 & 0.496 \\ \hline
% 	\end{tabular}}
% 	\caption{\textbf{Effectiveness comparison in simulations.}}\label{simluation_table}
% 	\vskip -0.2in
% \end{table*}

\subsection{Predicting multi-relations in scientific publications} \label{sec-DBLP}
We downloaded the DBLP dataset of scientific publication records from AMiner (Extraction and Mining of Academic Social Networks) \cite{tang2008arnetminer}. %The dataset contains 2,092,356 papers, 8,024,869 citations, 1,712,433 authors and 4,258,946 collaborations between authors, and a total of 264,025 distinct venues in which the papers were published. 
We built three graphs: Author $\times$ Author ($W^{(1)}$), Paper $\times$ Paper ($W^{(2)}$) and Venue $\times$ Venue ($W^{(3)}$). In $W^{(1)}$, the edge weight is the count of papers that both authors have co-authored; in $W^{(2)}$, the edge weight is the number of times both papers were cited by another paper; and in $W^{(3)}$, the edge weight is calculated using Jaccard similarity between the vectors of the two venues whose dimensions are a bag of citations. After filtering the vertices with zero and low degrees in each graph, we finally obtained $W^{(1)}$ with 13,823 vertices and 266,222 edges; $W^{(2)}$ with 11,372 vertices and 4,309,772 edges; and $W^{(3)}$ with 10,167 vertices and 46,557,116 edges, similar to the dataset used in \cite{liu2016cross}. We also built 12,066 triples in the form (Paper, Author, Venue) as positive multi-relations, given by the natural relationship that a paper is written by an author, and published in a specific venue. These triples were stored in the sparse initial tensor $\mathcal{Y}^0$  as input, whose dimensions match with the graph sizes.

We first performed 5-fold cross-validation with 3-fold training triples, 1-fold validation triples to select the best hyperparameters and 1-fold test triples for all the methods, using the 12,066 positive triples together with the same number of randomly sampled negative triples. Figure \ref{DBLP_figures}(A) shows the performance comparisons with standard deviations on all the 5 test folds.  We observed that LowrankTLP clearly outperforms the baselines in every fold, and the methods utilizing the graph information outperform the CP and CP-W, which do not use graph information.  We also randomly sampled 0.1\%, 10\%, 50\% and 90\% of positive triples as training data to test the rest of the positive triples together with the same number of randomly sampled negative triples. The random samplings are repeated 5 times for each percentage.  Using the optimal hyperparameters chosen from the previous 5-fold cross-validation, we compared the performance of all the methods on various percentages of training/test data. Figure \ref{DBLP_figures}(B)\&(C) show that LowrankTLP consistently outperforms all the baselines in every training percentage. Remarkably, both LowrankTLP and ApproxLink achieve average AUC $\approx$ 0.76 and MAP $\approx$ 0.8 when there are only 0.1\% of training data; LowrankTLP, ApproxLink and TOP are more robust to sparse input, comparing with CP-based methods GraphCP and GraphCP-W, which also use TPG information; the methods utilizing knowledge graphs perform consistently better than CP and CP-W, which do not use knowledge graph, demonstrating that associations among the tensor entries carried by the manifolds in the TPG are useful to enhance the prediction performance.
%\vspace{-0.1in}

\begin{figure}[t]
	\begin{center}
		\includegraphics[width=1\columnwidth]{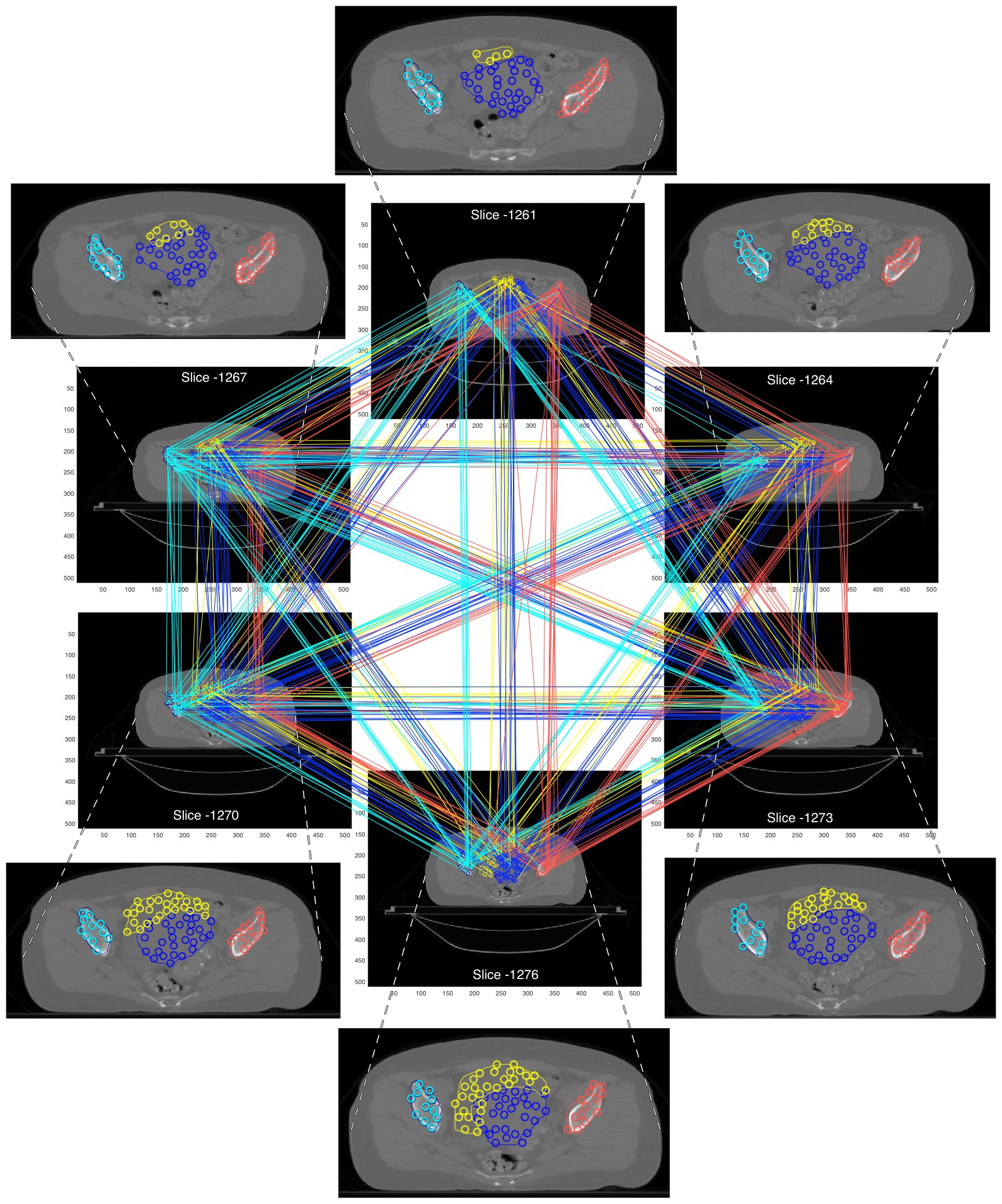}
		\vskip -0.1in
		\caption{\textbf{Example of aligning 6 CT scan images.} Each type of segmented region in the images is represented by a different color in the alignment. The links connect all the pairs of the spots in a 6-tuple with one from each image to represent one alignment.}
		\label{img-image_alignment}
	\end{center}
	\vskip -0.1in
\end{figure}

\begin{figure}[t]
	\begin{center}
		\resizebox{\columnwidth}{!}{\begin{tabular}{cc}
				\includegraphics[width=0.5\textwidth,clip,trim=30 0 30 0]{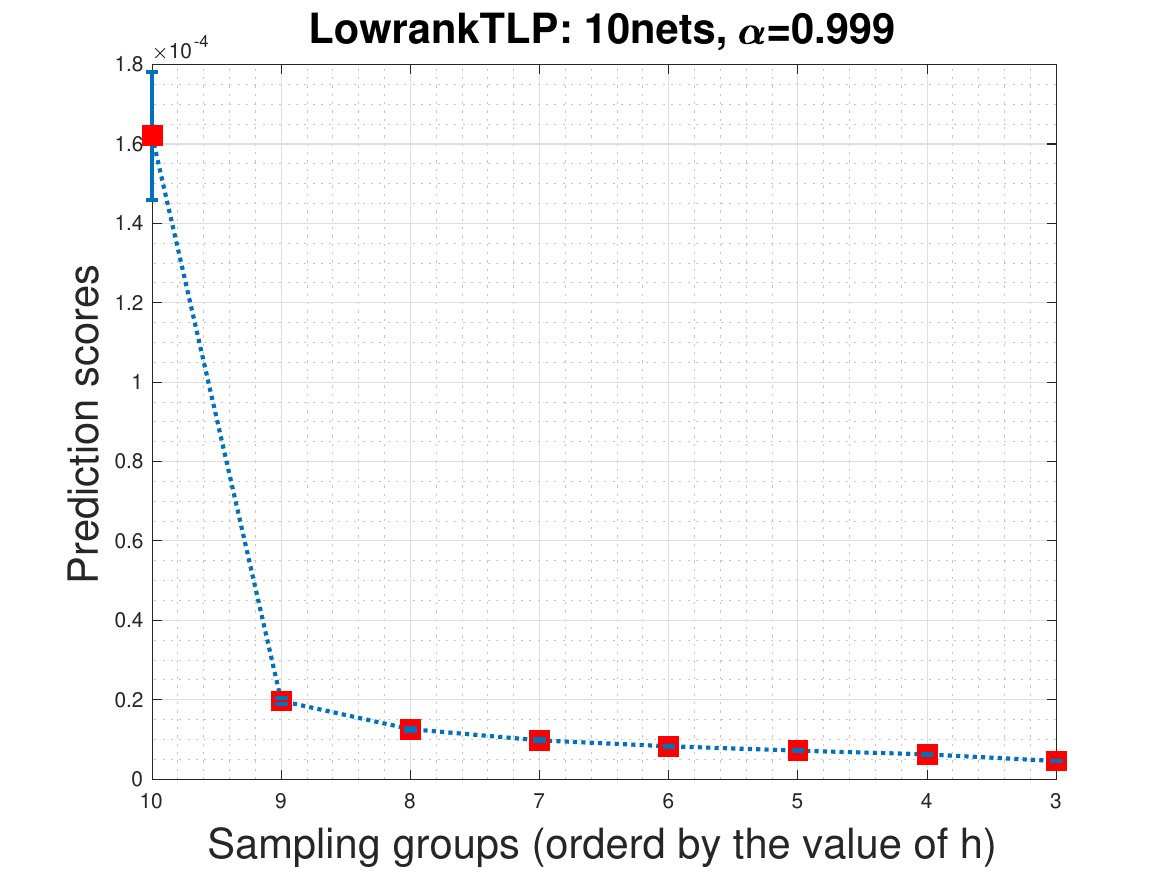} & 
				\includegraphics[width=0.5\textwidth,clip,trim=30 0 30 0]{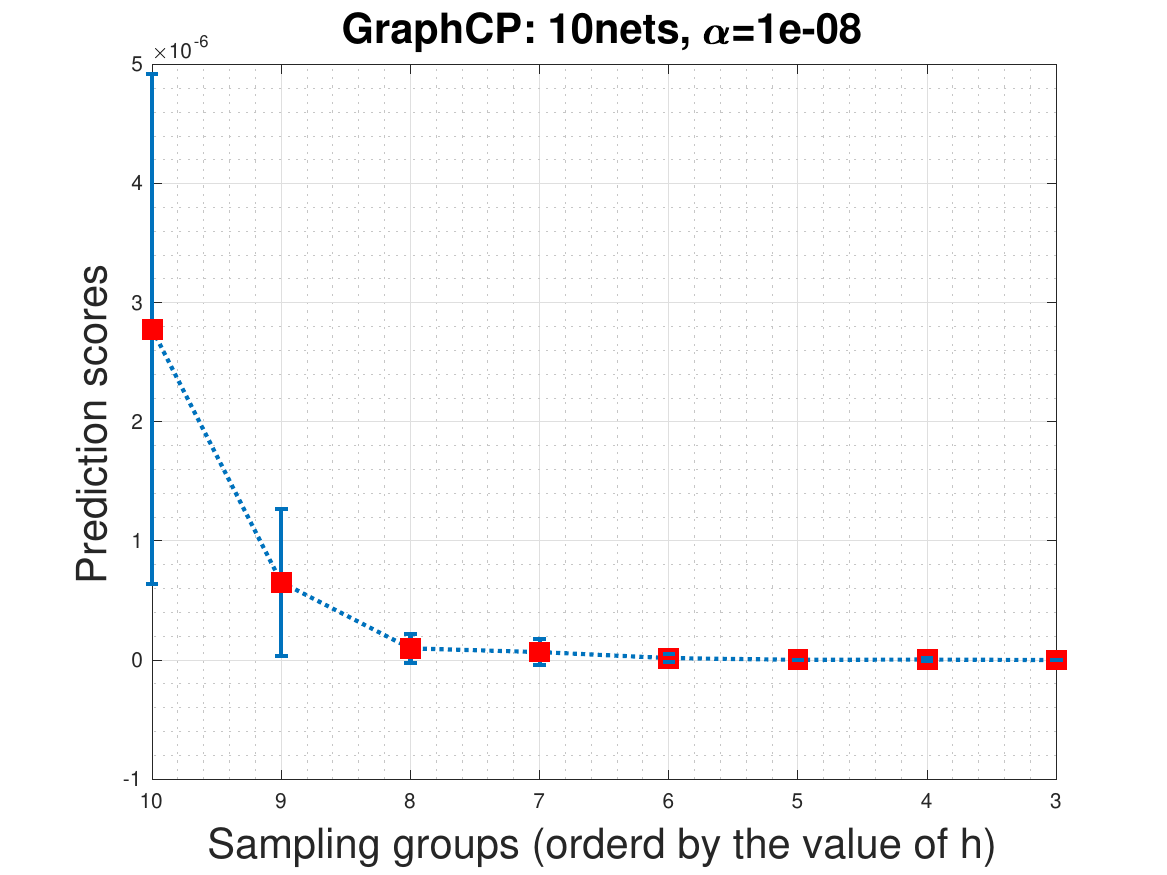} \\
				
				\multicolumn{2}{c}{{(A) Alignment of 10 CT scans}} \\
				%\vspace{-8pt}
				\includegraphics[width=0.5\textwidth,clip,clip,trim=30 0 30 0]{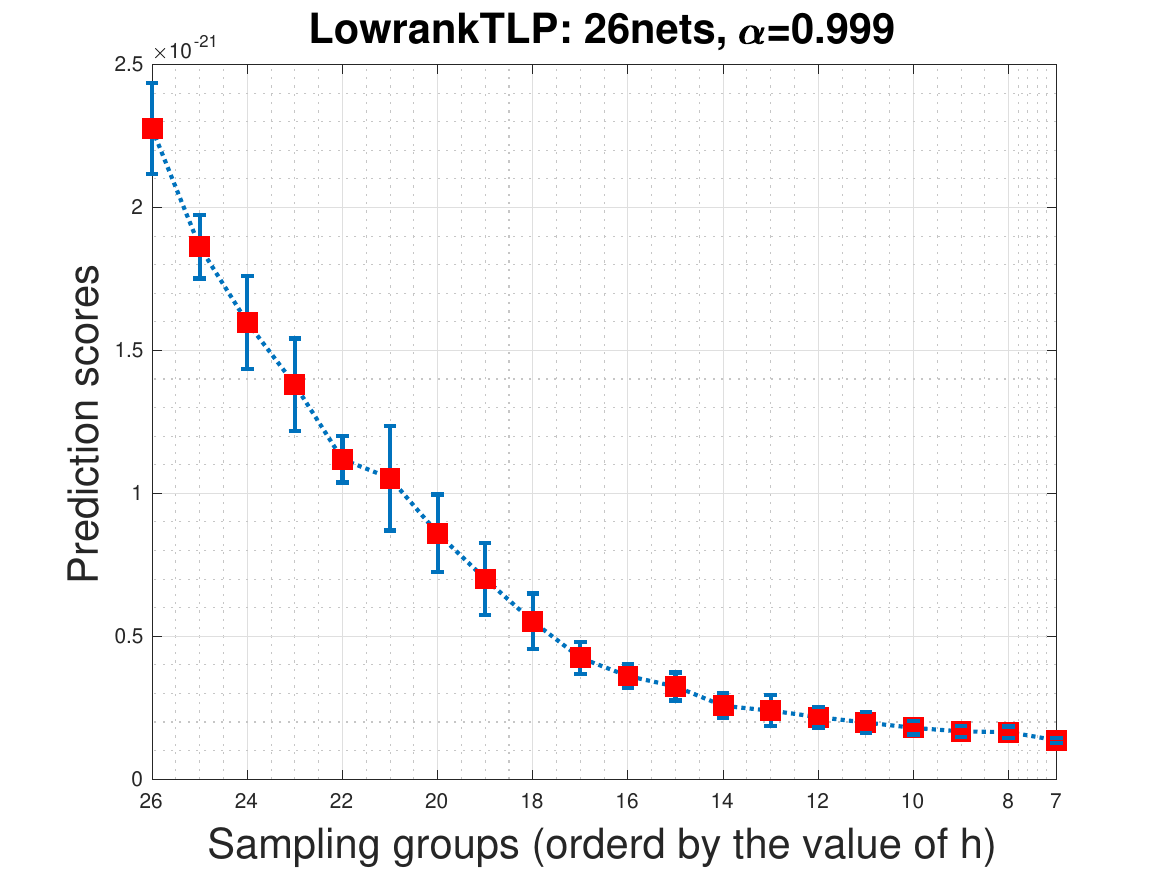} & 
				\includegraphics[width=0.5\textwidth,clip,clip,trim=30 0 30 0]{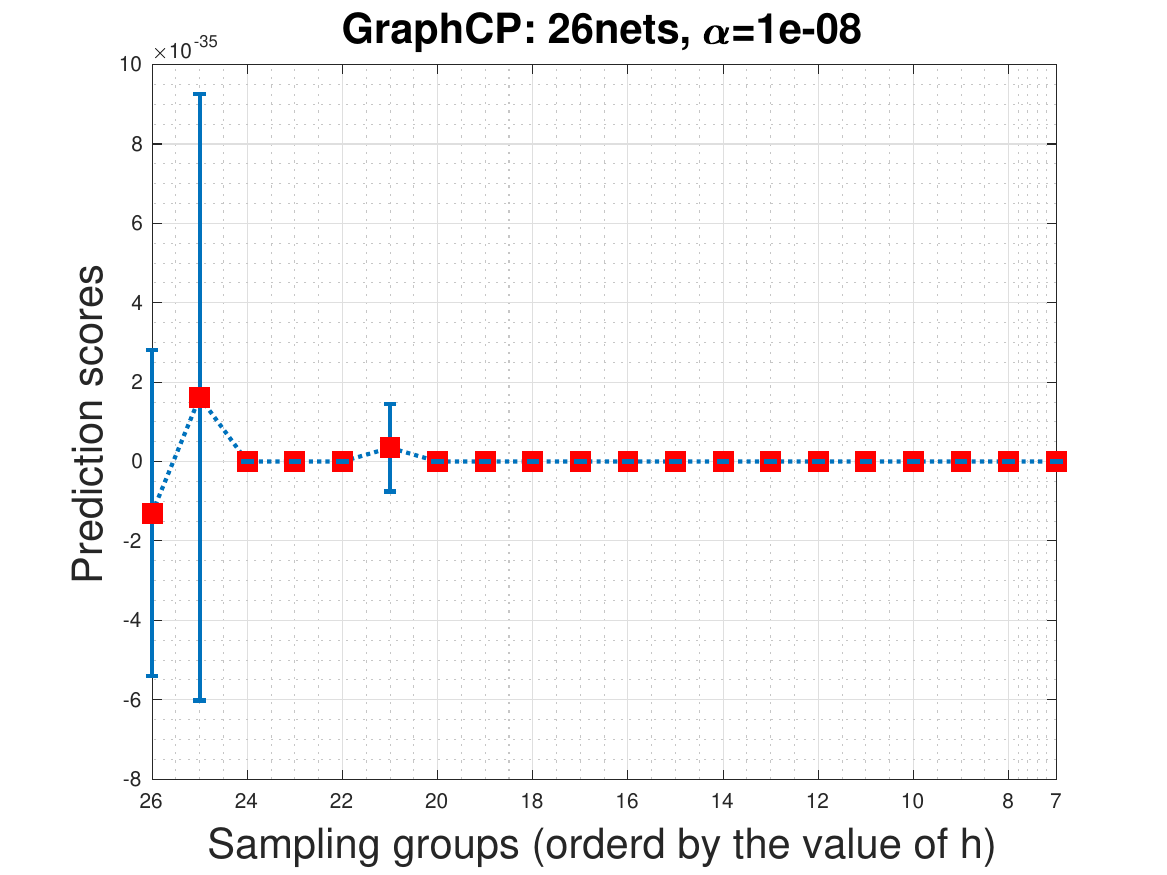} \\
				\multicolumn{2}{c}{(B) Alignment of 26 CT scans}\\
		\end{tabular}}
		\caption{\textbf{Results of aligning 10 and 26 CT scan images.} The average and standard deviation of the prediction scores of the $10^4 $ $n$-tuples in each sampling group ordered by the homogeneity score $h$ is shown.}\label{CT_26_results}
	\end{center}
	\vskip -0.3in
\end{figure} 
\subsection{Alignment of CT scan images} \label{sec-CT}
We obtained a dataset of 134 CT scan images of an anonymized female patient. The scans were acquired on a Philips Brilliance Big Bore CT Scanner,  and each image has 512 $\times$ 512 pixels with a slice thickness of 3mm. We used a subset of 26 images which contain the same set of four segmented regions manually annotated by a radiologist. When working with CT scan images, a radiologist is interested in matching the segmented regions across the images. We represent this situation by aligning a set of sampled spots across the images to detect if they belong to the same type of segmented region. To construct a graph for each CT image, we first sampled from each segmented region in each image a number (proportional to the region size) of spots.  Then, we calculated the similarity between the spots using the RBF function:
$s(x_i, x_j)= \text{exp}(-\frac{||x_i - x_j||^2}{\sigma})$ if $\phi(x_i) \ne \phi(x_j)$ and otherwise 1,
%\begin{equation}
%s(x_i, x_j)= 
%\begin{cases}
%1,& \text{if } \phi(x_i) = \phi(x_j) \\
%e^{-\frac{||x_i - x_j||_2^2}{\sigma}},              & \text{otherwise, }
%\end{cases}
%\end{equation}
where $x_i$ and $x_j$ are the coordinates of the two spots; $\phi(x)$ represents the region where the spot $x$ is located; $\sigma=10$ is the width of RBF function. The pairwise similarity scores between the spots in two different images were obtained by the color density difference between the spots, calculated using a RBF function with $\sigma=10^{-2}$. The initial tensor $\mathcal{Y}^0$ was then generated in CP-form using these cross-image spots similarity matrices. For example, to align 7 images, $\binom{7}{2}=21$ similarity matrices were generated. In this setting, the number of spots can be different across the images. Therefore, it is possible that one spot in an image is matched to more than one spot in another image after the alignment.

\begin{table}[t]
	\centering
	\resizebox{\columnwidth}{!}{\begin{tabular}{|c|c|c|c|c|c|c|}
			\hline
			\multicolumn{1}{|l|}{}         & \multicolumn{4}{c|}{\textbf{LowrankTLP}}             &      &         \\ \hline
			& \textbf{$ k = 10$} & \textbf{$k = 10^2$} & \textbf{$k = 10^3$} & \textbf{$k = 10^4$} & \textbf{CP-form $\mathcal{Y}^0$}   & \textbf{GraphCP} \\ \hline
			\textbf{4 images}                       & 0.59    & 0.91     & 0.91      & 0.91       & 0.61 & 0.73    \\ \hline
			\textbf{5 images}                       & 0.67    & 0.80     & 0.89      & 0.89       & 0.66 & 0.75    \\ \hline
			\textbf{6 images}                       & 0.78    & 0.78     & 0.84      & 0.84       & 0.69 & 0.78    \\ \hline
			\multicolumn{1}{|l|}{\textbf{7 images}} & 0.75    & 0.73     & 0.80      & 0.83       & 0.72 & 0.76    \\ \hline
	\end{tabular}}
	\caption{\textbf{Performance of CT image alignment.} The accuracy is measured as the \% of accurately aligned spots in the first image to the correct spots in the rest of the images.\label{CT_table}}
	\vskip -0.4in
\end{table}

The set of query tuples were selected if the color densities between each pair of the spots in a tuple are all above a threshold. The alignment accuracy was measured by the top-1 match of each spot. Specifically, for each spot in the first graph, we took the sub-tensor of dimension $(n-1)$ associated with the entry in the first dimension to find the entry of the highest score in the sub-tensor. Then, we checked if the features of the aligned spots from all the other images in the maximum entry were the same as the spot in the first dimension. Table \ref{CT_table} shows the comparisons of LowrankTLP and GraphCP, using CP-form tensor $\mathcal{Y}^0$ as their initialization. With $k \geq 100$, LowrankTLP achieves much higher accuracy than GraphCP in almost all the cases. It is also interesting that with $k=10,000$, LowrankTLP is able to align 7 images with an accuracy of 0.83, which means 83\% of the spots in the first graph is perfectly matched with a spot from the same type of segmented region in all other 6 images. An example of 6 aligned images is shown in Figure \ref{img-image_alignment}. It is clear that the aligned spots are consistent across the images. 
%Note that we are able to align as many as 10 images with LowrankTLP in the experiment. However, the evaluation of the 10-way multi-relations is too computationally difficult to enumerate the aligned spots in the sub-tensor. Thus, the results are not shown.

More importantly, to further measure the scalability of LowrankTLP on a larger number of real graphs, we performed an additional evaluation by aligning 10 and 26 CT scan images. Since it is not computationally feasible to enumerate every entry of the $10$-way tensor and the $26$-way tensor, we generated a list of candidate $n$-tuples for performance evaluation. Given $n$ images to be aligned, we randomly sampled a list of $n$-tuples of spots. The $n$-tuples were then grouped by their homogeneity score $h$, where $h$ is defined as the maximum number of spots that are from the same type of segmented region in the $n$-tuple. For example, if there are 3, 4, 10 and 9 spots in a 26-tuple from the 4 types of regions, respectively, the homogeneity score of this 26-tuple will be $\text{max}(3, 4, 10, 9) = 10$. Based on the homogeneity score, we generated sampling groups of varying $h$ to evaluate the alignment of $n = 10$ and $n = 26$ images. We expect that the sampling groups of larger $h$ also receive higher prediction scores on the $n$-tuples in the groups. GraphCP is the only baseline that is both applicable and scalable in this experiment for comparison.

The average and standard deviation of the prediction scores for each sampling group are shown in Figure \ref{CT_26_results}. In the alignment of 10 images shown in Figure \ref{CT_26_results} (A), we observe that LowrankTLP generates a much larger average score for $h = 10$ compared with the sampling groups with $h<10$, and the average score decreases consistently and monotonically as $h$ decreases. GraphCP is also able to identify the group of $h = 10$ but the variance is large and a flatter tail is observed after $h=6$.  In the aligment of 26 images shown in Figure \ref{CT_26_results} (B), GraphCP completely fails to distinguish the most significant group $h=26$ from the other groups, whereas LowrankTLP maintained the same clear decreasing trend as $h$ decreases. This comparison implies LowrankTLP is more applicable to high-order TPG of a large number of graphs in real-world problems. As discussed in the implementation details in Section 7.1, the graph hyperparameter $\alpha$ of GraphCP was set to be very small when the number $n$ of graphs is large. In Appendix \ref{sec:Supplementary} Figure \ref{fig:app}, we also provide more comprehensive comparisons of LowrankTLP and GraphCP by varying the graph hyperparameter $\alpha$. Very similar results are observed. %The results clearly show that the graph manifolds information play a key role for enhancing the prediction performance of LowrankTLP, given the observation that the decreasing trend of the curve becomes clearer as $\alpha$ increases.

%\textcolor{red}{In order to show the advantages of LowrankTLP when a large number of graphs is used, we performed additional evaluation by aligning up to 26 CT Scan images. Since we cannot evaluate every entry of the multi-relations, we generated a list of candidate tuples. Given $n$ images to be aligned, we sampled a set of $n$-tuples in which only $k \in [\lceil\frac{n}{4}\rceil : n]$ images are represented by the same segmented regions in each tuple. $10^4$ $n$-tuples were generated for each $k$. We expect that LowrankTLP is able to obtain larger scores when more images are represented by spots of the same segmented region in the $n$-tuples. Figure \ref{CT_26_results} (A) and (B) shows the results of the analysis. When 10 images are aligned, shown in Figure \ref{CT_26_results} (A), we can notice that LowrankTLP (left) generates much larger scores when all the 10 elements in the tuple are from the same segmented region in each image. GraphCP (right) was also able to identify the case with all images having the same feature, but with a lower difference between 9 and 10. Moreover, GraphCP exhibits higher variance than LowrankTLP across the $10^4$ samplings. When using 26 images, shown in Figure \ref{CT_26_results} (B), GraphCP (right) fails to identify the most significant cases, where LowrankTLP (left) has a clear trend according to the number of images with the same feature in the tuple.  }
\subsection{Alignment of PPI Networks} \label{sec-ppi}
%\textbf{IsoRankN} \cite{liao2009isorankn} and \textbf{BEAMS} \cite{Alkan2013beams}: both methods are developed for the global alignment of multiple protein–protein
%interaction networks (PPIs). The inputs are PPIs $W^{i} ,i=1,\dots n$ for $n$ different species, and ${n}\choose{2}$ matrices each of which stores the pair-wise protein sequence similarities between two species. The outputs are the groups of proteins from different species where the proteins within each group should have a high functional consistency.
%\begin{table}[h]
%	\centering
%	\begin{tabular}{l|llll}
%		& \textbf{HS} & \textbf{DM} & \textbf{SC} & \textbf{CE} \\ \hline
%		\textbf{\# of Proteins}     & 10,403      & 7,396      & 5,524       & 2,995      \\
%		\textbf{\# of Interactions} & 109,822     & 49,991      & 165,588     & 9,711      
%	\end{tabular}
%	\caption{Number of proteins and associations in PPI networks.}\label{tb-ppi}
%\end{table} 

We downloaded the IsoBase dataset \cite{singh2008global,liao2009isorankn,park2010isobase}, containing protein-protein interactions (PPI) networks for five species: \textit{H. sapiens} (HS), \textit{D. melanogaster} (DM), \textit{S. cerevisiae} (SC), \textit{C. elegans} (CE) and \textit{M. musculus} (MM). The \textit{M. musculus} network only contains 776 interactions and is dropped from the analysis. After removing proteins with no association in the PPI networks, there are 10,403, 7,396, 5,524 and 2,995 proteins and 109,822, 49,991, 165,588 and 9,711 interactions in the HS, DM, SC and CE PPI networks, respectively. The dataset also contains cross-species protein sequence similarities as BLAST Bit-values for all the pairs of species. Similar to the CT scan experiment, we generated the input tensor $\mathcal{Y}^0$ in CP-form whose dimensions are matched with the number of proteins in the corresponding species, by using the pairwise BLAST sequence similarity scores.  In addition, the annotations of the proteins with 37,463 gene ontology (GO) terms below level five of GO are also provided for evaluation. We generated a set of query tuples of proteins with high sequence similarity between all the protein pairs in the tuple. These tuples can then be classified as true multi-relations if all the annotated proteins in the tuple share at least one common GO term, and otherwise false multi-relations.  
%This process was similar to how the candidates entries were generated in our CT Scan experiment. At the end of the process, we have a set of tuples, with an associated score, which we can use to select the set of entries we want to use in the evaluation.
The experiments were performed using three species (HS, DM and SC) and four species by adding CE. Around 3M tuples were generated among three species and about 163M among four species. 

\begin{figure}[t]
	\begin{center}
		\resizebox{\columnwidth}{!}{\begin{tabular}{cc}
				\includegraphics[width=0.5\textwidth,clip,trim=0 0 15 0]{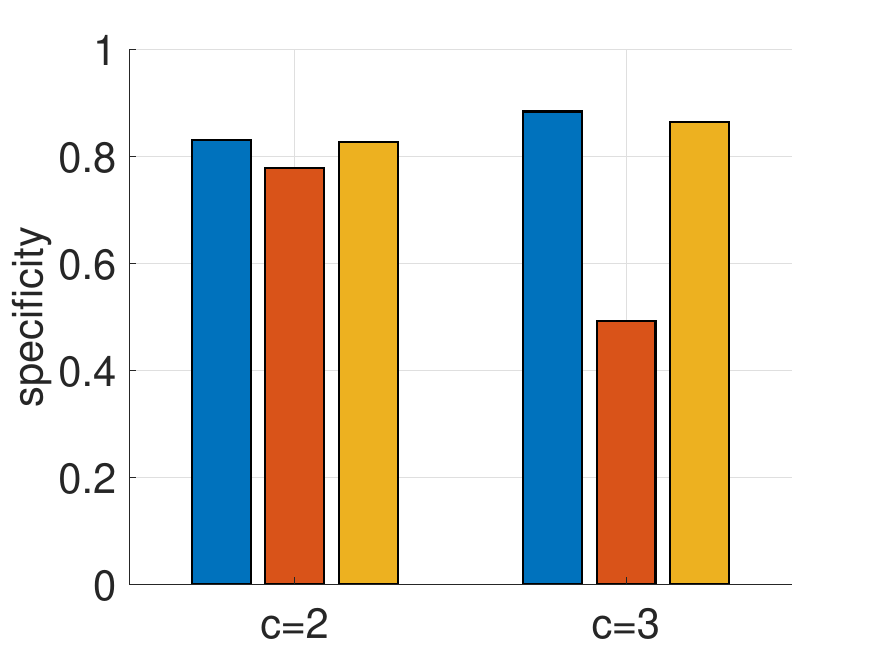} & 
				\includegraphics[width=0.5\textwidth,clip,trim=15 0 0 0]{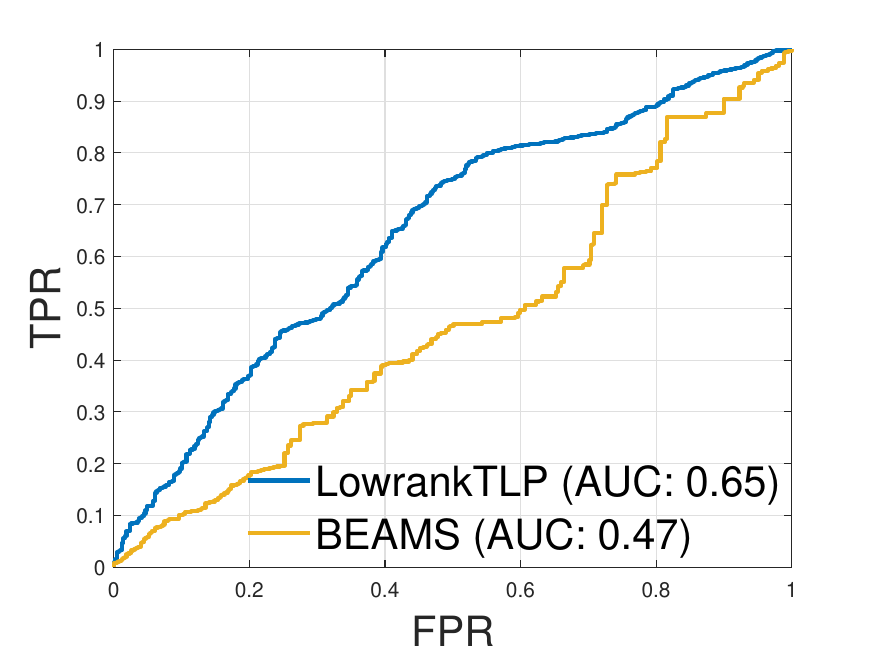} \\
				
				\multicolumn{2}{c}{{(A) Alignment of HS/DM/SC networks}} \\
				%\vspace{-8pt}
				\includegraphics[width=0.5\textwidth,clip,trim=0 0 15 0]{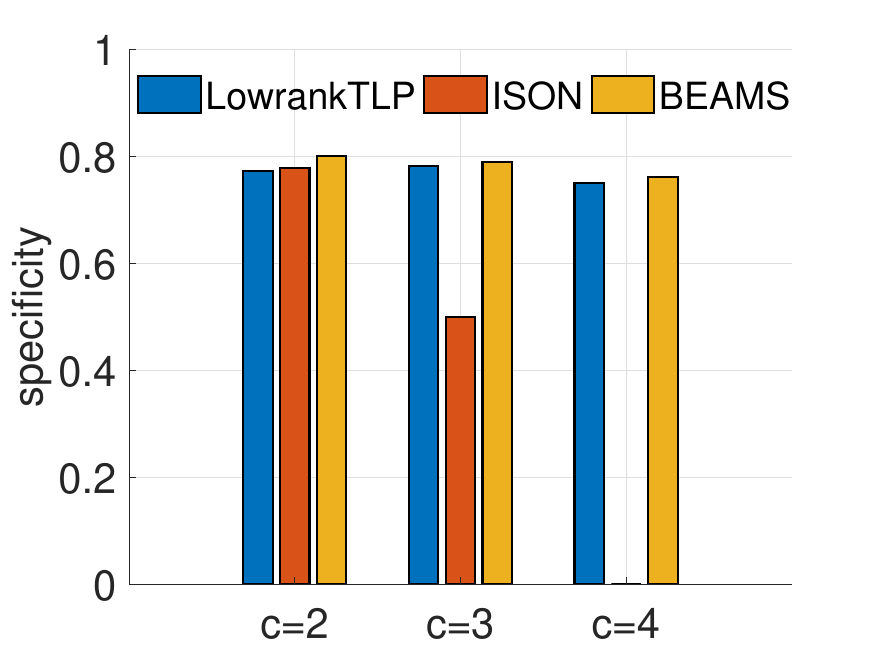} & 
				\includegraphics[width=0.5\textwidth,clip,trim=15 0 0 0]{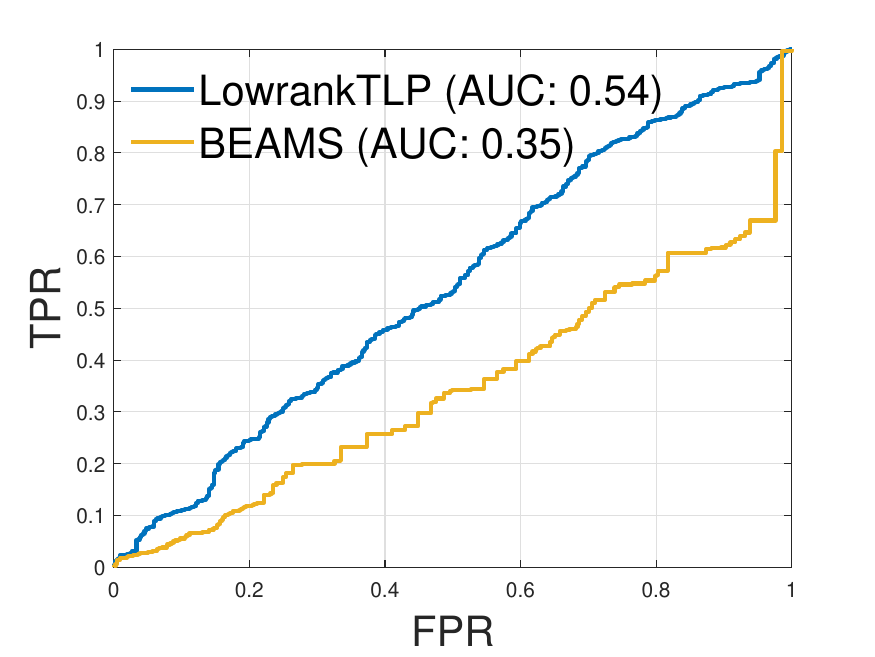} \\
				\multicolumn{2}{c}{(B) Alignment of HS/DM/SC/CE networks}\\
		\end{tabular}}
		\caption{\textbf{Results of PPI network alignment.} In both (A) and (B), the figure on the left shows the specificity of the detected clusters containing different number of species, and the figure on the right shows the AUC curves between consistent and inconsistent query entries by prediction among the clusters reported by BEAMS. \label{PPI_results}}
	\end{center}
	\vskip -0.3in
\end{figure}

Similar to the post-processing in the evaluation in \cite{Alkan2013beams}, after applying LowrankTLP to generate the prediction scores for all the query tuples, the tuples were sorted for a greedy merge as protein clusters for standard evaluation of PPI network alignment. A cluster of size $n$ is defined as a set of proteins with at least one protein from each of the $n$ species. The greedy merge scans the tuples and adds the tuple that only contains proteins not seen yet as a new cluster. Otherwise, the proteins that are already in some other clusters are removed from the tuple, and the remaining proteins are added as a smaller cluster. In the evaluation, the specificity is defined as the ratio between the number of consistent clusters and the number of annotated clusters, where an annotated cluster is a cluster in which at least two proteins are associated to at least one GO term, and a consistent cluster is the one in which all of its annotated proteins share at least one GO term. In the left plot in Figure \ref{PPI_results} (A), for both clusters of size 2 and 3, LowrankTLP performs better than both BEAMS and IsorankN in the alignment of the three networks. The left plot in Figure \ref{PPI_results} (B) shows that LowrankTLP performed similarly or slightly worse than BEAMS in every cluster size in the alignment of four networks. IsorankN is not able to detect any cluster of size 4.

To further compare LowrankTLP with BEAMS, we analyzed the detailed ranking of the annotated clusters with at least one protein from each species reported by BEAMS. Specifically, we enumerated all the tuples containing one protein from each species from each cluster and then applied LowrankTLP to calculate the scores of all the tuples in the output tensor. We re-ranked these tuples by the scores and annotated them as consistent or inconsistent multi-relations by GO annotations. The AUC by their rankings is shown in the right plots in Figure \ref{PPI_results}. In both the three-network alignment and the four-network alignment, LowrankTLP ranks the consistent multi-relations  above the inconsistent multi-relations with AUC larger than 0.5. Notice that since we only check the very top of the predictions (those predicted as true multi-relations), the AUC is less than 0.5 for BEAMs results.  

\section{Conclusion}
In this study, we introduced a new algorithm LowrankTLP to improve the scalability and performance of label propagation on tensor product graphs for multi-relational learning. The theoretical analysis shows that the global optimal solution minimizes an estimation error bound for recovering the true tensor from the noisy initial tensor for multiple graph alignment, and provides the data-dependent transductive Rademacher bound for binary hyperlink prediction. In the experiments, we demonstrated that LowrankTLP well approximates label propagation on the normalized tensor product graph to achieve both the better scalability and performance. We also demonstrated that LowrankTLP, capable of taking either a sparse tensor or a CP-form tensor as input, is a flexible approach to meet the requirements of multi-relational learning problems in a wide range of applications. In all the experiments, we also observed that it does not require a huge rank to achieve a good prediction performance even if the size of a tensor product graph is exponential of the size of the individual graphs. This observation supports that the direct and efficient analysis of the entire spectral of the tensor product graph is a better approach. In the future, we will analyze the spectral of the tensor product graphs to develop an automatic strategy of choosing the rank $k$, for more efficient application of LowrankTLP to multi-relational learning problems.

\ifCLASSOPTIONcaptionsoff
  \newpage
\fi



%\bibliographystyle{IEEEtran}

% H.~Kopka and P.~W. Daly, \emph{A Guide to {\LaTeX}}, 3rd~ed.\hskip 1em plus

% \end{thebibliography}

\bibliographystyle{IEEEtran}
\bibliography{Bibliography}

\pagebreak
\appendices
\section{Additional Definitions and Lemmas}\label{sec:lemma}
\begin{defn}
\textbf{CANDECOMP/PARAFAC decomposition (CP):}\\ 
	\label{def:CPD}
	An $n$-way tensor $\mathcal{X} \in \mathbb{R}^{I_1\times I_2\times ... \times I_n}$ of rank $r$ can be written as
	\begin{align}
		\mathcal{X}&= \sum_{c=1}^r \boldsymbol{a}_c^{(1)} \circ \boldsymbol{a}_c^{(2)} \circ \dots \circ \boldsymbol{a}_c^{(n)} \nonumber\\
		& = \llbracket A^{(1)}, A^{(2)}, \dots, A^{(n)} \rrbracket \text{,} \nonumber
	\end{align}
	where $\boldsymbol{a}_c^{(i)}$ is the $c$-th column of factor matrix $A^{(i)} \in \mathbb{R}^{I_i \times r}$.\\
	\emph{Vectorization property:} The vectorization of CP-form is $vec(\mathcal{X}) = (A^{(n)}\odot A^{(n-1)}\odot  \dots \odot A^{(1)}) \boldsymbol{1}$, where $\boldsymbol{1}$ is a vector with all-ones. 
\end{defn}

\begin{defn}
\textbf{Tucker decomposition:}\\  \label{def:tucker}
	An $n$-way tensor $\mathcal{X} \in \mathbb{R}^{I_1\times I_2\times ... \times I_n}$ can be decomposed into a core tensor $\mathcal{G} \in \mathbb{R}^{r_1\times r_2\times ... \times r_n}$ and factor matrices \{$A^{(i)} \in \mathbb{R}^{I_i \times r_i} :   i=1, \dots, n$\} as 
	\begin{align}
	\mathcal{X} & = \mathcal{G} \times_1 A^{(1)} \times_2 A^{(2)} \dots \times_n A^{(n)} \nonumber\\
	& =  \llbracket \mathcal{G}; A^{(1)}, A^{(2)}, \dots,  A^{(n)} \rrbracket \nonumber.
	\end{align}
	\emph{Vectorization property:} The vectorization of $\mathcal{X}$ is $vec(\mathcal{X}) = (A^{(n)} \otimes A^{(n-1)} ...\otimes A^{(1)})vec(\mathcal{G})$. 
\end{defn}
%\vspace{-10pt}
\begin{lem}
	If $A, B, C$ and $D$ are matrices of such size that one can form the matrix products $AC$ and $BD$, then $(A \otimes B)(C \otimes D) = (AC) \otimes (BD)$. \label{l1}
\end{lem}
\begin{lem}
	If matrices $A, B, C$ and $D$ are of such size that one can form the operation $(A \odot B), (C \odot D), (A^T C)$ and $(B^TD)$, then equality $(A \odot B)^T (C \odot D) = (A^T C) \circledast  (B^TD)$ holds. \label{l2}
\end{lem}
\begin{lem} 
\label{l4}
	Let $\lambda_1,\dots, \lambda_n$ be eigenvalues of $A$ with corresponding eigenvectors $\boldsymbol{x}_1,\dots,\boldsymbol{x}_n$, and let $\mu_1,\dots,\mu_m$ be eigenvalues of $B$
	with corresponding eigenvectors $\boldsymbol{y}_1,\dots,\boldsymbol{y}_m$. Then the eigenvalues and eigenvectors of $A \otimes B$ are $\lambda_i \mu_j$ and $\boldsymbol{x}_i \otimes \boldsymbol{y}_j$, $i=1,\dots,n$, $j=1,\dots,m$. 
\end{lem}
\begin{lem}\label{l5}
Let matrix $\tilde{W}^{(i)}$ denote the best rank-$k_i$ approximation to  $W^{(i)}$ per Eckart-Young-Mirsky theorem \cite{eckart1936approximation}. The matrix $\otimes_{i=1}^n \tilde{W}^{(i)}$ is not guaranteed to be the best rank $\prod_{i=1}^n k_i$ approximation to $\otimes_{i=1}^n W^{(i)}$.
\end{lem}
\section{Transductive Rademacher Complexity}
\label{sec: appendix_TRC}
The  \emph{transductive Rademacher complexity} and the data-dependent error bound for binary transductive learning proposed in \cite{el2009transductive} are given below in  Definition \ref{def: appendix_TRC} and Theorem \ref{the: appendix_TRC}.
\begin{defn}
\label{def: appendix_TRC}
Given a fixed set $\Phi_{l+u} = \{(\boldsymbol{x}_i, y _i ):  i =1,\dots, l+u  \}$ of sample-label pairs drown from an unknown distribution, w.l.o.g.,  the training set sampled uniformly without replacement from $\Phi_{l+u}$ is denoted as $\Phi_l =  \{(\boldsymbol{x}_i, y _i ):   i =1,\dots, l  \}$, and the test set is denoted as $X_u = \{\boldsymbol{x}_i :   i =l+1,\dots, l+u  \}$. Define $\mathcal{H}_{out} \subseteq \mathbb{R}^{l+u} $ as a set of vectors $\boldsymbol{h} = (h(\boldsymbol{x}_1),\dots,h(\boldsymbol{x}_{l+u}))^T$ output by a transductive algorithm using the set $\Phi_l$ and $X_u$ over all possible training/test set partitions, such that $h(\boldsymbol{x}_i)$ is the soft label of example $\boldsymbol{x}_i$. The \emph{transductive Rademacher complexity} is defined as
\begin{align*}
R_{l+u}(\mathcal{H}_{out}) = (\frac{1}{l}  +\frac{1}{u}) \mathbb{E}_{\boldsymbol{\sigma}} \Big[\sup_{\boldsymbol{h} \in \mathcal{H}_{out}}  \boldsymbol{\sigma}^T  \boldsymbol{h} \Big],
\end{align*}
where $\boldsymbol{\sigma} = (\sigma_1,\dots,\sigma_{l+u})^T$ is a vector of i.i.d random variables such that $\sigma_i = 1$ with probability $p$,  $\sigma_i = -1$ with probability $p$ and $\sigma_i = 0$ with probability $1-2p$. We set $p = \frac{lu}{(l+u)^2}$ as in \cite{el2009transductive}.
\end{defn}

\begin{thm}
\label{the: appendix_TRC}
Let $c_0 = \sqrt{\frac{32\ln{(4e)}}{3}}$, $Q = \frac{1}{l}  + \frac{1}{u}$ and $G = \frac{l+u}{(l+u-0.5)(1-0.5/\max{(l,u)})}$. For any fixed positive real $\gamma$, with probability of at least $1-\delta$ over the random training/test set partitioning, $\forall \boldsymbol{h} \in \mathcal{H}_{out}$,
\begin{align*}
\mathcal{L}_{u}^{\gamma}(\boldsymbol{h})\leq & \widehat{\mathcal{L}}_{l}^{\gamma}(\boldsymbol{h}) + \frac{R_{l+u}(\mathcal{H}_{out})}{\gamma}  + c_0 Q \sqrt{\min{(l,u)}} \nonumber \\ &+ \sqrt{\frac{GQ}{2}\ln(\frac{1}{\delta})},
\end{align*}
where $\widehat{\mathcal{L}}_{l}^{\gamma}(\boldsymbol{h})= \frac{1}{l} \sum_{i=1}^l \ell_{\gamma} (h(\boldsymbol{x}_i),y_i)$ and $\mathcal{L}_{u}^{\gamma}(\boldsymbol{h})= \frac{1}{u} \sum_{i=l+1}^{l+u} \ell_{\gamma} (h(\boldsymbol{x}_i),y_i)$ are the $\gamma$-margin empirical and test error respectively, with $\ell_{\gamma} (a,b) = 0$ if $ab > \gamma$ and $\ell_{\gamma} (a,b) = \min{(1,1-\frac{ab}{\gamma})}$ otherwise.
\end{thm}

\section{Proofs}\label{sec:proofs} 
\subsection{Proof of Theorem \ref{the: perturbation}} \label{proof 1}
\begin{proof}
We prove Theorem \ref{the: perturbation} by induction \\
When $n=2$ we have
\begin{align}
\textbf{top\_bot\_2k} &( \otimes_{i=1}^{2}\boldsymbol{\lambda}^{(i)}) = \textbf{top\_bot\_2k}(\boldsymbol{\lambda}^{(2)}\otimes \boldsymbol{\lambda}^{(1)}) \nonumber \\
&=\textbf{top\_bot\_2k}(\boldsymbol{\lambda}^{(2)}\otimes \textbf{top\_bot\_2k}(\boldsymbol{\lambda}^{(1)})) \label{eqn:step2}\\
&=\textbf{top\_bot\_2k}(\boldsymbol{\lambda}^{(2)}\otimes \textbf{top\_bot\_2k} (\Gamma^{(1)})), \nonumber
\end{align}
where Equation \eqref{eqn:step2} is based on the observation that the $k$ largest (smallest) elements in the outer product can only have at most $k$ different numbers from $\boldsymbol{\lambda}^{(1)}$. Thus, taking the top(bottom)-$k$ in $\boldsymbol{\lambda}^{(1)}$ guarantees the $k$ largest (smallest) elements in the outer product will be kept. Suppose when $n=m>2$ we have 
\begin{align}
&\textbf{top\_bot\_2k}( \otimes_{i=1}^{m}\boldsymbol{\lambda}^{(i)}) \nonumber \\ &= \textbf{top\_bot\_2k}(\boldsymbol{\lambda}^{(m)}\otimes \textbf{top\_bot\_2k}( \Gamma^{(m-1)})) \nonumber
\\ &=\Gamma^{(m)},
\end{align}
then, when $n=m+1$ the following equations hold. 
\begin{align}
\textbf{top\_bot\_2k} &( \otimes_{i=1}^{m+1}\boldsymbol{\lambda}^{(i)}) = \textbf{top\_bot\_2k}(\boldsymbol{\lambda}^{(m+1)} \otimes (\otimes_{i=1}^{(m)} \boldsymbol{\lambda}^{(i)})) \nonumber\\
&= \textbf{top\_bot\_2k}(\boldsymbol{\lambda}^{(m+1)}\otimes\textbf{top\_bot\_2k}(\otimes_{i=1}^{(m)} \boldsymbol{\lambda}^{(i)})) \nonumber \\
&=  \textbf{top\_bot\_2k}(\boldsymbol{\lambda}^{(m+1)}\otimes\textbf{top\_bot\_2k}(\Gamma^{(m)})) \nonumber
\end{align}
(End of Proof)
\end{proof}
\subsection{Proof of Proposition \ref{prop:error}}\label{proof 2}

\begin{proof}
Let $\sigma_1 > \sigma_2 > \dots > \sigma_N$ be the sorted eigenvalues of matrix $S$.
Since $ \sigma_i \in [-1,1], \ \text{for} \  i=1,\dots, N$ and $\alpha \in (0,1)$, by Eckart-Young-Mirsky theorem, the nonzero eigenvalues of $A_k$ are $\{\frac{1}{1-\alpha \sigma_i} :  i=1, \dots, k \}$ and the perturbations are given as
\begin{align*}
||A_k  - A||_2 & = \frac{1}{1-\alpha\sigma_{k+1}} \ \text{and} \\
||A_k  - A||_F & = \sqrt{\sum_{i=k+1}^N (\frac{1}{1-\alpha\sigma_i})^2}.
\end{align*}
Using $\{\sigma_i : i=1,\dots, k\}$ as eigenvalues and their corresponding eigenvectors of $S$ to construct a rank-$k$ matrix $L$, and define $B= (I-\alpha L)^{-1}$, we have $||\hat{A}-A||_2 \leq ||B -A||_2$ and $||\hat{A} - A||_F \leq ||B -A||_F$ according to the definition of $S_k$ in Section \ref{sec:optimization}. Thus, inequalities in Proposition \ref{prop:error} hold if we can prove $||B -A||_2 < ||A_k -A||_2$ and $||B -A||_F < ||A_k -A||_F$. We first obtain the perturbations as 

\begin{align}
&||B  - A||_2 = \frac{\alpha |\sigma^*|}{1-\alpha\sigma^*} \ \text{and} \nonumber \\
&||B  - A||_F = \sqrt{\sum_{i=k+1}^N (\frac{\alpha |\sigma_i|}{1-\alpha \sigma_i})^2}, \nonumber
\end{align}
where 
$\sigma^* = \text{argmax}_{\sigma \in \{\sigma_{k+1} ,\dots, \sigma_N\}} \frac{\alpha |\sigma|}{1-\alpha\sigma}$.
Now we need to show the inequalities \eqref{eq:sp} and \eqref{eq:sp1} are valid. 
\begin{align}
&||B -A||_2 < ||A_k -A||_2 \label{eq:sp}\\
&||B -A||_F < ||A_k -A||_F \label{eq:sp1}
\end{align}
 It is easy to prove  Inequality \eqref{eq:sp1} by the fact that $\alpha|\sigma_i|<1$. To show Inequality \eqref{eq:sp}, we have to consider three special cases: firstly, if $\sigma^* > 0$ and $\sigma_{k+1} \geq 0$ then we have $\sigma^* = \sigma_{k+1}$, thus Inequality \eqref{eq:sp} holds by the fact that $\alpha |\sigma^*| < 1$; secondly, if $\sigma^* < 0$ and $\sigma_{k+1} \geq 0$ we have $\frac{\alpha |\sigma^*|}{1-\alpha\sigma^*} < 1$ and $\frac{1}{1-\alpha\sigma_{k+1}} \geq 1$, thus Inequality \eqref{eq:sp} holds; finally, if  $\sigma^* < 0$ and $\sigma_{k+1} <0$ we have $|\sigma^*| \geq |\sigma_{k+1}|$, and $\frac{\alpha |\sigma^*|}{1-\alpha\sigma^*} < \frac{1}{1-\alpha\sigma^*} \leq \frac{1}{1-\alpha\sigma_{k+1}}$, thus Inequality \eqref{eq:sp} holds. Overall, we have shown
\begin{align*}
&||\hat{A} -A||_2 \leq ||B -A||_2 < ||A_k -A||_2 \ \text{and}\\
&||\hat{A} -A||_F \leq ||B -A||_F < ||A_k -A||_F.
\end{align*}

\end{proof}

\newpage
(End of Proof)
\newpage
\section{Supplementary Files}\label{sec:Supplementary}
\begin{table}[hbt!]
	\centering
	\resizebox{\textwidth}{!}{\begin{tabular}{llll}
	\hline
\multicolumn{4}{|c|}{\textbf{Task 1: hyperlink prediction}}                                                                                                                                                                                                                                                                                                                                                                                                 \\ \hline
\multicolumn{1}{|l|}{Experiment}       & \multicolumn{1}{l|}{Input relations}                                                                                                  & \multicolumn{1}{l|}{Query set}                                                                                      & \multicolumn{1}{l|}{Knowledge graph}                                                                                                            \\ \hline
\multicolumn{1}{|l|}{Simulation} & \multicolumn{1}{l|}{observed $n$-way relations}                                                                                         & \multicolumn{1}{l|}{held-out test $n$-way relations}                                                                  & \multicolumn{1}{l|}{\begin{tabular}[c]{@{}l@{}}n graphs generated by permuting a percentage\\ of edges from a common random graph\end{tabular}} \\ \hline
\multicolumn{1}{|l|}{DBLP}       & \multicolumn{1}{l|}{\begin{tabular}[c]{@{}l@{}}sampled known\\ (author, paper, venue)-relations\end{tabular}}                         & \multicolumn{1}{l|}{\begin{tabular}[c]{@{}l@{}}held-out known\\ (author,paper,venue)- relations\end{tabular}}       & \multicolumn{1}{l|}{\begin{tabular}[c]{@{}l@{}}Author $\times$ Author, Paper $\times$ Paper\\ and Venue $\times$ Venue graphs\end{tabular}}                          \\ \hline
\multicolumn{4}{|c|}{\textbf{Task 2: multiple graph alignment}}                                     \\ \hline
\multicolumn{1}{|l|}{Experiment}       & \multicolumn{1}{l|}{Input relations}                                                                                                  & \multicolumn{1}{l|}{Query set}                                                                                      & \multicolumn{1}{l|}{Knowledge graph}                                                                                                            \\ \hline
\multicolumn{1}{|l|}{CT scans}   & \multicolumn{1}{l|}{\begin{tabular}[c]{@{}l@{}}RBF similarities across spots sampled\\ from each pair of CT scan images\end{tabular}} & \multicolumn{1}{l|}{\begin{tabular}[c]{@{}l@{}}alignment scores of spots across\\ multiple images\end{tabular}}     & \multicolumn{1}{l|}{\begin{tabular}[c]{@{}l@{}}RBF similarities between spots sampled within\\ each CT scan image\end{tabular}}                 \\ \hline
\multicolumn{1}{|l|}{PPI}        & \multicolumn{1}{l|}{\begin{tabular}[c]{@{}l@{}}BLAST sequence similarities between\\ proteins from each pair of species\end{tabular}} & \multicolumn{1}{l|}{\begin{tabular}[c]{@{}l@{}}alignment scores of proteins across\\ multiple species\end{tabular}} & \multicolumn{1}{l|}{\begin{tabular}[c]{@{}l@{}}protein-protein interactions (PPI) networks for\\ different species\end{tabular}}                \\ \hline
\end{tabular}}
\caption{\textbf{Summary of datasets in the experiments \label{summary_table}}}

\end{table}

\begin{figure}[hbt!]
	\begin{center}
		\resizebox{\columnwidth}{!}{\begin{tabular}{ccc}
				\includegraphics[width=0.5\textwidth]{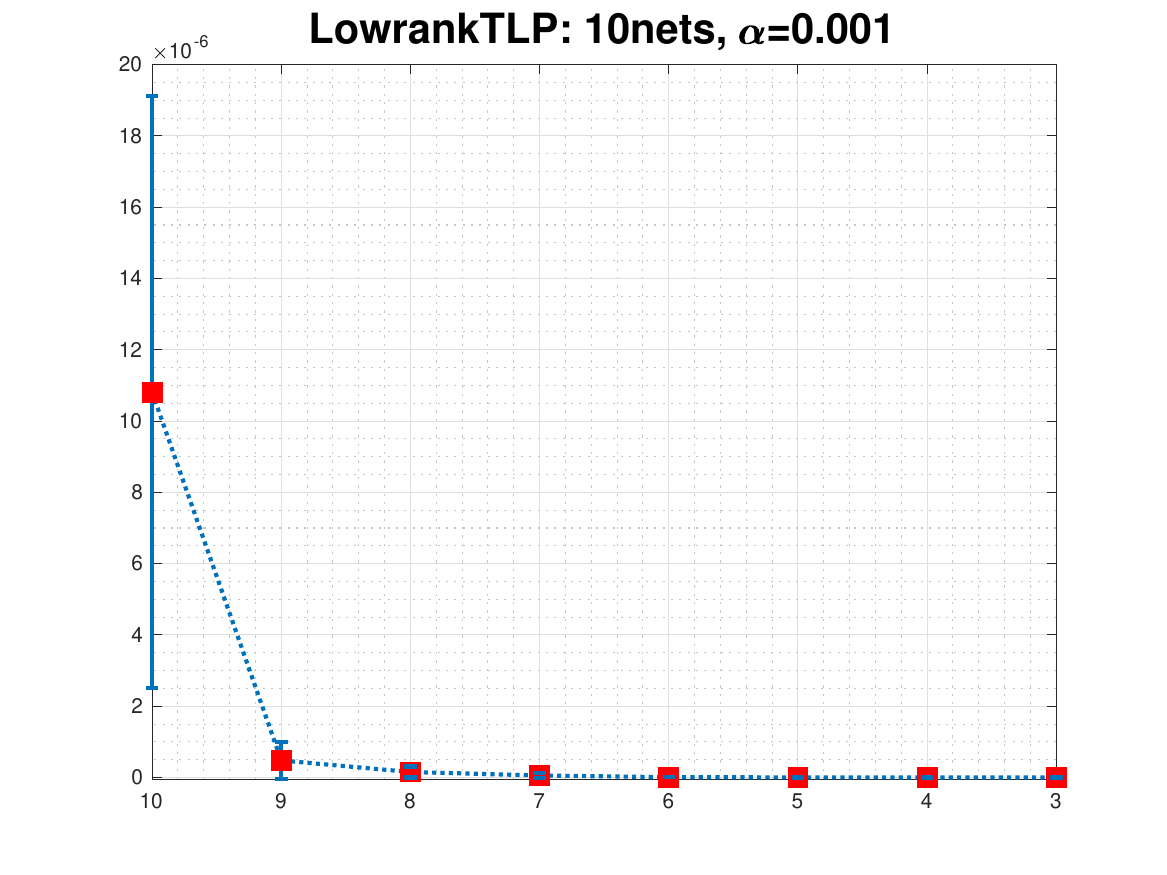} & 
				\includegraphics[width=0.5\textwidth]{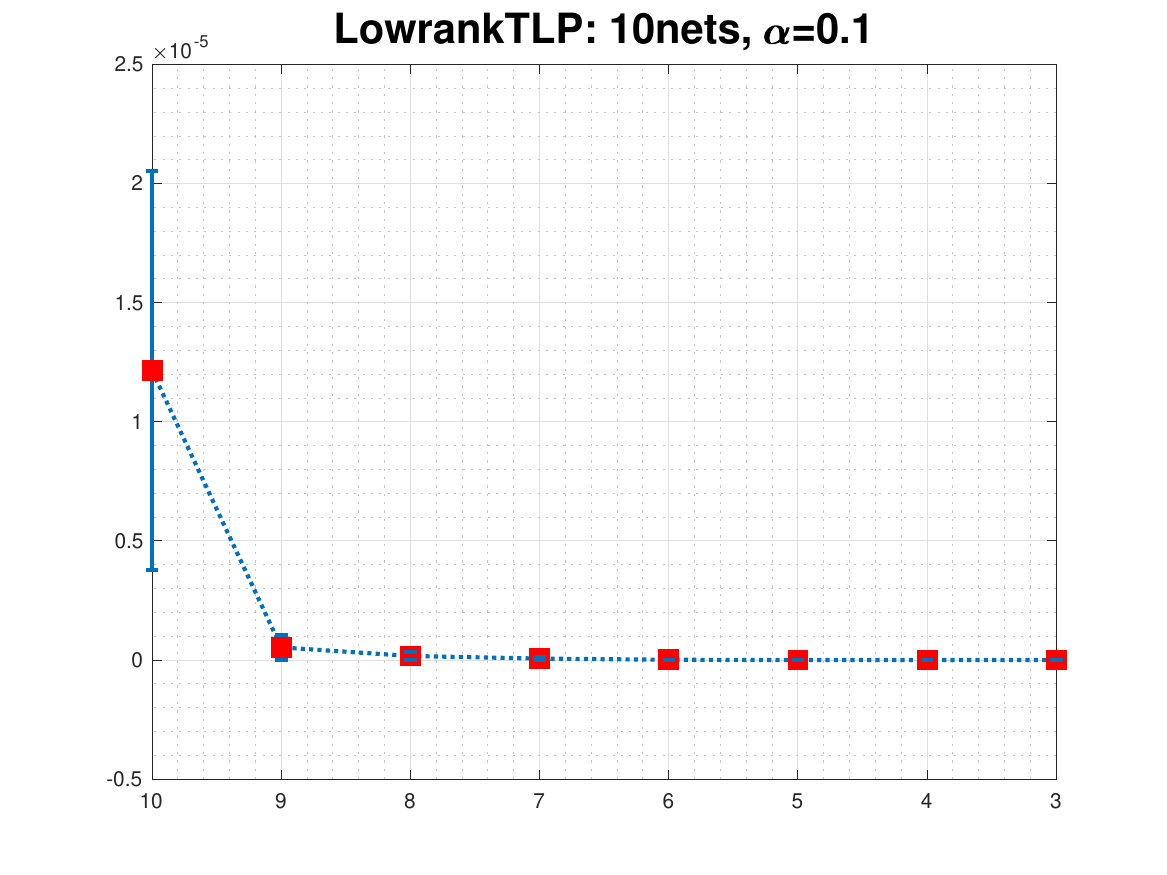} & \includegraphics[width=0.5\textwidth]{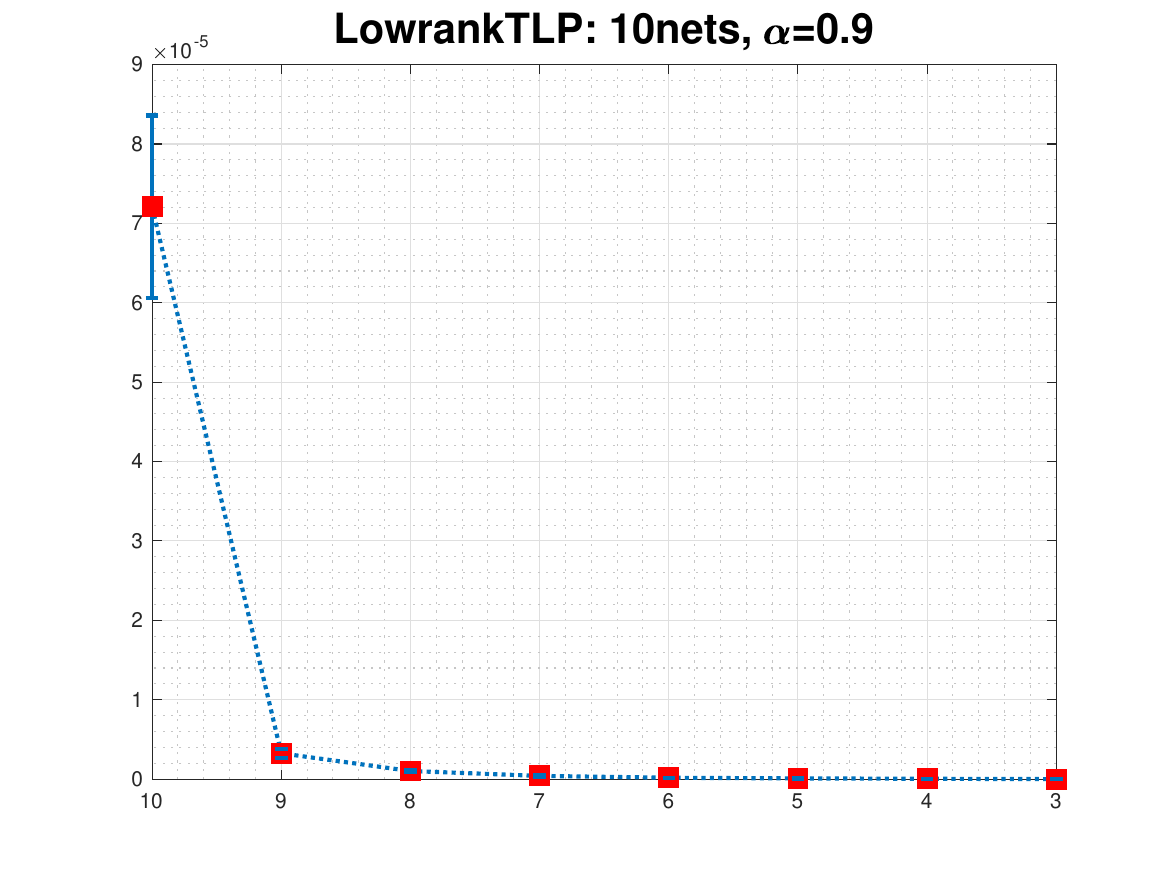} \\
				%\vspace{-8pt}
				\includegraphics[width=0.5\textwidth]{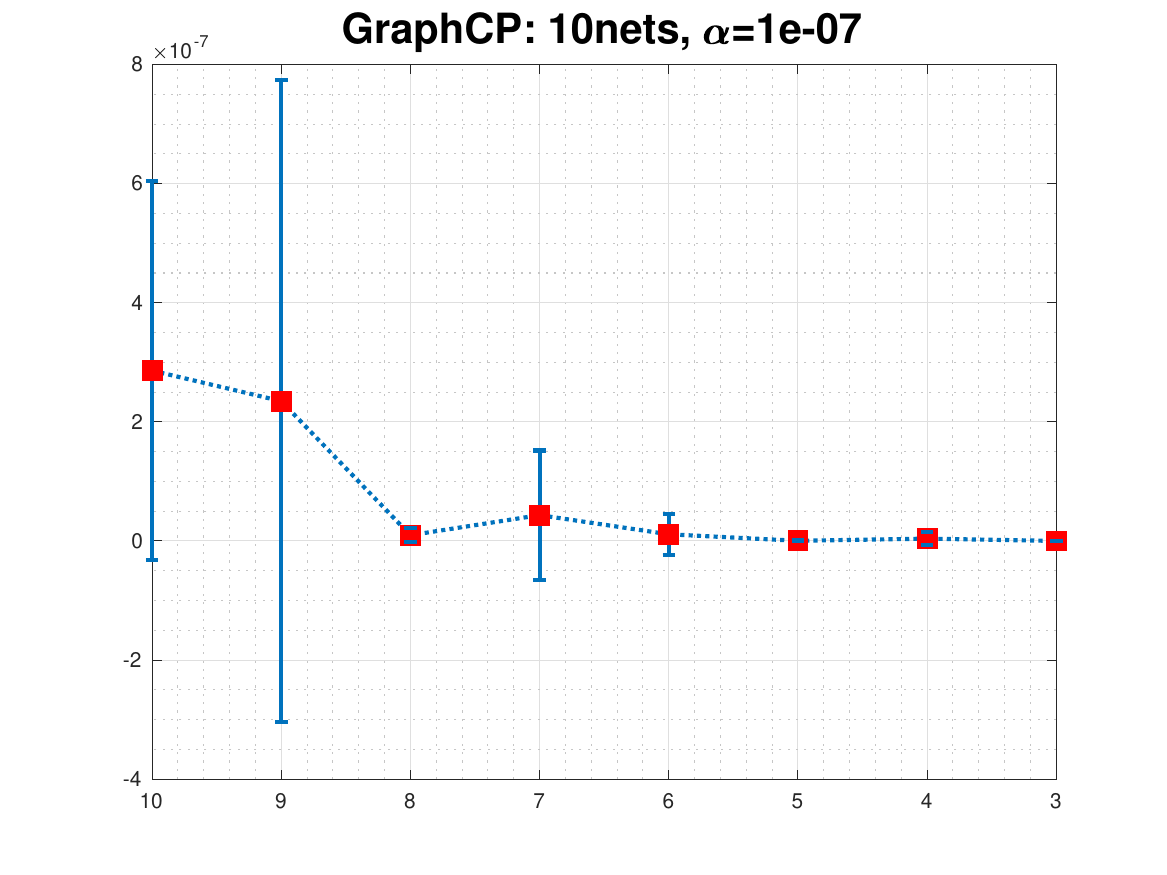} & 
				\includegraphics[width=0.5\textwidth]{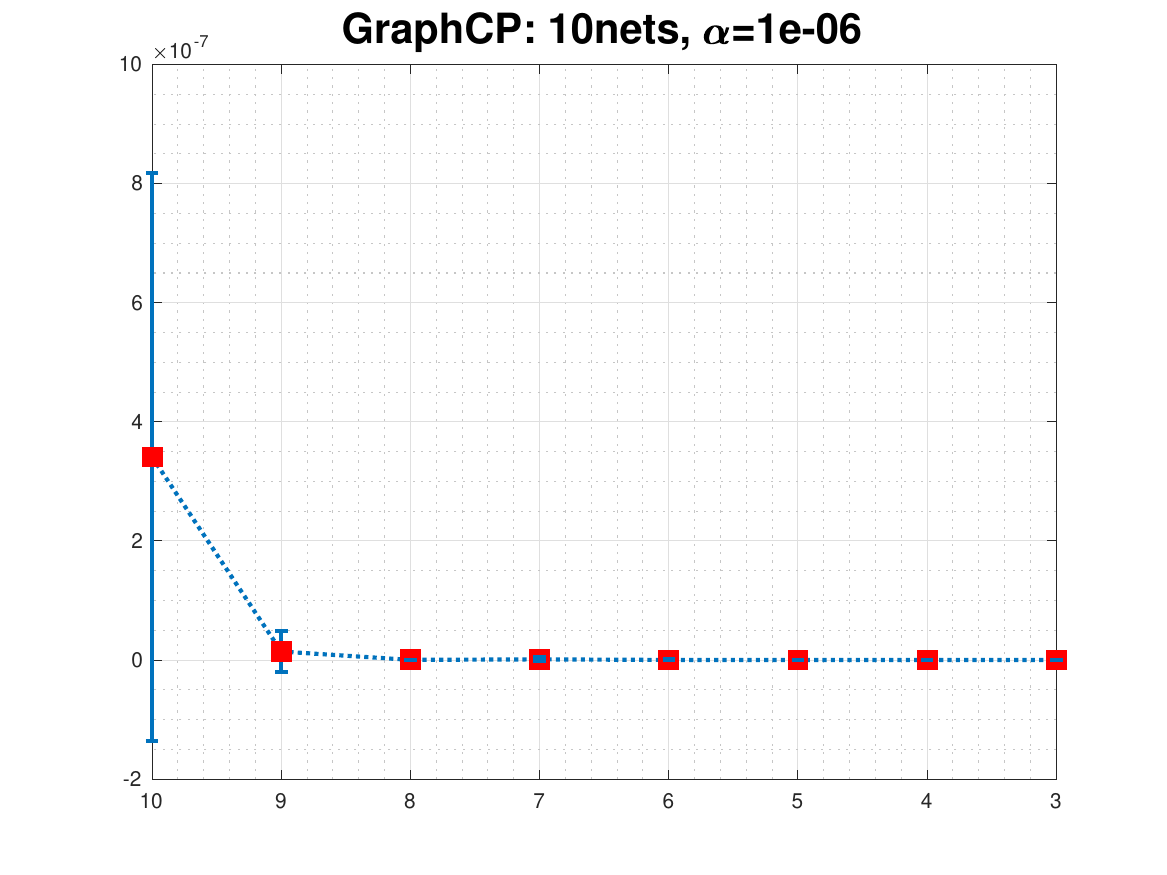} & \includegraphics[width=0.5\textwidth]{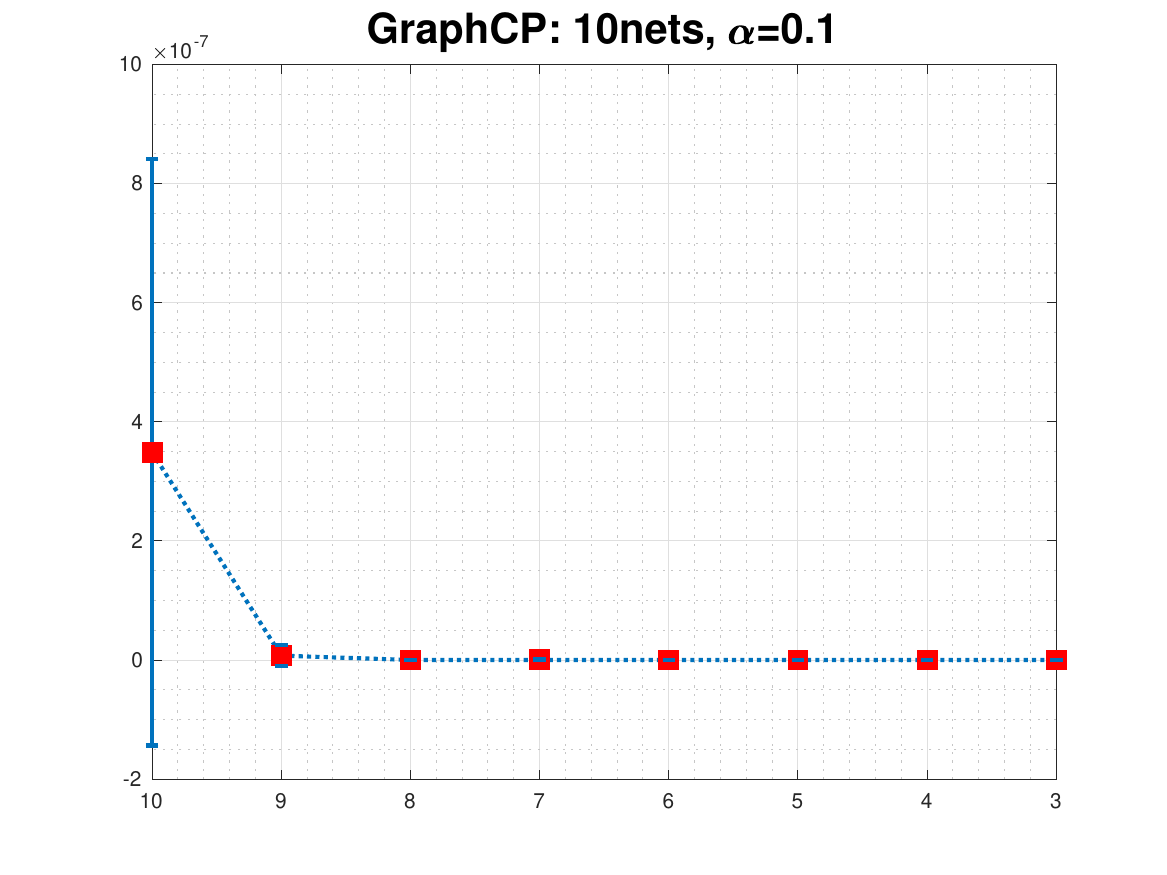} \\
				\multicolumn{3}{c}{(A) Alignment of 10 CT scans}\\
								\includegraphics[width=0.5\textwidth]{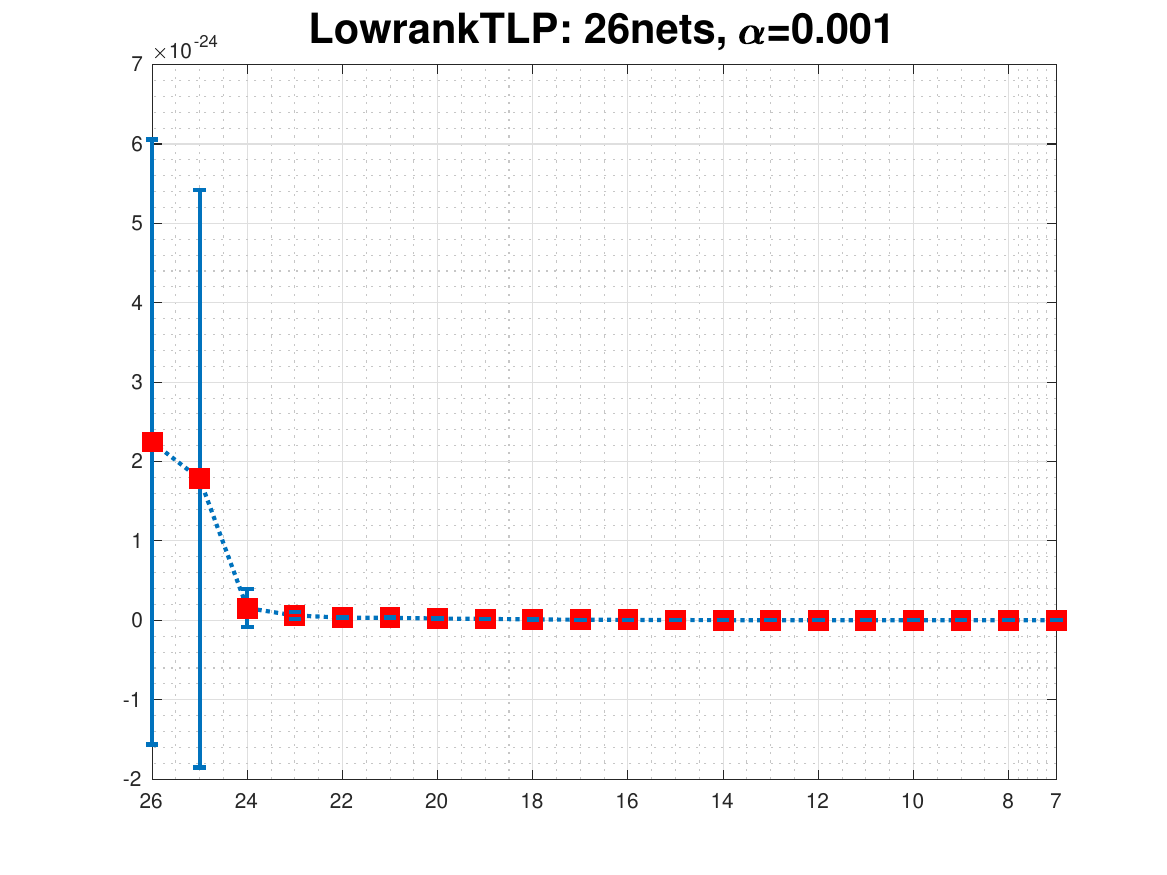} & 
				\includegraphics[width=0.5\textwidth]{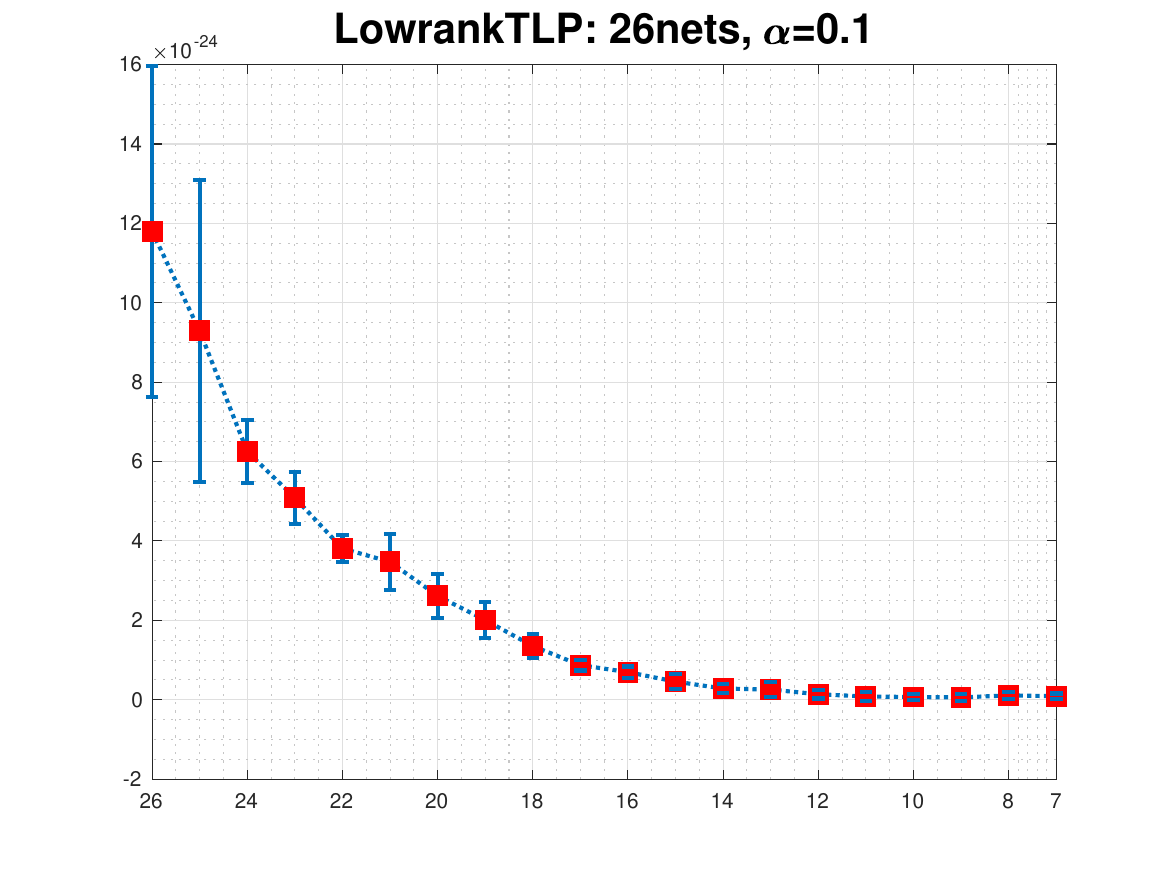} & \includegraphics[width=0.5\textwidth]{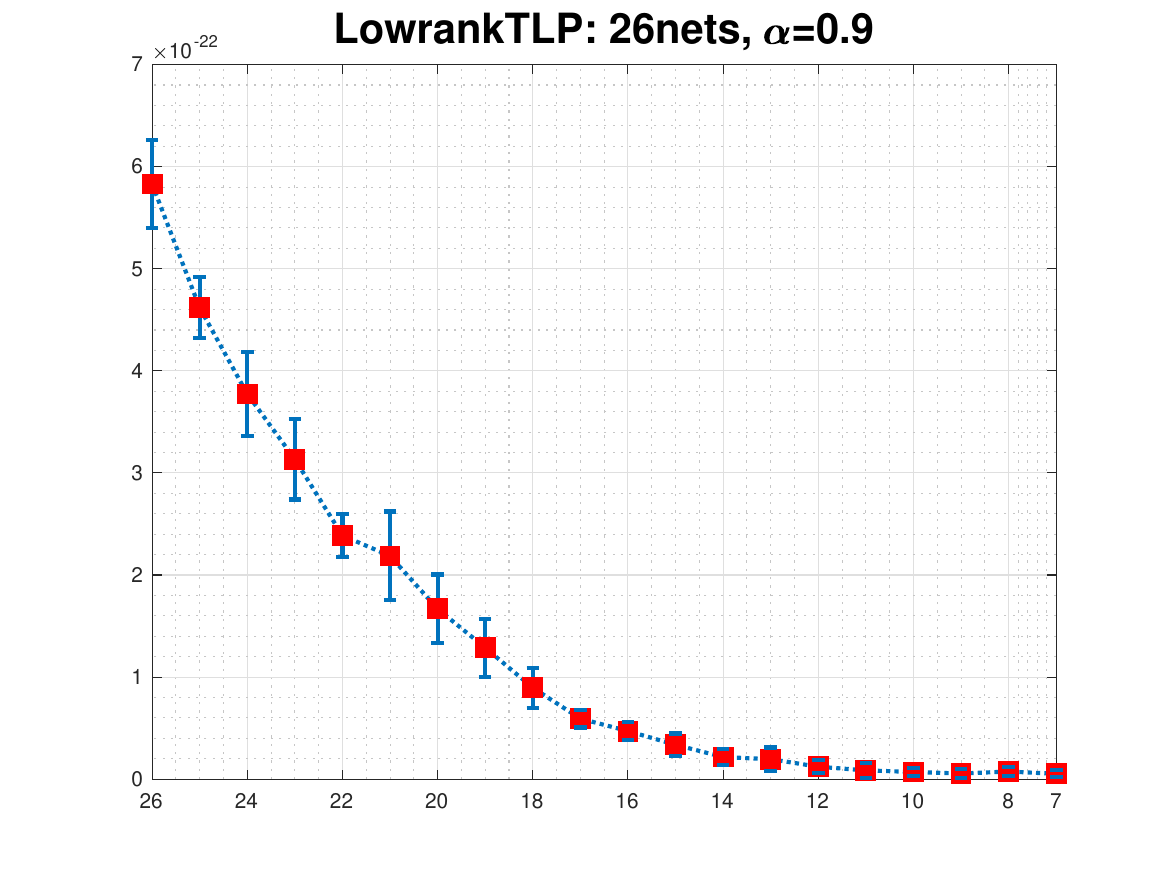} \\
				%\vspace{-8pt}
				\includegraphics[width=0.5\textwidth]{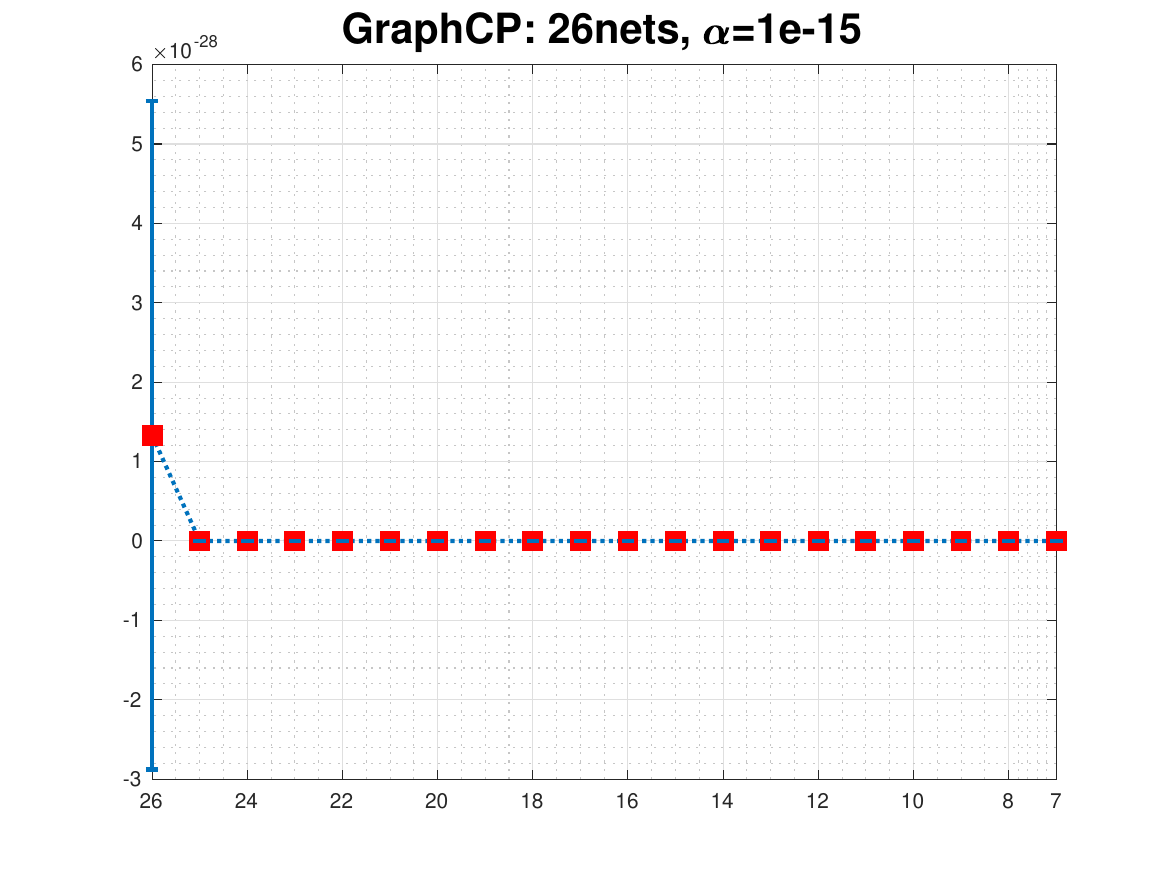} & 
				\includegraphics[width=0.5\textwidth]{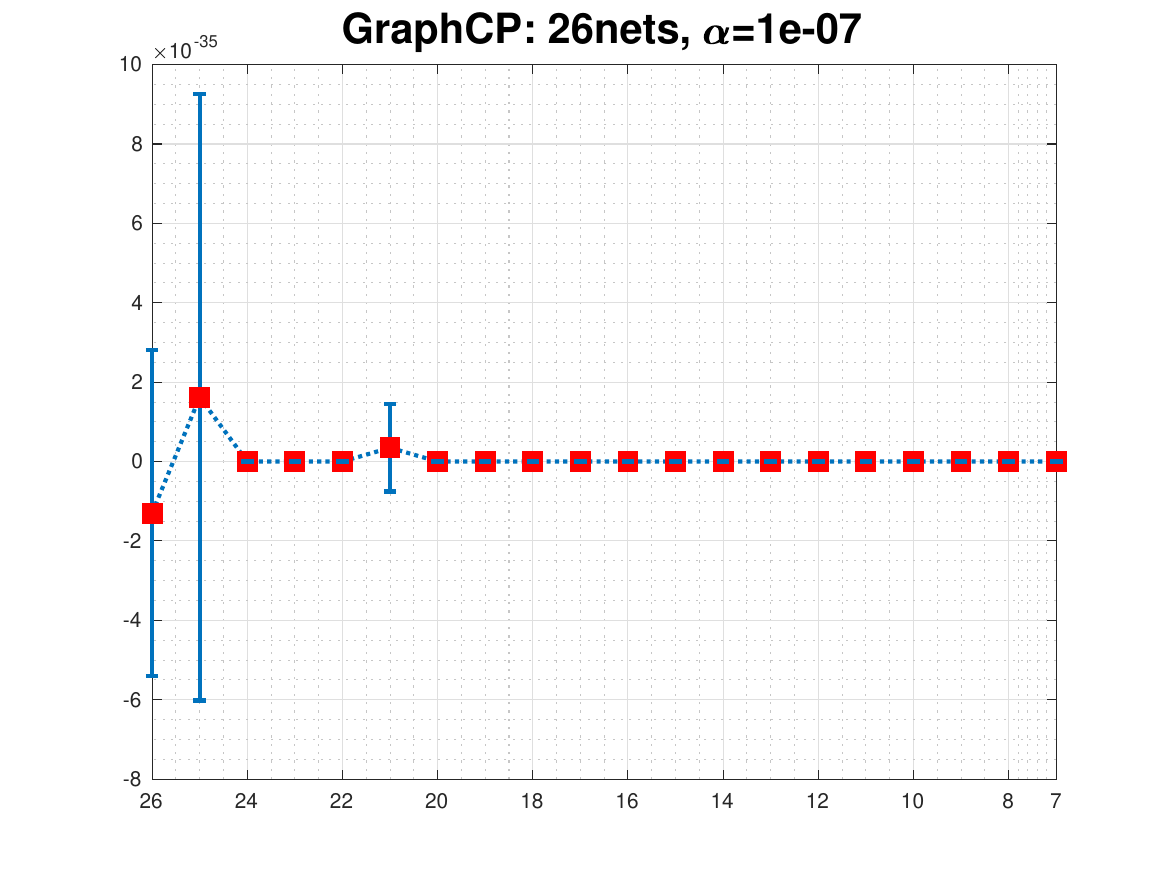} & \includegraphics[width=0.5\textwidth]{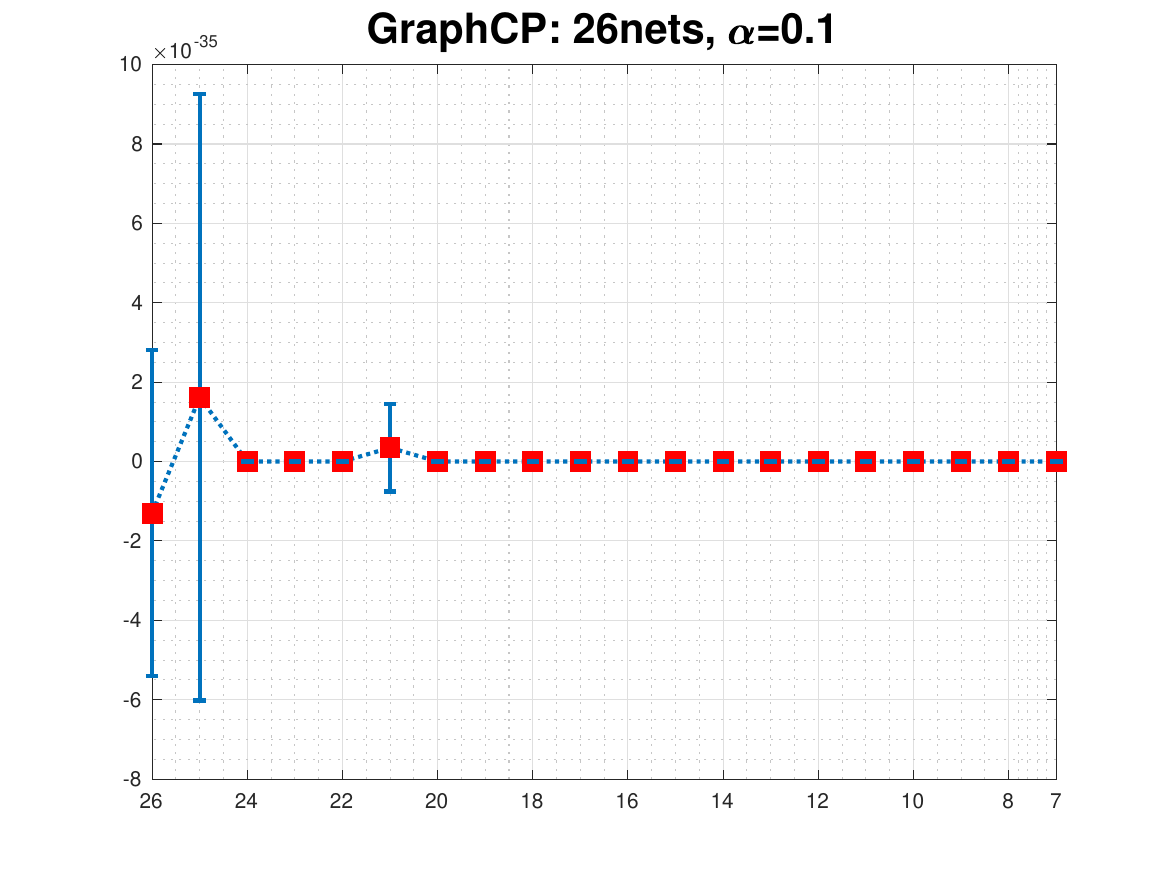} \\
				\multicolumn{3}{c}{(B) Alignment of 26 CT scans}\\
		\end{tabular}}
		\caption{\textbf{CT scan images alignment by varying the graph hyperparameter $\alpha$.} The x-axis are the sampling groups ordered by the value of $h$ defined in Section \ref{sec-CT}; the y-axis are the prediction scores.}\label{fig:app}
	\end{center}
\end{figure} 
\end{document}